\documentclass[acmtog]{acmart}

\usepackage{booktabs} 

\usepackage{url}
\usepackage{gensymb}

\usepackage{xcolor}
\usepackage{xspace}
\usepackage{subfigure}
\usepackage{multirow}
\usepackage{soul}
\usepackage{siunitx}

\renewcommand{\[}{\begin{equation}}
\renewcommand{\]}{\end{equation}}

\definecolor{darkgreen}{HTML}{3C8031}

\usepackage[ruled]{algorithm2e} 

\SetAlFnt{\small}
\SetAlCapFnt{\small}
\SetAlCapNameFnt{\small}
\SetAlCapHSkip{0pt}

\usepackage{soul}

\begin{document}

\title{Imaging for All-Day Wearable Smart Glasses}

\author{Michael Goesele} 
\author{Daniel Andersen} 
\author{Yujia Chen} 
\author{Simon Green} 
\affiliation{\institution{Meta Reality Labs Research}}
\author{Eddy Ilg} 
\affiliation{\institution{University of Technology Nuremberg*\thanks{*Contribution made while affiliated with Meta Reality Labs Research}}}
\author{Chao Li} 
\author{Johnson Liu} 
\author{Grace Kuo} 
\author{Logan Wan} 
\author{Richard Newcombe} \affiliation{\institution{Meta Reality Labs Research}}

\begin{teaserfigure}
\centering
\includegraphics[width=\textwidth]{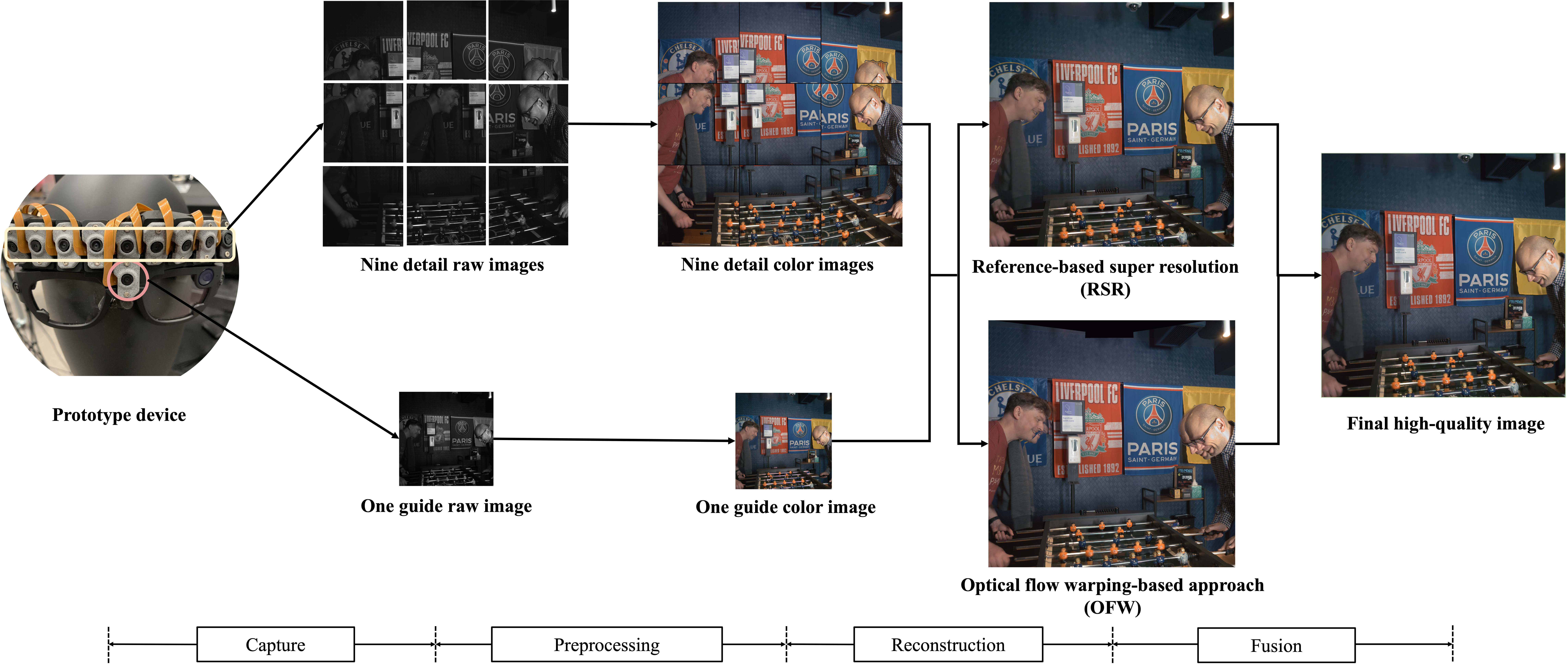}
         \caption{\label{fig:teaser}Full picture of the proposed distributed camera system. 
        Our software pipeline reconstructs a superresolved color output image. 
        Using a distributed camera prototype, we capture raw images mosaiced with a Bayer pattern that contain one color channel per pixel. In a preprocessing step, we obtain full color images, which serve then as input to two different reconstruction techniques -- reference-based superresolution (RSR) and the optical flow warping-based approach (OFW). 
        RSR usually has artifacts on the high-frequency details such as textured wall in the background while OFW has artifacts in the occluded regions (please zoom in to see the artifacts). The final fusion step combines the results from both techniques, yielding high-quality result images.
        }
\end{teaserfigure}

\begin{abstract}
\section*{Abstract}
In recent years smart glasses technology has rapidly advanced, opening up entirely new areas for mobile computing.  We expect future smart glasses will need to be all-day wearable, adopting a small form factor to meet the requirements of volume, weight, fashionability and social acceptability, which puts significant constraints on the space of possible solutions. Additional challenges arise due to the fact that smart glasses are worn in arbitrary environments while their wearer moves and performs everyday activities. In this paper, we systematically analyze the space of imaging from smart glasses and derive several fundamental limits that govern this imaging domain. We  discuss the impact of these limits on achievable image quality and camera module size --  comparing in particular to  related devices such as mobile phones. We then propose a novel distributed imaging approach that allows to minimize the size of the individual camera modules when compared to a standard monolithic camera design. Finally, we demonstrate the properties of this novel approach in a series of experiments using synthetic data as well as images captured with two different prototype implementations. 
\end{abstract}


\maketitle

\section{Introduction}

The last decade has brought superb advancements in mobile device imaging, primarily driven through the use-cases of mobile phone photography, video capture and video calling. We now routinely take photos and videos with our phones that significantly surpass the performance of specialized point and shoot cameras substantially larger than a phone available a decade ago. This is in part due to the miniaturization of camera module hardware including high performance CMOS sensors with small pixel size, highly optimized optical elements and assembly methods as well as MEMS-based focusing and optical image stabilization systems. However, just as important has been the rise of novel software processing that further enhances image quality, e.g., in terms of demosaicking, low light performance and video stability. Perhaps even more profound is the previously exotic, but now standard appearance of camera arrays on most high-end mobile devices - with multiple cameras enabling a single device to seamlessly capture a wide range of field of views at digital resolutions that match the optical performance of the single dedicated lens. The result is effortless capture, preview and sharing of imagery within one device. 

In this paper we look to the new frontier and problem spaces for imaging systems, namely enabling imaging for a broad range of new applications in wearables, and specifically in glasses form-factors. The rise of smart glasses with photography and video applications in recent years has also opened the door for applications in contextualized AI that allow users to query the world, and answer questions posed by a user in real-time by providing the egocentric view of the wearer to a new generation of deep learned ML computer vision models \cite{Ego4D2022CVPR}. The potential for super-human level applications of AI in wearable devices, brings with it the new challenges of requiring ever increasing field-of-view, depth-of-field, signal to noise ratio increases, higher frame-rate, lower motion blur, higher dynamic range data, and improved low light performance. These challenges must be resolved to enable applications which expect to match the performance of the human perception system but crucially require significantly reduced sizes relative to the human visual system. 

The future, always-on use-cases in contextualized AI demand all-day wearable devices which impose strict weight limitations~\cite{Kim:2021:WCP} for wearer comfort. Together with industrial design and social acceptability considerations, these in turn limit size, weight, power usage and the compute capabilities of the components significantly more than for mobile phone devices or other new imaging application areas, e.g., in autonomous vehicles and home robotics, capture systems (e.g., Project Starline~\cite{lawrence2021project}), or larger form-factor head-mounted displays as used in today's mixed reality applications.

In turn, we ask whether glasses form-factor imaging systems can ever match mobile phone level photography and videography performance, and look to uncover the limits of imaging systems in all-day wearable devices. What trade-offs do we face when attempting ever smaller, lower weight form-factors, and what physical limits are there in matching human level visual sensing, such that any down-stream task of the imaging system is not in principle limited by the imaging system itself? 
For the first time, we look across all requirements and constraints (excluding power consumption) for imaging systems in glasses form factors, and analyze them to uncover the fundamental tradeoffs in achieving photography and videography matching current mobile phone capabilities. We uncover the clear tradeoffs that exist for tiny imaging systems that, if solved, unlock not only wearable devices, but the adjacent areas of small form-factor robotics.

Our key contribution is a distributed imaging system that provides a design space that can enable significantly shrinking the size of the individual cameras  as used in mobile phones today by switching to an array of much smaller cameras. We achieve this by introducing a novel configuration of a camera array, distributed across the glasses frame, built with near diffraction limited cameras and varying refractive optical configurations. This replaces the single large camera or small number of cameras in mobile phone systems. This approach draws on the concepts of light fields \cite{levoy1996light} and lumigraphs \cite{gortler1996lumigraph} as well as super resolution and is different than recent approaches that aim to minimize the depth of the camera module without constraining the other two dimensions \cite{Venkataraman:2013:PUT,Chakravarthula:2023:TON}. The hardware is co-designed with the first algorithmic prototype for a processing pipeline that converts imagery from the array into a single high-resolution output image. While production of a single perspective image is not a prerequisite for AI applications, the value of demonstrating the potential for photorealistic capture from such a distributed system lies in enabling a direct comparison with mobile phones, and also enabling a simple interface to any application that today uses a single image or video as input. 

We structure the remainder of the paper to first derive the tradeoffs in the imaging system's properties, given constraints on the camera module form-factor. We then use data from Project Aria \cite{Engel:2023:PAA} to derive an understanding of how users move in natural situations and how these movements impact the achievable imaging quality. We then provide an overview of the proposed distributed imaging system and provide an analysis of its performance. Finally, we include a comprehensive discussion of the open issues in developing such a system further, especially under the competing constraints of enabling mobile phone class imagery versus the potential of such wearable imaging systems for use in AI applications.

\section{Previous Work}
\label{sec:prev}

Smart glasses have a long history, going at least back decades with Mann's seminal quest for an augmediated life \cite{Mann:2013:MAL}. In this paper, we focus in particular on devices that include a camera system. Prominent examples of this emerging product category include the (now discontinued) Google Glass,  Snap Spectacles \cite{Bipat:2019:AUC} as well as Ray-Ban Meta Smart Glasses, all of which make different trade-offs in terms of functionalities and feedback to the user. Related to this, the  Project Aria device \cite{Engel:2023:PAA} is a research vehicle designed to enable capturing research data using a smart glasses form factor device. It includes a versatile sensor suite, most importantly an 8\,mega pixel color point of view (POV) camera, two monochrome scene cameras with VGA resolution and inertial measurement units (IMUs) to support localizing the devices via simultaneous location and mapping (SLAM) and visualinertial odometry (VIO) \cite{engel2014lsd,mur2017orb,mourikis2007multi}.

\subsection{Imaging from Smart Glasses}

Imaging from cameras on smart glasses poses unique challenges that are different from those posed by handheld stand-alone cameras or mobile phones. For example, the extrinsic calibration of cameras can change dramatically due to the flexibility of the frames, which bend when worn by a user \cite{Wang_2023_CVPR}. 
In addition, the design of smart glasses combined with societal and individual factors affect usage scenarios for the cameras in smart glasses. For example, smart glasses with sunglasses-style lenses tend to be used more frequently in outdoor settings. Likewise, people tend to use smart glasses cameras  to capture novel events that can only be experienced from the wearer's own point of view \cite{Bipat:2019:AUC}. 
These use-case scenarios determine the properties of the image capture process, including the environmental illumination level and camera motion induced by the user. Earlier research has already analyzed the natural hand motion in a hand-held device such as  a mobile phone \cite{Wronski:2019:HMS}.
Head motion needs to be considered so it does not degrade image quality, but in certain cases it can be used to improve image quality, by providing extra information in a burst capture of multiple images for achieving super resolution \cite{Wronski:2019:HMS}. 

\subsection{Computational Imaging}
To go beyond the quality of what captured single images can offer, we can capture multiple images with different settings and use computational imaging techniques to combine the captured images, using all the information available in the raw data.
Multiple photos taken with varying exposure times can be merged together to create high dynamic range (HDR) images \cite{debevec1997recovering}, which requires registration between the individual photos if the scene and camera are not fully static \cite{Gallo:2015:LNR}. Photos captured from multiple perspectives can be combined together to create a 3D representation of the scene which allows rendering from novel viewpoints \cite{mildenhall2021nerf}. For extremely noisy raw images captured in dark environments, given enough input images, the rendered output can outperform dedicated deep raw denoisers applied to the raw input images \cite{mildenhall2022rawnerf}. Multiple photos captured at high frame rate with slight hand tremor from the capturing device can be combined together to create a super-resolution image by utilizing the aliasing in the captured images \cite{Liba:2019:HMP}. We will use similar ideas in order to recover high quality images when capturing a scene from smart glasses.

\subsection{Distributed Camera Systems}

Distributed camera systems use multiple sensors in order to jointly exceed a single camera's limitations, e.g. in terms of resolution, field of view, dynamic range or frame rate.

One approach is to use image stitching to obtain a high resolution image from a set of low resolution images.
The most salient work about image stitching is Autostitch \cite{brown2003recognising}, which introduced a generic pipeline. 
Autostitch was later further improved by introducing gain compensation, automatic straightening steps and an efficient bundle adjustment implementation \cite{brown2007automatic}. 
When extending image stitching methods to videos, temporal consistency needs to be enforced to avoid jittering in video \cite{yuan2017multiscale,perazzi2015panoramic}). 

Another important class of methods are the light field and lumigraph approaches \cite{levoy1996light,gortler1996lumigraph}. At a high level, they capture a slice of the plenoptic function and use it to re-render the scene from novel viewpoints and.
These approaches have later been extended in different directions such as  the ability to reconstruct dynamic events as captured by camera array with staggered camera trigger time using an optical flow-based  algorithm \cite{wilburn2005high}. Nomura et al. \shortcite{nomura2007scene} adopted a very smart way to generate scene collages using up to 20 cameras attached to a flexible plastic sheet, combining the benefits of both multiple cameras and temporal multiplexing.
Sahin and  Laroia \cite{Sahin:2017:LLC} describe the Light L16 camera, which combines ideas from both of these approaches to create fused images from the system's 16 cameras.

Besides the two classes of image synthesis methods, there has been some recent work on image fusion that takes images captured from multiple cameras with different properties, e.g. the multiple cameras on mobile phones. A U-Net like structure can help transfer features from low-quality images to high-quality ones \cite{trinidad2019multi}. Wu et al. \shortcite{wu2023efficient} also proposed a practical solution to this problem by considering the imperfection in reference images.

\subsection{Image and Video Super-Resolution}

Recently, super-resolution (SR) methods have emerged as a powerful tool that can be used to enhance raw captured images. One category of SR methods, reference-based super resolution methods (Ref-SR) that aim to super resolve a low resolution (LR) image with the help of a high resolution (HR) reference image, are particularly relevant for distributed camera systems, since multiple cameras provide a natural source of reference for SR applications. The Ref-SR problem can be formulated as a neural texture transfer problem to leverage texture detail with stronger robustness \cite{zhang2019image}. 
 $C^2$-matching \cite{jiang2021robust} improves the matching quality by explicitly considering the domain gap between HR and LR images. Attention mechanisms \cite{wang2021dual,pesavento2021attention,cao2022reference} and geometric information \cite{zou2023geometry} can also be added to the process to further improve the SR quality. It has also been shown that decoupling the super resolution task and the texture transfer task contributes to improved Ref-SR quality \cite{huang2022task}. Zhang et al. \shortcite{zhang2023lmr} extended the original Ref-SR problem to use multiple HR reference images instead of a single one. 

When considering the video super-resolution problem, we face new challenges such as maintaining temporal consistency and efficient computation. On the other hand, we also benefit from new information contained in multiple related frames. Ref-SR has been extended to the video reconstruction problem by propagating information between video frames in an efficient way \cite{lee2022reference,zou2023refvsr++,kim2023efficient}. Burst mode capture and SR reconstruction can be combined by rapidly capturing multiple images and taking advantage of the similarity amongst them \cite{bhat2021deep}.

\section{Fundamentals}
\label{sec:fundamentals}

In this section, we review basic concepts used to describe the properties of a camera system and discuss the relationship between them. We then look at two specific examples, a modern smart phone camera and the human eye, in order to use them as reference in the remainder of the paper.

To first order, a camera system can be defined by its \textit{field of view} (FOV), the range of angles it covers in horizontal (HFOV), vertical (VFOV), and diagonal (DFOV) direction, and the \textit{angular resolution}, typically given in arc minutes (arcmin) per pixel. Together they determine the number of physical pixels on the image sensor and thus the pixel count and size of the captured images. The actual size of each individual pixel, more specifically the optically active area, determines the \textit{sensitivity} of the sensor, i.e., its ability to capture images in dark lighting environments. The size of the sensor and the FOV determine together the required \textit{focal length} of the lens. Additionally, given the \textit{aperture size} of the lens, we can compute the \textit{depth of field} (DOF), i.e., the range of scene distances that can be in focus at the same time when imaged onto the sensor. All of these jointly determine to first order the size of the camera module, which integrates the lens, sensor as well as mechanical and electrical components.
As described above, a camera module has a fixed focus range. An \textit{auto-focus} system can shift that range depending on the scene content. Further, an \textit{optical image stabilization} (OIS) system can partially compensate for camera movements during an exposure, allowing the use of longer exposure times without introducing excessive motion blur. Both of these systems increase the size of the camera module and introduce additional power requirements.

The items above create a high-dimensional trade-off space, which is a simplified view of the full trade-off space. Any imaging system represents a specific point in this trade-off space. In the following, we give two examples for such systems. 

\subsection{Example Systems}
The main camera of an Apple iPhone 14 Pro,\footnote{\url{https://support.apple.com/kb/SP875?locale=en_US}} a modern high end smart phone, covers a 
FOV of 74$\times$53\,degree
with an average angular resolution of 0.49\,arcmin per pixel, yielding images with 48\,megapixel resolution. 
Given the 1.22\,$\mu$m pixel size\footnote{\url{https://www.gsmarena.com/apple_iphone_14_pro_max-11773.php}} yielding an optically active area of 9.76\,mm$\times$7.32\,mm, the additional space required for auto-focus and optical image stabilization as well as the size of the lens, the resulting camera module fits well onto a smart phone but is much too big and heavy for smart glasses. 

The human eye is a foveated imaging system that has the highest angular resolution of about 1 arcmin in a narrow foveal region with about 2 degree diameter. This angular resolution is equivalent to 20/20 vision, the normal (or corrected to normal) vision acuity at 20 foot distance tested by classical eye charts such as the Snellen chart. Overall, the human eye has a field of view of 151$\times$125\,degree (HFOV$\times$VFOV) \cite{Howard:1995:BVS} with rapidly decreasing angular resolution away from the fovea region \cite{Rosenholtz:2016:CLP}. The eye is able to focus at different distances with accommodation range strongly decreasing with age \cite{Mordi:1998:PAM}. Its overall size is close to spherical with a diameter of approximately 24\,mm \cite{Bekerman2014VariationsIE}. The human eye can see in a wide range of illumination conditions, covering a dynamic range of 14 orders of magnitude (ranging from $10^{-6}$ to $10^8 \frac{cd}{m^2}$ with reduced acuity in dark conditions \cite{Fang:2015:EHC,Hood:1986:STL}. The eye is able to perceive as few as 5 to 8 photons \cite{Hecht:1942:EQV} or even a single photon \cite{Tinsley:2016:DDS}). 

Matching  the performance of these systems on smart glasses is a worthwhile target. It would mean that we  achieve an image quality equivalent to a modern smart phone. It would also mean that we enable the glasses' machine perception algorithms to use the same quality input data as provided by the eye to the human visual system under arbitrary illumination conditions. However, the size, weight and power requirements of these systems are prohibitively large and do not fit the constraints for all-day wearable devices. In fact, space on smart glasses is severely restricted and will be an even more critical bottleneck as additional functionalities such as displays will be integrated in the future. Weight is also very important for comfort with standards for regular glasses recommending a maximum weight of the frame of 32\,g \cite{ISO12870}. In the  next section, we take a close look at the specific conditions and the trade-off space for smart glasses, which  will allow us to define the gap between the systems discussed in this section and an ideal smart glasses camera system.

\section{Limits for Smart Glasses}
\label{sec:limits}
All-day wearable smart glasses operate in a very specific setting: they are placed on a user’s head and capture and process data while the user is moving naturally in unconstrained lighting environments. This results in two major types of quality degradation: blur and noise. 
At the same time, they have constraints on size and weight as discussed in the previous section. 
Together these define the space of possible solutions and determine the overall viability of imaging systems for all-day wearable smart glasses.
In the following, we focus on the discussion of three aspects:
\begin{itemize}
\item{the limits on optical resolution due to diffraction and the connection to the achievable depth of field,}
\item{the impact of head motion on imaging on smart glasses, and}
\item{the amount of light available in typical scenarios and its impact on image quality.}
\end{itemize}

\subsection{Optical Resolution: Diffraction and Depth of Field} \label{sec:optical_resolution_limits}

To determine the achievable angular resolution of a camera, we consider two fundamental optical properties: diffraction and depth of field (DOF). Diffraction is caused by the wave nature of light and limits the achievable spot size on the sensor to
\begin{equation} \label{eq:diffraction_spot_size}
\varnothing_{\text{diff}} = 2.44 \lambda \tfrac{f}{D},
\end{equation}
where $\varnothing_{\text{diff}}$ is the spot size diameter, $f$ is the focal length of the lens, and $\lambda$ is the wavelength of the light (assumed to be 500\,nm for this analysis). 
$D$ is the entrance pupil diameter, defined as the image of the limiting aperture as seen through the lens from the object side. Note that the physical diameter of the lens cannot be smaller than $D$ but may be larger.
Using  $f$ we can convert from spot size to angular resolution with
\begin{equation}\label{eq:diffraction_angle_res}
    \delta\theta_{\text{diff}} = \tan \left(\frac{\varnothing_{\text{diff}}}{2f} \right)\approx  1.22 \tfrac{\lambda}{D}.
\end{equation}
Here we applied the small angle approximation, $\tan(x) \approx x$, and assumed that two spots are distinguishable if they are separated by the spot radius (similar to the Rayleigh criterion \cite{Rayleigh:1879:IOS}).

However, this resolution is only achievable at the focal plane of the lens, and limited DOF can further reduce the resolution. If we assume that it is generally desirable for optical infinity to be in focus, we can analyze the DOF by examining the hyperfocal distance $H$, which is the distance beyond which all content is in focus: 
\begin{equation}\label{eq:hyperfocal}
    H = \frac{fD}{2\varnothing_{\text{geo}}},
\end{equation}
where $\varnothing_{\text{geo}}$ represents the circle of confusion diameter.
As in Eq.~\ref{eq:diffraction_angle_res}, we can relate the circle of confusion to the system angular resolution:
\begin{equation}\label{eq:geo_angle_res}
    \delta\theta_{\text{geo}} = \tan \left(\frac{\varnothing_{\text{geo}}}{2f}\right) \approx \frac{\varnothing_{\text{geo}}}{2f}.
\end{equation} 

Finally, if the sensor pixel pitch, $p$, does not provide the resolution to adequately sample the spot size on the sensor, the angular resolution becomes limited by the pixel pitch. The instantaneous field of view (IFOV), which is the angle subtended by a single pixel, represents the pixel-limited angular resolution:
\begin{equation} \label{eq:ifov}
\text{IFOV} = \frac{p}{f}.
\end{equation}

Combined, the overall angular resolution, $\delta \theta$, is the maximum of the geometric resolution, the diffraction-limited resolution, and pixel-limited resolution:
\begin{equation} \label{eq:overall_reesolution}
\delta \theta = \max (\: \delta\theta_{\text{geo}}\:,\: \delta\theta_{\text{diff}}\:,\: \text{IFOV} \:).
\end{equation}

In this section, we assume that the pixel pitch is sufficiently small in order for it to not be the limiting factor, allowing us to analyze the relationship between diffraction and DOF independently of the sensor choice. 
To make the trade-offs more clear, we plug Eq.~\ref{eq:geo_angle_res} into Eq. \ref{eq:hyperfocal}, which gives a relationship between hyperfocal distance and angular resolution that is independent of system focal length,
\begin{equation} \label{eq:hyperfocal_trade_space}
    H = \frac{D}{4 \delta\theta} \quad \text{when} \quad D \geq 1.22 \frac{\lambda}{\delta\theta}.
\end{equation}
Here, the inequality defines a region that is not physically possible due to diffraction.

\begin{figure}[tp]
    \centering
    \includegraphics[width=0.8\linewidth]{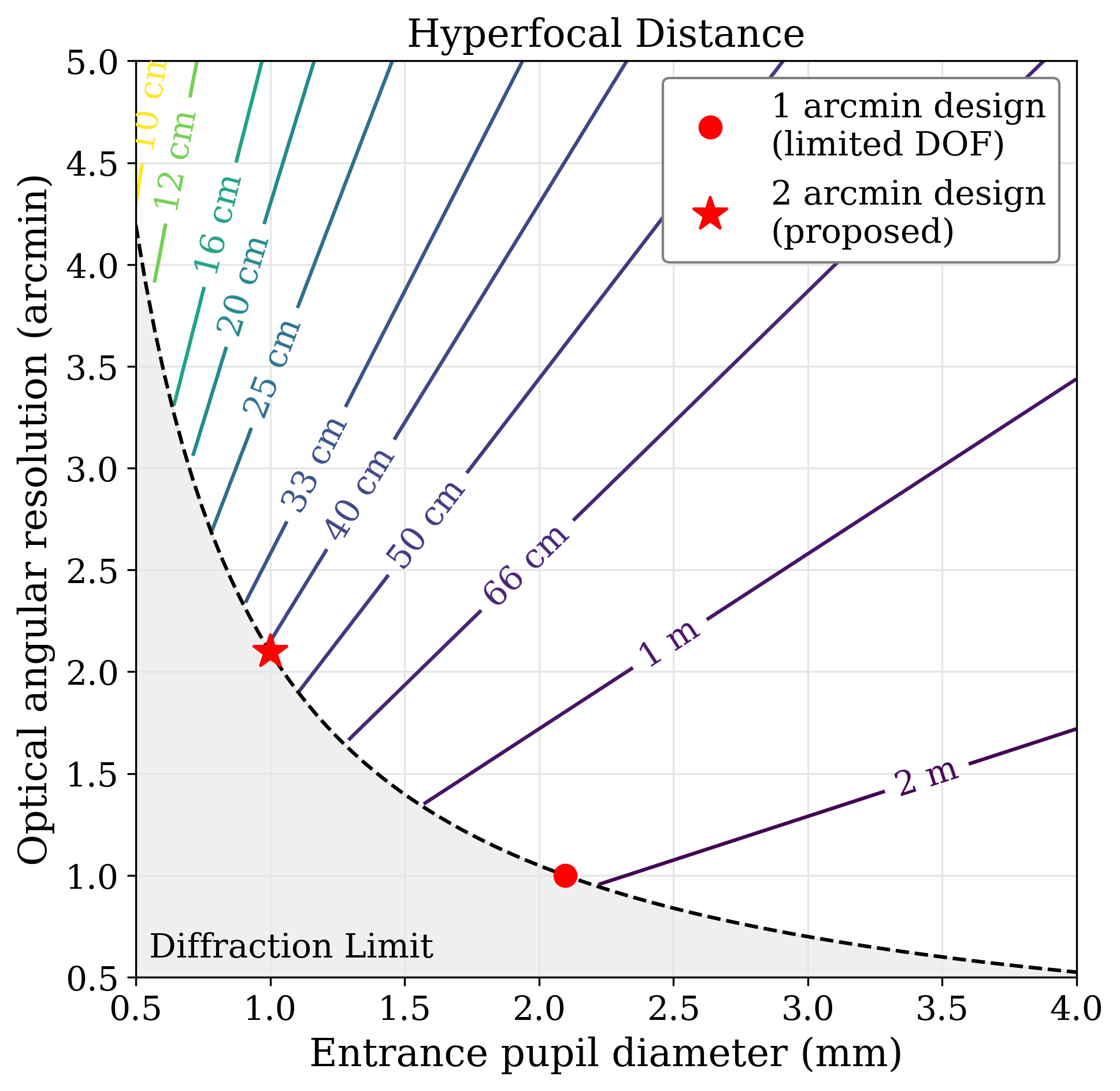}
    \caption{Trade-off for a fixed focus camera between angular resolution, entrance pupil diameter (which sets the minimum lens size), and depth of field (DOF),  represented  by the hyperfocal distance. Lenses within the gray region on the lower left are not physically possible due to diffraction, so achieving 1\,arcmin resolution (red diamond) requires an entrance pupil of at least 2\,mm. In addition, the long hyperfocal distance of such a lens means that only distant content, beyond about 2\,m, is in focus. Therefore, auto-focus is necessary adding additional size, weight and power consumption. Scaling back the resolution to 2 arcmin (red star) enables a long enough DOF that auto-focus is not needed. Further, this design can be achieved with a smaller lens diameter, making the system more compact. Given current technology, this is our recommended trade-off for imaging on smart glasses.
    }
    \label{fig:hyperfocal}
\end{figure}

This equation, which is plotted in Fig.~\ref{fig:hyperfocal}, represents some key trade-offs for imaging on smart glasses: it relates the resolution $\delta \theta$, the depth range over which we can achieve that resolution [$H \rightarrow \infty$], and the minimum lens (and thus camera module) diameter $D$, which is important when considering form factor. 
Trying to match the the properties of the human eye as described in Sec.~\ref{sec:fundamentals}, an ideal system would achieve at least 1\,arcmin resolution over a range from a comfortable reading distance (about 40\,cm) 
to optical infinity in a form factor that is as small as possible. 

Let us consider the trade space with these requirements in mind: If we desire a near human 1\,arcmin resolution,  Fig.~\ref{fig:hyperfocal} shows this is only possible with a lens diameter greater than 2\,mm due to diffraction (red diamond). Furthermore, the hyperfocal distance (about 2\,m) is much further than the desired distance of 40\,cm , which means that auto-focus is necessary to achieve this resolution over the full depth range. Auto-focus \cite{Zhang:2018:ASE} represents a range of challenges: the actuator and control electronics add size and weight, need a task-appropriate driving algorithm, and require more power. In particular, for always-on sensing on smart glasses, the actuator may need to be driven continuously, which requires more power than for photography applications where only a single snap shot is needed.

There is, however, an alternate design choice: By reducing the entrance pupil diameter to 1\,mm while keeping the resolution as high as possible (Fig.~\ref{fig:hyperfocal}, red star), we can achieve in-focus imagery over the full desired range. Although this additional DOF comes at the expense of resolution, the smaller diameter means that the system can be more compact, which is highly desirable. 
Finally, the lower resolution has the additional benefit of reducing the impact of motion blur due to head motion as shown in the next section.

\subsection{Head motion}
\label{sec:head_motion}

The human body is constantly in motion; in particular, the human head is capable of fast rotation with a rotational velocity of several hundred degrees per second. As a result, any image sensors on current or future headsets must address the issue of motion blur to avoid losing high-frequency scene details. This is especially important for high resolution image capture and for applications such as QR code reading from smart glasses that rely on high frequency scene details. In this section, we distinguish between two cases: general head motion while the glasses wearer performs regular tasks and user-stabilized head motion, where the wearer deliberately keeps the head still, e.g., in order to deliberately capture an image. While this allows us to draw conclusions about the required exposure times, we need to also relate this to the available light, which we will discuss in Sec. \ref{sec:available_light}.

\subsubsection{General Head Motion}
\label{sec:head_motion_general}

\begin{figure}[tp]
    \includegraphics[width=\linewidth]{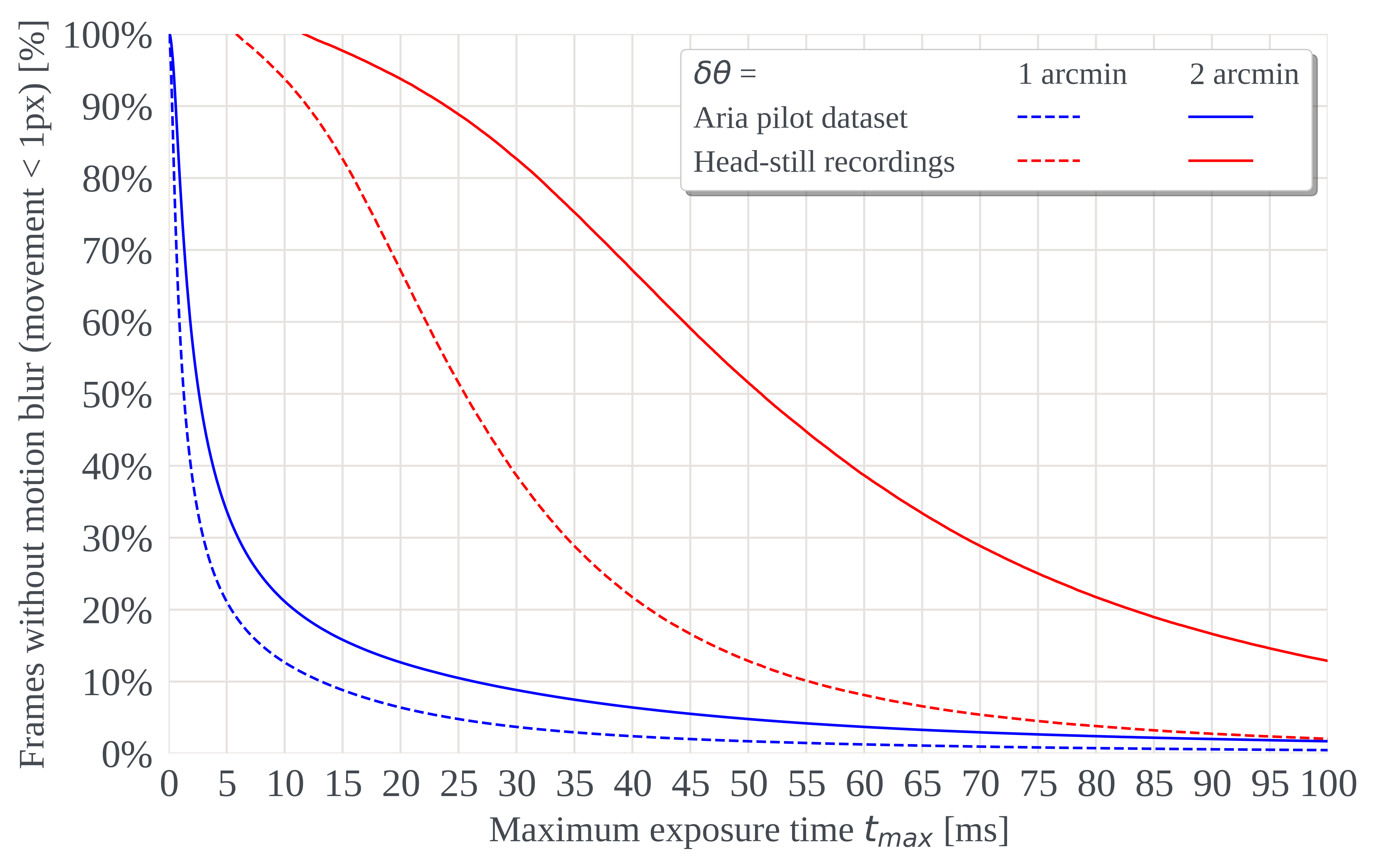}
    \caption{Percentage of egocentric image data with minimal motion blur (movement < 1px) as a result of exposure time, for a camera with $\delta\theta$ = 1 arcmin (dashed lines) and $\delta\theta$ = 2 arcmin (solid lines). Normal user movements put strict limitations on the achievable exposure time ({blue}). Users are, however, able to consciously hold their head still if desired, allowing for significantly longer exposure times ({red}).
    }
    \label{fig:motion_exposure}
\end{figure}

First, we examine head motion and its implications in the general scenario where a wearer is navigating their environment and performing everyday activities. For this analysis we selected 100 egocentric recordings from the Aria Pilot Dataset \cite{APD}, and aggregate the recordings' rotational velocity samples from the IMU data stream (see Sec.~\ref{sec:head_motion_user_stabilized_supplementary} in the supplemental material). Given this data and the angular resolution of the camera, we can determine a maximum exposure time that avoids motion blur. 

Based on the analysis in Sec.~\ref{sec:optical_resolution_limits}, and assuming a circle of confusion with a size less than 1px, we define an exposure to have minimal motion blur so long as the point spread function of a point source solely due  to head motion does not exceed 1\,pixel in size. Assuming that large-scale rotational velocity is relatively linear, we can infer the maximum exposure time $t_{\max}$ with minimal motion blur as
\[
 t_{\max} = \frac{\delta\theta}{\omega}.
\label{eq:max_exposure_time_minimal_motion_blur}
\]
The blue lines in Fig.~\ref{fig:motion_exposure} show that even with a short exposure time of 5\,ms, less than 35\,\% of captured egocentric data with this distribution  would yield minimally motion-blurred frames under an angular resolution of 2\,arcmin.

\subsubsection{User-stabilized head motion}
\label{sec:head_motion_user_stabilized}

While the human body is constantly in motion, there are regular periods in which head motion is stabilized either consciously or unconsciously (e.g. while reading, watching TV, or purposely posing). To analyze this separate class of human movement, we recorded a set of ten Project Aria sequences in which wearers sat, stood, or read a book while consciously keeping the head as still as possible. We extracted sub-sequences from these recordings for which all IMU rotational velocities were below 3 degrees per second, which we empirically found well-bounded by the ``keeping head still'' behavior of the Aria wearer (see Sec.~\ref{sec:head_motion_user_stabilized_supplementary} in our supplementary material for details). The red lines in Fig.~\ref{fig:motion_exposure} show the distribution of maximum exposure times for minimally-motion-blurred images when considering only this category of wearer behavior. When compared with general head-motion, egocentric cameras can safely increase exposure times during head-still activities without incurring motion blur. Furthermore, back-and-forth subconscious stabilizing motions by the user means that the previous assumption of linear rotational velocity is actually an upper-bound in the case of head-still activities (see Sec.~\ref{sec:head_motion_user_stabilized_supplementary} in the supplementary material).

\subsection{Available Light}
\label{sec:available_light}

Decreasing the exposure time is the primary way to reduce the motion blur discussed in the previous section. However, it also decreases the amount of signal captured and leads to a lower signal to noise ratio (SNR) in the captured images. We therefore analyze in this section the number of photons that are captured per pixel under different lighting conditions and exposure times, and discuss how this impacts image quality in terms of signal to noise ratio.

\subsubsection{Physical Model}
In this section, we model the number of photons impinging on a sensor pixel for a natural scene lit at an illuminance level $E$ given in lux following \citet{Alakarhu:2007:ISI}. Given $E$, we first calculate the number of photons $N_{\text{ph}}(\lambda)$ of a given wavelength $\lambda$ impinging per time unit onto a surface in the scene as:
\begin{equation} \label{eq:scene_photons}
    N_{\text{ph}}(\lambda) = \frac{1}{683 \tfrac{\text{lm}}{\text{W}} } \cdot\frac{s(\lambda)}{ \int s(\lambda)V(\lambda) d\lambda} \cdot \frac{\lambda E}{hc}
\end{equation}
where $V(\lambda)$ is the standard luminosity function (CIE 1926), $h$ is the Planck constant, $c$ is the speed of light, and $s(\lambda)$ is the  spectrum of the light (we assume CIE Standard Illuminant A, representing typical domestic lighting). 
Next, we determine the total number of photons per pixel on the sensor, $N_{\text{ph}}$, which depends on both scene and camera characteristics:
\begin{equation} \label{eq:photons_per_pixel}
    N_{\text{ph}} = \frac{RT}{4} \cdot \frac{D^2p^2t}{f^2} \int N_{\text{ph}}(\lambda) C(\lambda) d\lambda.
\end{equation}
Here, $R$ is the scene reflectance, $T$ is the lens transmission, $D$ is the entrance pupil diameter, $f$ is the focal length, $p$ is the pixel pitch, and $t$ is the exposure time. $C(\lambda)$ represents wavelength-dependent transmission, which most notably includes the color filter array.

To showcase the upper limit of available light in an ideal case, we assume no transmission loss in the lens ($T = 100\%$) and assume ideal color filters with perfect transmission (420\,nm-500\,nm for blue, 500\,nm-600\,nm for green, and 600\,nm-680\,nm for red). To describe an average scene, we set $R = 18\%$ which represents a middle gray.
Applying these lighting and camera assumptions in Eqs.~\ref{eq:scene_photons}-\ref{eq:photons_per_pixel} yields the following simplified upper bound between illuminance and photons per pixel:
\begin{equation}
    N_{\text{ph}} = k \cdot \frac{D^2p^2t}{f^2} \cdot E 
    = k \cdot {D^2(\text{IFOV})^2t} \cdot E, 
\end{equation}
where $k_\text{RGB}$ is a constant for each color channel:
\begin{equation}
 k_\text{RGB} =(0.6262, 1.654, 1.271) \cdot \frac{10^{14}}{s\cdot m^2\cdot lux}. 
 \end{equation}

\subsubsection{Impact on Image Quality}

To connect the number of photons $N_{\text{ph}}$ to image quality, we consider the best-case scenario SNR, in which all photons hitting the sensor are converted to electrons (perfect quantum efficiency) and there is no camera read noise. The only noise source is then Poisson noise, caused by the discrete nature of photons:
\begin{equation}
 \text{SNR}\leq\sqrt{N_{\text{ph}}}\,\mathrm{.}
\end{equation}
As a general rule of thumb, images with SNR=40 are considered ``first excellent'' images, and images with SNR=10 are considered ``first acceptable'' images \cite{ISO12232}. Therefore, to achieve ``acceptable'' images, even in this best-case scenario, requires at least 100 photons per pixel.

\begin{figure}[tp]
    \includegraphics[width=\linewidth]{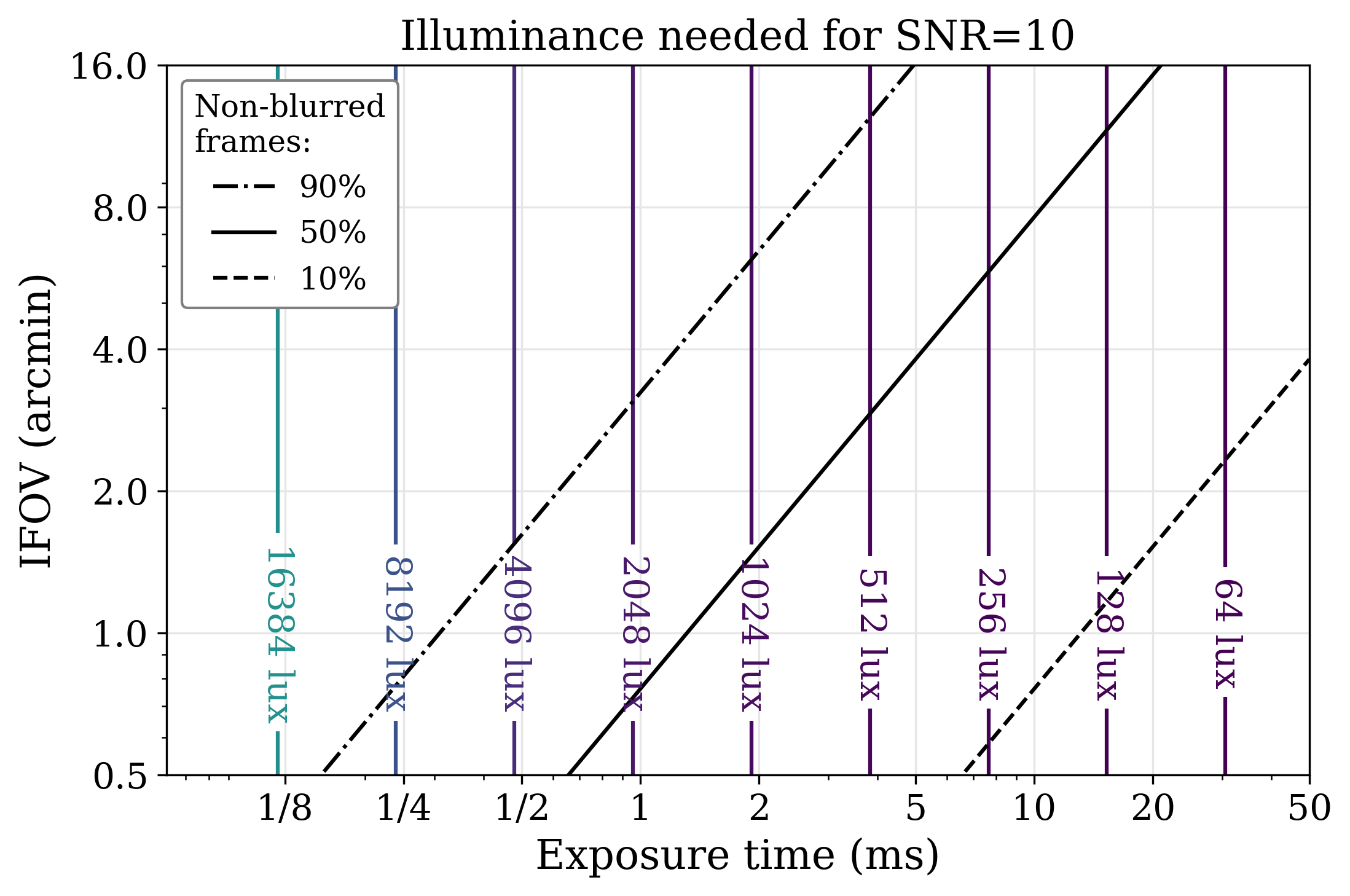}
    \caption{Illuminance levels required to achieve SNR=10 for a hypothetical camera module with f-number 1.8 and a sensor pixel pitch of 1\,$\mu$m. The black lines relate this to the exposure time required to avoid motion blur as shown in Fig. \ref{fig:motion_exposure}. E.g., reaching SNR=10 and 50\,\% non-blurred frames under the Aria pilot dataset motion distribution using a camera with an IFOV of 2\,arcmin requires an exposure time of 2.6\,ms and an illuminance of approx. 800 lux.
    }
    \label{fig:snr10}
\end{figure}

Fig. \ref{fig:snr10} plots along the x axis the illuminance $E$ needed in order to collect enough photons to reach SNR=10, assuming a hypothetical camera module with f-number 1.8 and a sensor pixel pitch of 1\,$\mu$m. This SNR level can, e.g., be reached at an illuminance $E=2048$\,lux with exposure times of 1\,$\mu$s or more. The diagional lines mark the maximum exposure time possible to reach a given percentage of non-blurred frames under the Aria pilot dataset motion distribution.
Fig. \ref{fig:snr10} shows that for a typical illumination level of a bright office room of about 500\,lux \cite{pears1998strategic} more than half of the images captured under the motion distribution of the Aria pilot dataset with an IFOV of 2\, arcmin will either be blurred or have a SNR<10 and almost  all images captured at an  illumination level of a typical family room (50\,lux) will be severely degraded by either noise or motion blur. In contrast, capturing in bright, sunny outdoor environments with $E=50,000$\,lux will almost always yield high quality results.

\subsection{Summary}
Combining the insights from the analyses in this section, it is not impossible to match the imaging performance of our two reference systems discussed in Section \ref{sec:fundamentals}, namely an iPhone 14 Pro and the human eye, which both have a maximum angular resolution of 1\,arcmin or better. However, any such system 
will need to be equipped with an auto-focus system in order to cover the desired focus range from about 40\,cm to infinity as discussed in Section \ref{sec:optical_resolution_limits}. Moreover, achieving high quality results is only possible under narrow conditions when the scene is very strongly lit and the user deliberately limits the motion of their head, significantly limiting its usefulness. 

On the other hand, lowering the angular resolution to 2\,arcmin yields a different point in the design space with several advantages compounding each other: First, we do not need to add an auto-focus system in order to keep all desired scene content in focus. Second, the image resolution and thus the number of pixels and the sensor size reduces by a factor of four compared to an imaging system that covers the same field of view at 1\,arcmin, saving both on system size and power needed to operate it. Third, the lower angular resolution means that the imaging system is less sensitive to motion blur and allows the use longer exposure times under the same lighting and motion conditions, improving the signal to noise ratio significantly. Finally, linking back to our original goal of minimizing the form factor, the sensor and lens can be made significantly smaller, leading to a more desirable form factor of the camera module.

\section{Distributed Imaging Concept}
\label{sec:plato}
While reducing the angular resolution to 2\,arcmin as discussed in the previous section leads to a very beneficial point in the design space, its impact on the size of the camera module is limited. Ultimately, one of the key drivers is the resolution of the output images, which directly determines the size of the sensor and thus the size of the lens. 
In order to further miniaturize the camera module as discussed in the introduction to facilitate easy and flexible integration into an all-day wearable glass-form factor,
we propose a distributed imaging system that has the goal to replace the single monolithic camera module considered so far with multiple, more compact modules. 
This allows for the seamless integration of cameras into all-day wearables in order to make them socially acceptable.
In the following, we will discuss three options for this, which effectively split the camera module in different dimensions and contain different levels of redundancy.

\begin{figure*}[tp]
\includegraphics[width=0.32\textwidth]{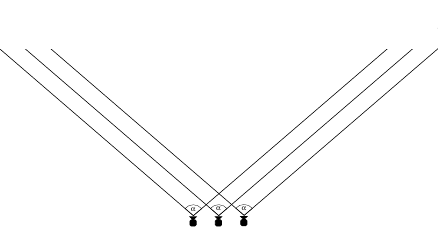}\hfill
\includegraphics[width=0.32\textwidth]{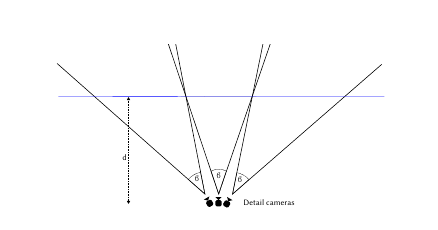}\hfill
\includegraphics[width=0.32\textwidth]{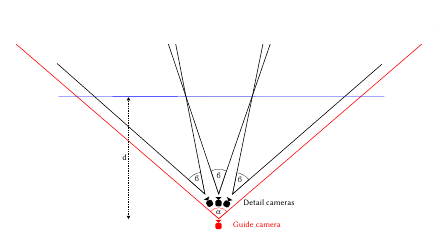}
\caption{\label{fig:camera_layouts}
Distributed camera layout design visualized in 2D. \textbf{Left:} three wide FOV cameras pointing to the same scene. \textbf{Middle:} three narrow FOV camera tiling together to cover the full scene. \textbf{Right:} one wide FOV camera covering the full FOV and three narrow FOV cameras tiling together to capture details in the scene. Note that close to the detail cameras, the fields of view do not overlap and $d$ represents the closest distance at which objects are captured by all detail cameras. 
}
\end{figure*}

One strategy is to keep all the small cameras at the targeted FOV, but capture the scene from slightly different perspectives (caused by their baselines) with a much reduced resolution, see  Fig.~\ref{fig:camera_layouts} (left). The advantage of this strategy is that each of the captured images contains a full view of the overall scene apart from occlusion and disocclusion artifacts. Assuming a perfect pinhole imaging model without diffraction, the set of cameras could capture enough high-frequency scene content to allow for a full reconstruction. However, all of this high-frequency content is lost in a practical system due to the blur induced by the lens and the finite size of the pixels in the sensor, making this option not practically viable.

The second strategy is to keep the angular resolution of  all cameras but reduce their field of view as shown in Fig.~\ref{fig:camera_layouts} (middle). Given a careful alignment, ensuring that the full space is covered from a given minimum distance $d$ until infinity, the cameras cover jointly the scene at the full resolution. There are, however, still two major issues: On one hand, integrating the images from the individual cameras into a single coherent photograph is error-prone given the small overlap between the images. On the other hand, the finite baseline between the cameras will still lead to occlusions in the scene. 

The third option, which we propose to follow,  combines the first two strategies and benefits from their advantages.
In this design, both high-resolution information and a global view of the entire region of interest are captured at the same time, 
see Fig.~\ref{fig:camera_layouts} (right).
The system consists of two types of cameras: A guide camera with low angular resolution and large field of view and multiple detail cameras with  high angular resolution but small field of view. 
We use the guide camera capture an image of the full field of view, called a \textit{ guide image}. 
The detail cameras are oriented in a way such that they jointly cover the  entire field of view 
of the guide camera at high resolution. Images captured by this type of camera are called \textit{detail images}. 
The advantage of this approach is that we can use the guide image as the backbone of the reconstruction, fusing the high-frequency details obtained from the detail images onto this view. The guide image also serves as a fallback that can cover any holes caused by occlusion, i.e., to cover any areas that are not imaged in any of the detail views. Fig.~\ref{fig:teaser} shows the concept diagram of our proposed distributed camera sensing system and the reconstruction approach used to recover a single fused image. 
We note that this opens up a flexible design space: Depending on the overall field of view, the resolution, the ratio of angular resolutions between guide and detail images as well as the desired size of the individual camera modules, different numbers of detail cameras can be used to tile the overall field of view.

With the camera designs in place, we still need a computational imaging algorithm that is capable of synthesizing a high resolution image from the raw information from all cameras, which will be discussed in the next section.

\section{Reconstruction Pipeline}
\label{sec:processing_pipeline}
In the previous section, we have discussed how to split a single camera into multiple cameras with small form factor that cover either the complete FOV with small angular resolution, which we refer to as the \emph{guide} image, or a small FOV with high angular resolution, which we refer to as \emph{detail} image. As shown in Figure~\ref{fig:camera_layouts}, in the end we need both -- complete FOV and high angular resolution. Using several such detail cameras allows capturing the complete FOV at the desired angular resolution. However, na\"ively stitching them together to a single high-resolution image is not possible, as the detail images are all from slightly different points of view. Correct reprojection of the detail images to the high resolution version of the guide (here called \emph{target}) view requires us to know the depth of the scene in order to establish reliable correspondences.

\begin{figure*}[tp]
\includegraphics[width=\linewidth]{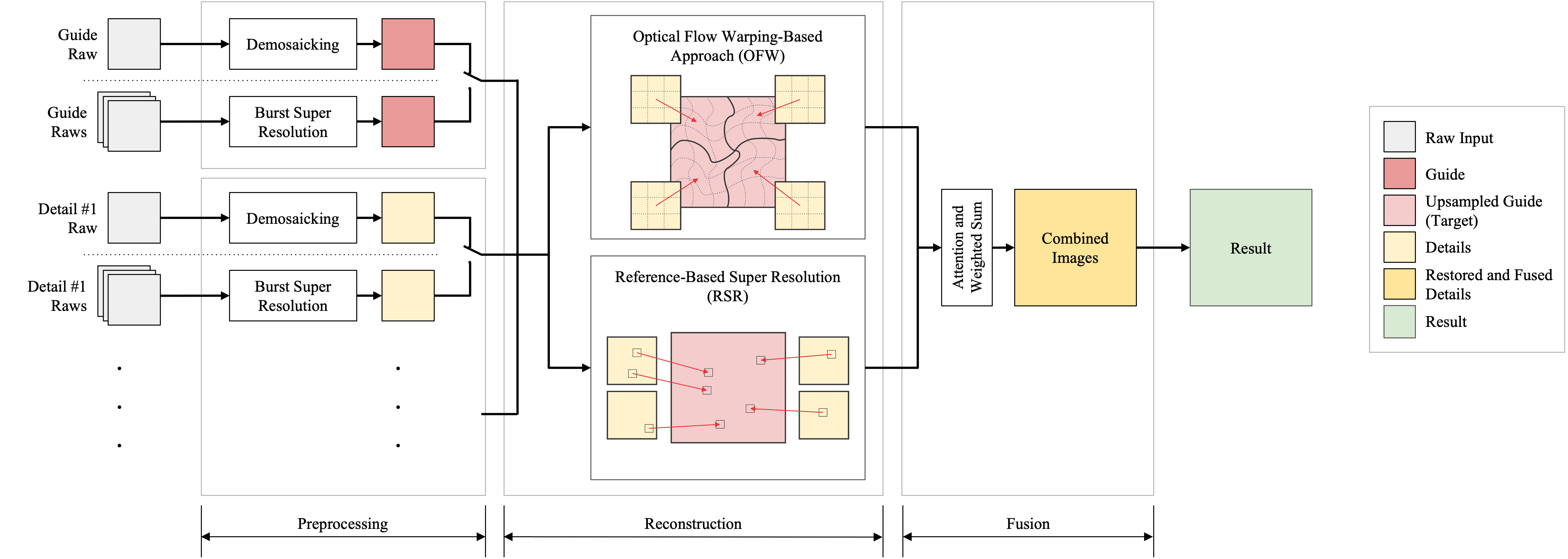}
\caption{\label{fig:pipeline_overview}
    Overview of the image processing pipeline. The input is either a single set of guide and detail images or respective bursts of images. After preprocessing, we present two orthogonal reconstruction approaches that achieve sharpness and robustness, respectively, and fuse the results of both together to obtain a high-quality photograph. 
}
\vspace{2ex}
\includegraphics[width=0.95\linewidth]{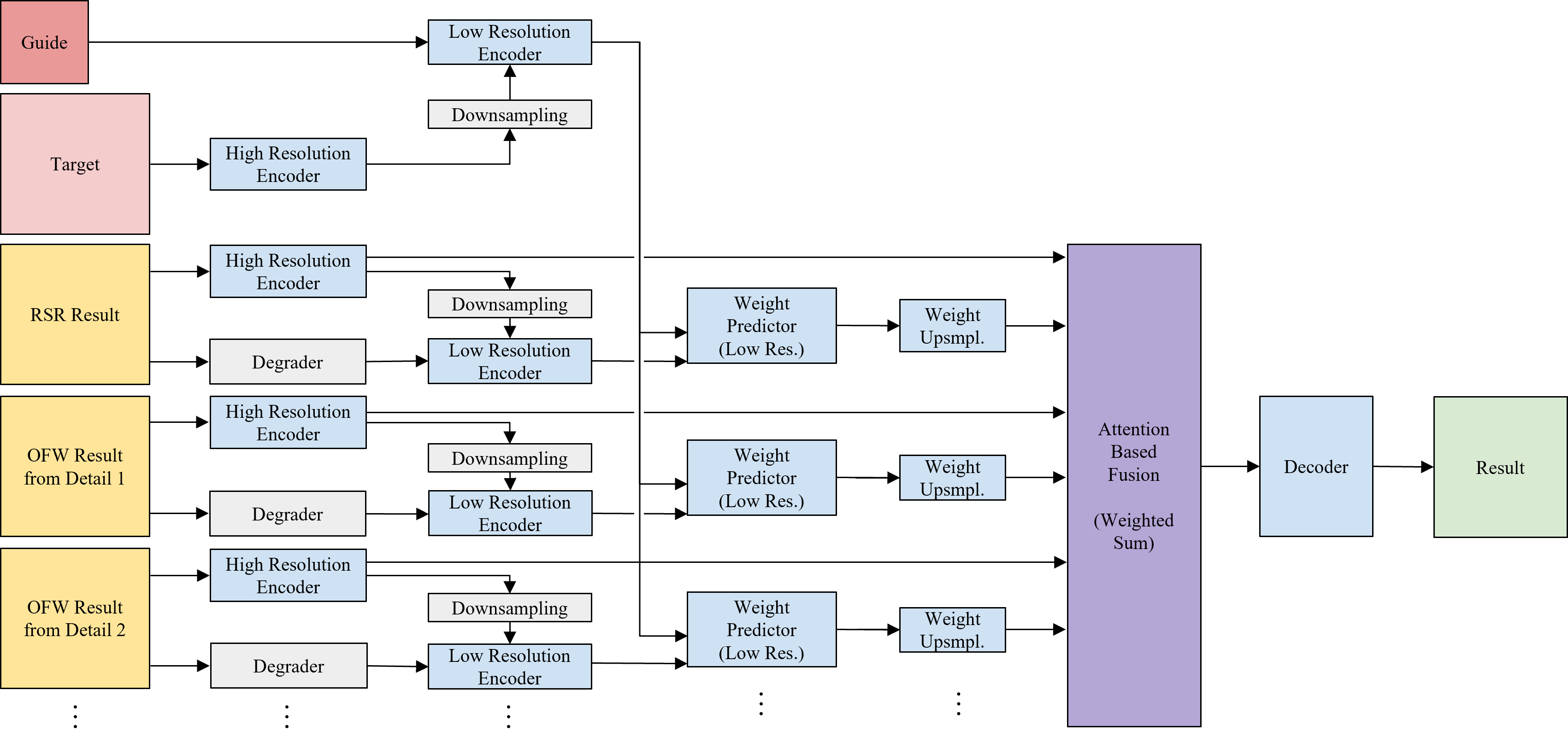}
\caption{\label{fig:fusion_arch}
    Architecture of the fusion network. 
    The ``degrader'' uses the known properties of our cameras to downsample and degrade the high-resolution images back to the quality of the input images. The output of the degrader is used to compute features on all low resolution images that are then fed into a weight predictor. The features of the high-resolution images are then fed into an attention module that computes their weighted sum. In the end, the combined features are decoded to the final result. 
    }

\end{figure*}

Depth estimation of the whole scene only from the detail images is not possible, though, as they overlap only by a small amount. To this end, we leverage the guide image that already captures the desired target view at a low angular resolution. Instead of depth, we estimate the optical flow with some soft epipolar constraints and subsequently warp the detail images to the target view. Assuming that the optical flow is correct this leaves all the high-frequency details captured in the detail images intact and can potentially provide the highest quality reconstruction. However, this approach fails gracelessly when the optical flow is incorrect. We therefore introduce a second orthogonal approach that leverages reference-based super resolution. This approach is very robust to incorrect correspondences, while it trades off some of the sharpness. Finally, we present a method to fuse the results of both approaches to obtain the best of both worlds. In the following, we will discuss the individual  components of our reconstruction pipeline (see also the overview in Figure~\ref{fig:pipeline_overview}).

\subsection{Preprocessing}
\label{sec:djdd}

The images captured by our prototype rigs are raw images with color filter array (Bayer pattern) applied, and need to be demosaicked to yield full RGB color images. 
We use deep joint demosaicking and denoising (DJDD) \cite{gharbi2016}, an end-to-end trainable approach, that provides high quality results. 

\begin{figure}
\centering
\includegraphics[width=\linewidth]{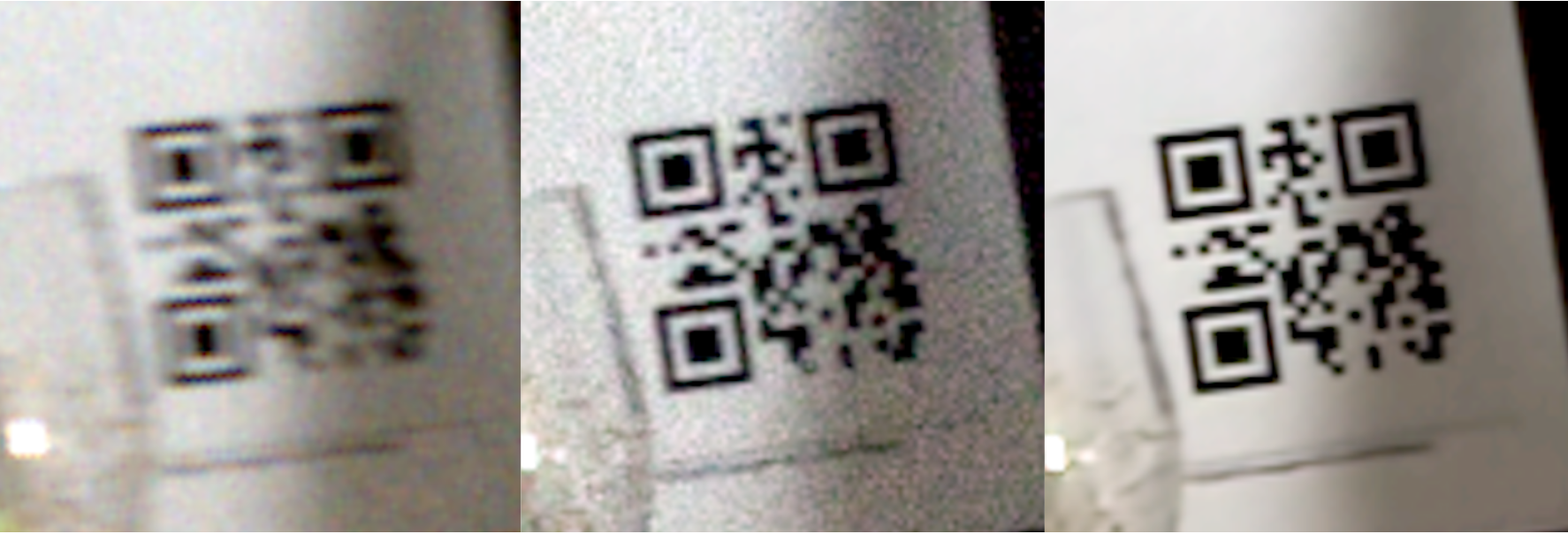}
\caption{\label{fig:burst_motion_blur}
    Comparing single image vs. burst mode. Left: capturing a single image with a long exposure time results in limited noise and high motion blur. 
    Middle: shortening the exposure time reduces the motion blur, but significantly increases the noise. Right:
    Using eight noisy images captured with short exposure time (same conditions as the middle one) and applying burst mode reconstruction results in a clean image. 
    }
\end{figure}

As discussed in Section~\ref{sec:head_motion}, head motion can lead to significant motion blur. 
We therefore add an alternative preprocessing option using burst mode. In this mode, we rapidly capture a set of images using shorter exposure times to limit the motion within each image's exposure time.  The high frame rate ensures that there is significant overlap between the images. 
We then apply a version of the deep-learned burst mode super resolution pipeline by Bhat et al. \shortcite{bhat2021deep}, where we remove the up-sampling components and use the pipeline only for denoising and demosaicking. An example result is shown in Fig.~\ref{fig:burst_motion_blur}. 
The first image with long-exposure time has limited noise, but the motion blur makes the QR code hard to read. The second image with short exposure time has limited motion blur, but is very noisy. Burst mode reconstruction effectively removes the noise and reduces motion blur. 

\subsection{Reconstruction}
\subsubsection{Optical Flow Warping-Based Approach (OFW)}
\label{sec:beta}

An intuitive approach to reconstruct the target image is to upsample the guide to the target resolution and estimate optical flow between the detail images and the resulting target image to subsequently warp the detail images to the target view. 
To obtain high-quality correspondences we introduce several adaptations of optical flow estimation to our particular setting that we will describe in the following paragraphs. In practice, we find that our adapted optical flow estimation approach provides maximally \emph{sharp} reconstructions in the majority of image regions. However, we also observe failure cases where the optical flow is not correct. It is important to note that with this technique, 
incorrectly warped detail images will result in severe artifacts when combined together by a simple blending. Even more so, regions that are visible in the target but occluded in the detail images can never be restored with this technique. Note, e.g., the double image of the hand in Fig. \ref{fig:gamma_vs_beta}, second row, ``OFW'', that is caused by these occlusion effects.

In order to be able to overlay the warped detail images correctly, we require optical flow at subpixel accuracy, which is non-trivial to obtain in general and even more difficult between images with different frequency content. We start by building upon RAFT \cite{teed2020raft} and make several extensions. First, we adapt the approach to work with two input images of different size by padding the input images to the same size. We then process each detail image individually by using the respective detail and the target as inputs. 

\textbf{Prewarping:} We found that directly using the detail and target requires solving for very large displacements, which does not work well in practice. Hence, we initialized the correspondences with a homography obtained from a plane at 100\,m distance. 
This also allows us to determine 
the region in the target image in which the detail can maximally be visible, and crop the target accordingly. 

\begin{figure}[tp]
\includegraphics[width=\linewidth]{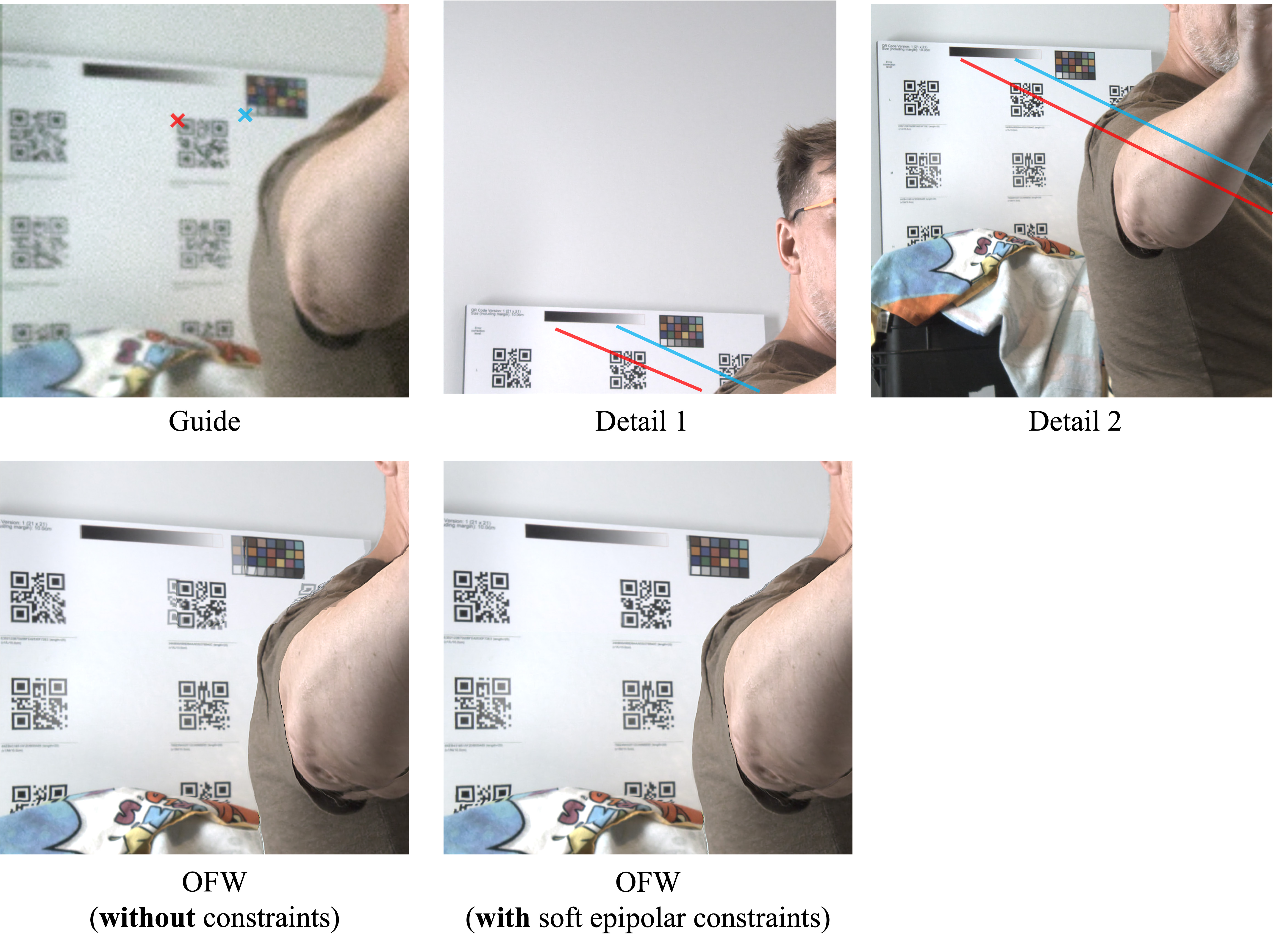}
\caption{\label{fig:epi}
    Effect of the soft epipolar line constraint in OFW. \textbf{Row 1:} Epipolar lines in two detail images for two selected pixels in the guide image over a depth range of 0.3\,m \ldots 100\,m. \textbf{Row 2:} Effect of  enforcing the soft epipolar line constraint. Note that the double image artifacts of the QR code in the top right are resolved by the soft epipolar constraint.
    }
\end{figure}

\textbf{Soft Epipolar Constraints:} Obtaining a subpixel-accurate calibration is difficult and costly. In addition, the smart glasses are typically  deformable \cite{Wang_2023_CVPR}, which may invalidate the calibration. To this end, we propose a soft epipolar constraint that allows for some inaccuracy in the calibration. Fig.~\ref{fig:epi} renders the epipolar lines for two pixels and shows the improvement after integrating the constraint. To implement it, we constrain the update steps in RAFT to lie within a tolerance of the epipolar lines. Furthermore, to reduce ambiguity even within a single RAFT iteration, we introduce a weighting that is inversely proportional to the distance to the epipolar lines and decreases the correlation scores of distant pixels.
    
\subsubsection{Reference-Based Super Resolution (RSR)}
\label{sec:gamma}
\begin{figure*}[tp]
\centering
\includegraphics[width=\linewidth]{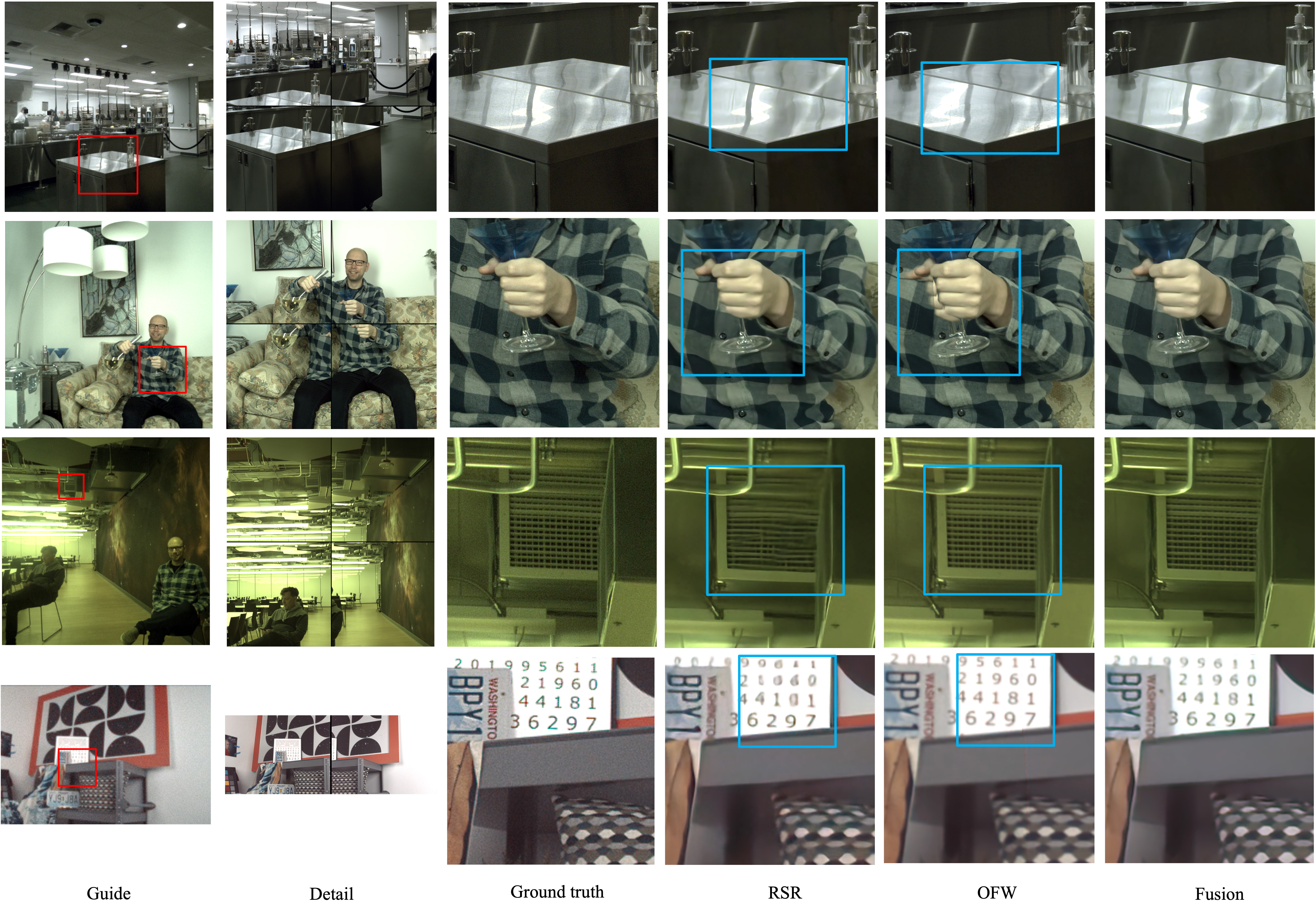}
\caption{\label{fig:gamma_vs_beta}
    Strengths and weaknesses of the optical flow warping-based (OFW) and reference-based super resolution (RSR) pipeline. \textcolor{red}{Red} highlighted regions are selected zoom-in regions. \textcolor{cyan}{Cyan} regions highlight the most different parts between the two pipelines. 
    \textbf{Column ``Detail''} only shows 4 out of 9 detail views that contain the most relevant content of the highlighted regions.
    \textbf{Row 1} shows a case where the RSR pipeline can keep the view-dependent effects such as specular highlights while the OFW pipeline fails.
    \textbf{Row 2} shows a case where the RSR pipeline succeeds in regions that are occluded in the detail, while the OFW pipeline fails due to different view directions from the detail images.
    \textbf{Rows 3 and 4} show cases where the OFW pipeline is able to restore high-frequency details such as thin structures and text.
    \textbf{Row 4} is also a real scene captured by a modified Aria head-worn device. It is clear that some numbers from the RSR pipeline are restored incorrectly.
    }
\end{figure*}

The optical flow warping-based pipeline produces the best results when the optical flow is correct but fails gracelessly otherwise and fundamentally cannot reconstruct areas that are occluded in the detail images due to the depth of the scene and the different viewpoints. We therefore introduce a second approach that leverages reference-based super resolution (RSR). This approach is very robust to incorrect correspondences, while trading-off the superior sharpness obtained by the OFW approach. In a final step, we fuse both results together. An illustration of the strengths and weaknesses of both approaches is shown by the qualitative results in Figure~\ref{fig:gamma_vs_beta}. Compare for example the hand holding the glass in the second row and the text in the third row.

We adopt the approach from \citet{jiang2021robust} referred to as C$^\mathrm{2}$-Matching. This approach aims to restore a high resolution image from a low-resolution image (the guide in our case) by transferring features from a so-called reference image (the details in our case). In the original work, the reference image can be an image of a different scene with similar content, which can be seen as providing a ``vocabulary'' to restore the current image. In our case, the reference is more similar as it is actually from the same scene. To transfer the features, nearest neighbor matches are computed in feature space that associate features from the low-resolution image to their closest counterparts in the reference. The features from the reference are then warped and jointly decoded with features from the low-resolution image. Note that the correspondences here are not spatially consistent and based on feature similarity alone. This means that a correspondence that is incorrect in terms of scene geometry will still associate a similar feature, and thus yield a relatively good restoration, which overall leads to high robustness to incorrect correspondences. On the other hand, the fact that exact correspondences are not available leads to ambiguity in the decoding---making it impossible to uniquely decode the right detail---and therefore the decoder will learn to produce an average of possible results, leading to blur in such cases. Alternatively, matching is inaccurate, which is especially problematic in regions that require accurate detail such as the panel with numbers in  Fig. \ref{fig:gamma_vs_beta}, last row.

\begin{figure}[tp]
\includegraphics[width=\linewidth]{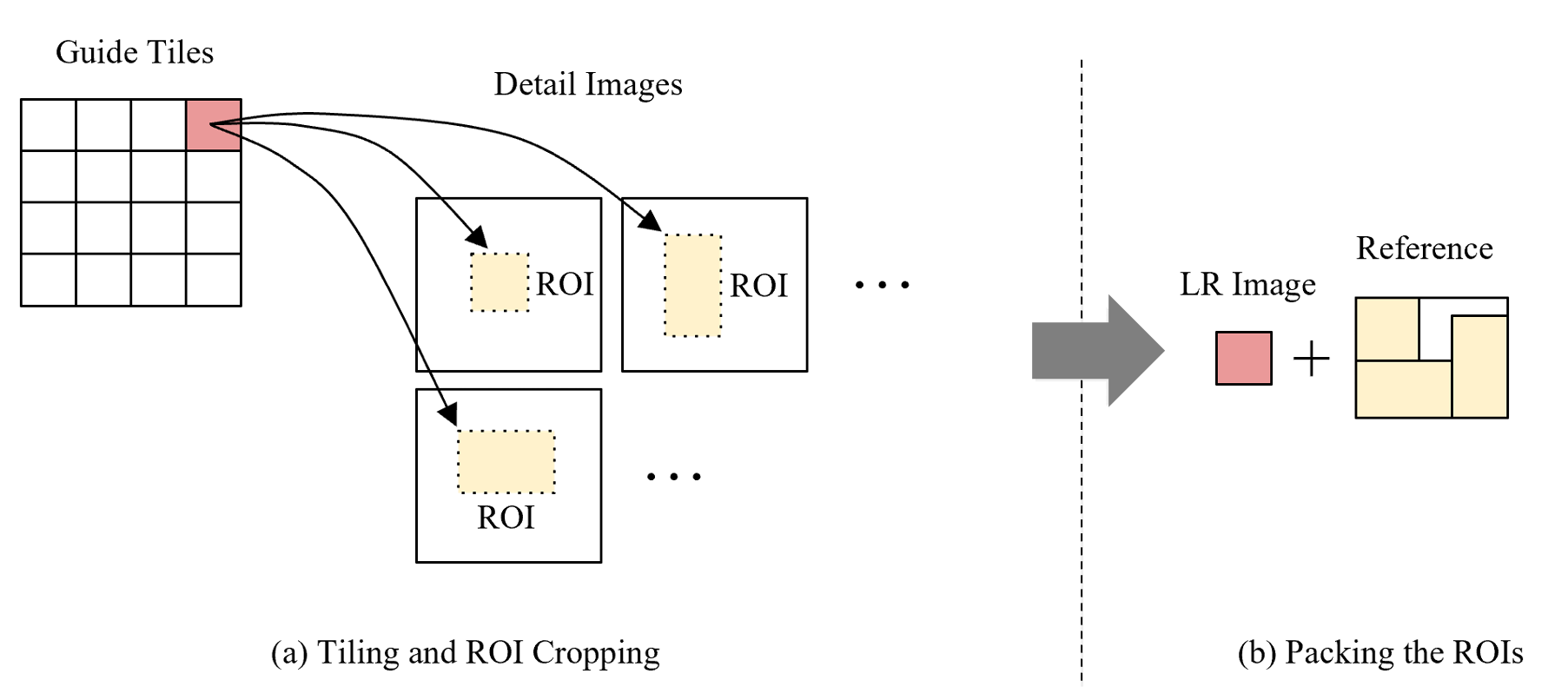}
\caption{\label{fig:gamma_inference}
    Overview of the ROI cropping and packing during the inference step of the reference-based super resolution approach. 
    }
\end{figure}

In order to deal with the memory requirements due to the large final image resolution, we first split the guide into tiles, as shown in Figure~\ref{fig:gamma_inference}(a). For each tile, we then use the same homography at 100\,m distance to determine in which regions in the detail images the tile may be visible (Figure~\ref{fig:gamma_inference}(a)) and refer to these regions as \emph{regions of interest} (ROIs). Subsequently, we pack the ROIs together to obtain a single reference image (Figure~\ref{fig:gamma_inference}(b)). We then feed the guide tile and the packed ROIs into C$^\mathrm{2}$-Matching and obtain a super-resolved target tile. We originally chose the tiles in such a way that they slightly overlap. In a final step, we assemble the tiles back together by using a soft blending between them.
We found that this works well in practice and avoids any boundary artifacts. It also serves as an additional regularization that reduces the number of false matches in the RSR pipeline.

\subsubsection{Fusion}
From the discussion in the previous sections and the results in Figure~\ref{fig:gamma_vs_beta}, we see that the OFW and RSR approaches complement each other. To obtain the benefits of both, we present a final pipeline stage that fuses both results together. To this end, we borrow the idea from Deep Burst Super Resolution~\cite{bhat2021deep}, which performs fusion of the warped feature maps from the different burst images. In our case, the setting is slightly different, as we have a guide image of lower resolution. For any pixel of the target image, we may have several results from the OFW approach from different detail images and one result from the RSR approach. An overview of our proposed architecture is shown in Figure~\ref{fig:fusion_arch}. The output of the fusion is then the final result of our pipeline.

\subsection{Depth Estimation}
\label{sec:depth}
\begin{figure}[tp]
\includegraphics[width=\linewidth]{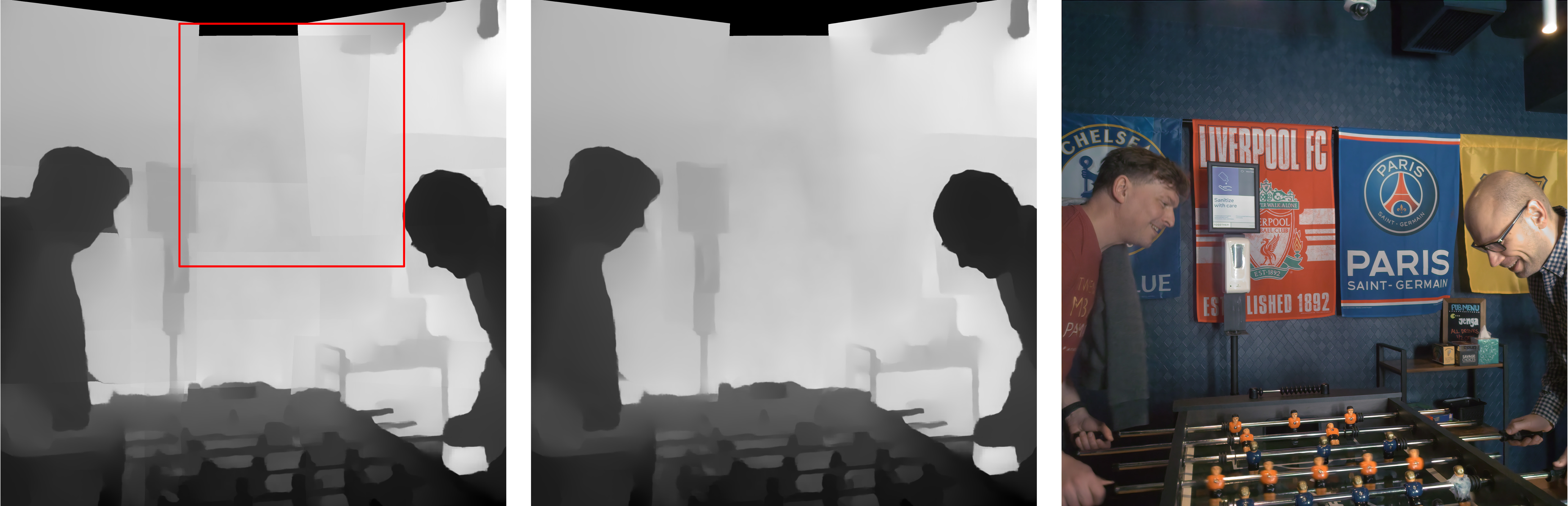} 
\caption{\label{fig:depth}
    Depth estimation based on OFW. \textbf{Left:} Na\"ive averaging of individual depth maps. \textbf{Middle:} Weighted blend with distance transform. \textbf{Right:} Reconstructed image. 
    The highlighted region illustrates the reduction of discontinuity on the flat wall compared to  na\"ively averaging individual depth maps. }
\end{figure}

Given the dense correspondences obtained in the OFW pipeline and the known camera calibration we can additionally reconstruct a depth map of the scene using a straightforward approach. Specifically, we determine the depth in the target view via triangulation from the OFW correspondences, yielding partial depth maps for each pair of guide and detail images. Since the individual depth maps are not fully consistent, we compute a weighted averaging with blending weights computed by applying a distance transform on the contour of each detail's valid mask. The weight decreases linearly with pixels furthest from the contour, having weight 1 while pixels at the contour have weight 0. Figure~\ref{fig:depth} shows an example of a reconstructed depth map. 

\subsection{SLAM and Online Calibration}
\label{sec:slam}

A common way to localize a smart glasses device in the environment is applying SLAM and VIO techniques on the image stream from two or more dedicated wide angle cameras \cite{engel2014lsd,mur2017orb,mourikis2007multi}. In addition, these techniques can also be used for online calibration, update each camera's state (such as their relative positions and intrinsic parameters) due to frame bending or hardware temperature change, which improves the pose accuracy further.
To be able to use VIO, we incorporated a Project Aria device \cite{Engel:2023:PAA} into one of our prototypes (shown in Fig. \ref{fig:teaser}), which provides inertial measurements and allows to compare the resulting accuracy against Project Aria's own VIO system.

\begin{figure}
\centering
\includegraphics[width=0.6\linewidth]{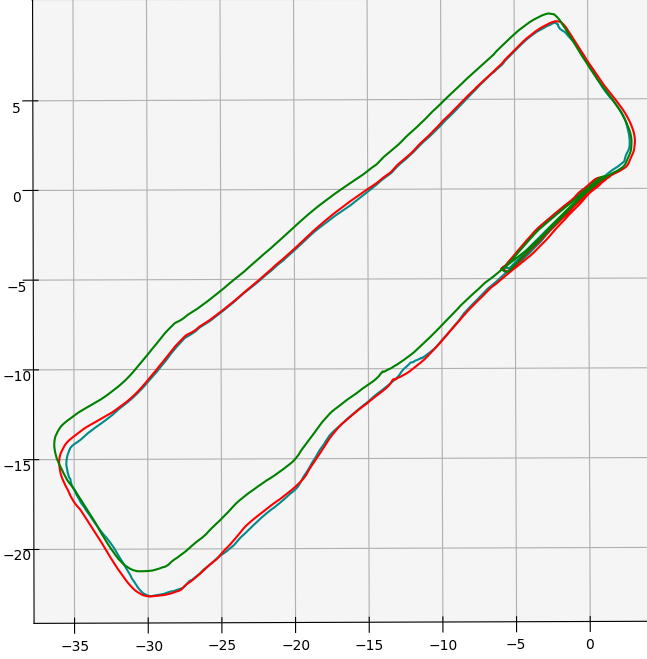} 
\caption{\label{fig:slam}
    SLAM results and comparison with the commercial state-of-the-art method. 
    \textcolor{cyan}{Cyan} trajectory is obtained from our modified VIO.
    \textcolor{red}{Red} trajectory is obtained after bundle adjustment.
    \textcolor{darkgreen}{Green} trajectory is from Project Aria's VIO.}
\end{figure}

In order to successfully run VIO based on the image stream from the detail and guide cameras plus inertial measurements from the Aria system, we needed to make several changes compared to a standard system.
First, since the images of the detail cameras needed to be downsampled from higher angular resolution in order to match the angular resolution of the  guide camera, the resulting resolution of the images was low.
However, by using a high  frame rate  ($>100Hz$) underlying the burst mode reconstruction, we were able to use Lucas-Kanade Tracking~\cite{lucas1981iterative} as the backbone feature point tracking method.
The fundamental matrix and RANSAC are further used to filter outlier feature points.
Second, we applied a global bundle adjustment to the result of this modified VIO to optimize the trajectory and improve the accuracy.
Fig.~\ref{fig:slam} shows comparisons among the trajectories obtained from our modified VIO before and after bundle adjustment and Project Aria's VIO.
The absolute trajectory error (ATE) between the result of our modified VIO and Aria's VIO is $0.582\,m$, which improves to $0.385\,m$ after applying bundle adjustment..

\subsection{Generalization to Video}

Our proposed image reconstruction pipeline can be adapted to generate video sequences. One approach is to capture synchronized frames from the guide and detail cameras, and process each set of frames independently to generate output images for a video sequence. However, this frame-by-frame approach requires all cameras to run at the same frame rate, which can be power- and compute-intensive.

A second approach introduces temporal sparsity by running the detail cameras at a reduced frame rate compared to the guide camera, and using the temporally-nearest detail frames as reference when reconstructing each guide frame. This achieves lower power and compute usage, but comes at the cost of temporal consistency, where dynamic scene content from the past or future is mapped incorrectly. One additional requirement of our method is epipolar guidance in both our OFW and RSR approaches, which however is invalidated when the guide and detail images come from different instants of time. For static parts of the scene, this problem could be fixed by re-establishing the geometric relationship using accurate SLAM trajectories.

\section{Evaluation}
\label{sec:results}

We evaluate our approach using synthetic as well as real scene data. Since  camera modules at the envisioned scale (in the following called \textit{tiny camera modules}) are not readily available, we build our experimental setups using larger cameras. In both cases -- synthetic and real scenes -- we carefully simulate the relevant properties of tiny camera modules, namely image noise and optical blur. Specifically, we created model of a detail camera module that integrates an off-the-shelf OmniVision OV01A10 image sensor with a lens with f-number 1.8, a focal length of 
1.925\,mm, and an entrance aperture diameter of a 1.1\,mm and a similar model for the guide camera, which we use as basis for  our simulations. 

\subsection{Experimental Setup}
\label{sec:results:prototypes}

We built two prototypes for our evaluation experiments with off-the-shelf cameras and lenses. Since target camera systems at the correct miniature scale are not readily available and would not allow for flexible experimentation, our approach is to build prototypes with small - but not miniaturized to scale - high-quality components. 
Two prototypes have been built to serve different goals. We built one desk-mounted prototype to capture static scenes, with cameras that achieve the best angular resolution (1\,arcmin for a limited depth of field) that can be provided by existing off-the-self small cameras, but is limited in frame rate and therefore not suitable for capturing dynamic scenes. To address the latter, we built a second head-mounted prototype that allows high-speed captures of video and can be used to evaluate burst mode and video reconstruction quality, while making compromises in angular resolution. 
Both prototypes have a full $3\times3$ (detail) $+\,1$ (guide) camera version that covers the desired FOV, but they are also flexible to make a lighter one such as a $2\times1$ (detail) $+\,1$ (guide) version for better wearing comfort, smaller data bandwidth request, shorter computation time and faster experiments.
More details about the specs of these two prototypes are provided in Appendix \ref{app:prototypes}.

The full resolution raw images captured from our prototypes' cameras are high-angular-resolution large-FOV images that do not match what we expect to see in the small form-factor cameras which we can install on glasses. 
Thus, we degrade all ten images (nine detail images + one guide image) to simulate the expected quality and resolution of small form factor cameras using lens and camera models. 
Using models of miniature cameras we obtained from simulation with Zemax\footnote{\url{https://www.zemax.com/}}, we are able to obtain key lens degradation parameters such as the point spread function (PSF). The captured images are then degraded to match the expected PSF of lens at the target scale. 
Additional noise is added in order to simulate target size cameras. Implementation details on how these degradation models are measured and applied to both synthetic and read-world test images are provided in Appendix \ref{app:degrade}.

\subsection{Training and Evaluation Details}

We trained the individual modules of the full pipeline separately. 
We retrained the deep demosaicking model following the deep demosaicking approach \cite{gharbi2016} on a subset of the Segment Anything dataset \cite{kirillov2023segment}.
For the OFW pipeline, we reused the pre-trained RAFT model \cite{teed2020raft}. For the RSR pipeline, we retrained all models involved in the pipeline with the Segment Anything dataset, except for the VGG networks that are used to extract features. During the training process, we augmented the training data further for our setting: a) using deep demosaicking and b) adding specular highlights. For the fusion pipeline, we trained it with a  synthetic dataset. We simulated capture rigs based on our existing prototypes, and synthesized the captured images using the simulated rig and random virtual scenes. These images also go through a pre-processing step to simulate the camera degradations. They are then fed into the already trained upstream pipelines (OFW and RSR pipelines). The output images from the two pipelines and the ground-truth images from simulation are used as input data for training.

For training the burst mode pipeline, we choose a dataset of high-resolution with significant amount of detail. To this end, we trained the burst mode pipeline on an image dataset captured by high-resolution cameras carried by drones flying over an urban city area. 
We first downsampled the drone images to greatly suppress the noise accumulated during the data capture process. Then, we take random crops and artificially added small translations and rotations to these cropped patches to synthesize the motion during the burst capture. Finally, we added degradation matching our camera sensors to the synthesized cropped patches to create the input burst images. We take the reference frame without degradation as the ground truth, and used L1 loss for the training. The downstream modules -- the data restoration module of the feature-based style transfer pipeline and the fusion pipeline -- are also retrained. The former was retrained with the same drone dataset, and the latter was retrained using simulated burst mode captured images from the synthetic dataset.

In evaluation experiments, we ran our test pipelines on a Lenovo ThinkStation P620 with one NVIDIA RTX A6000 GPU. The running time using the static image pipeline without burst mode is 120 seconds per image for the desk-mounted prototype setup and 22 seconds per image for the head-mounted prototype setup. The difference is  mainly by the different image resolutions. Running the same pipeline with burst mode for the head-mounted prototype setup requires 25 seconds to process a single image from a burst of eight images. Note that all of these are research implementations, and significant speedups could be achieved by optimizing them. 

\subsection{Comparison With Other Commercial Devices}

We compare the results of our novel approach with other devices in terms of image quality and  machine perception applications. 

\subsubsection{Photography}
\label{sec:comparison_photography}

\begin{figure*}
\centering
\includegraphics[width=\linewidth]{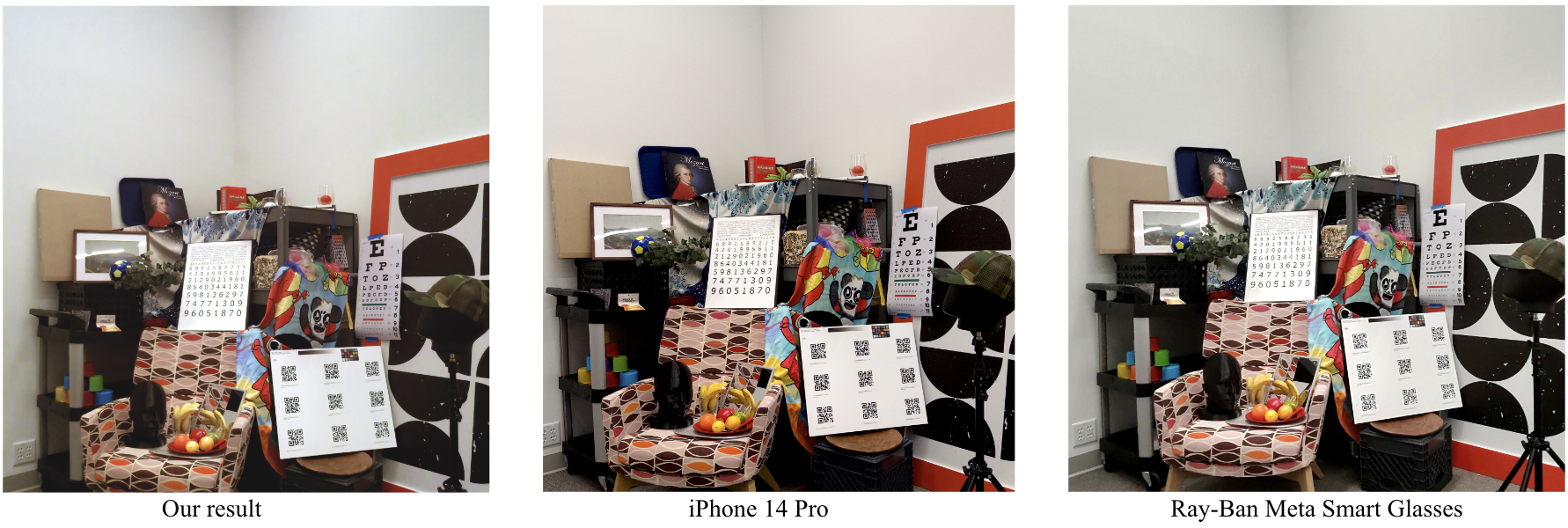}
\includegraphics[width=\linewidth]{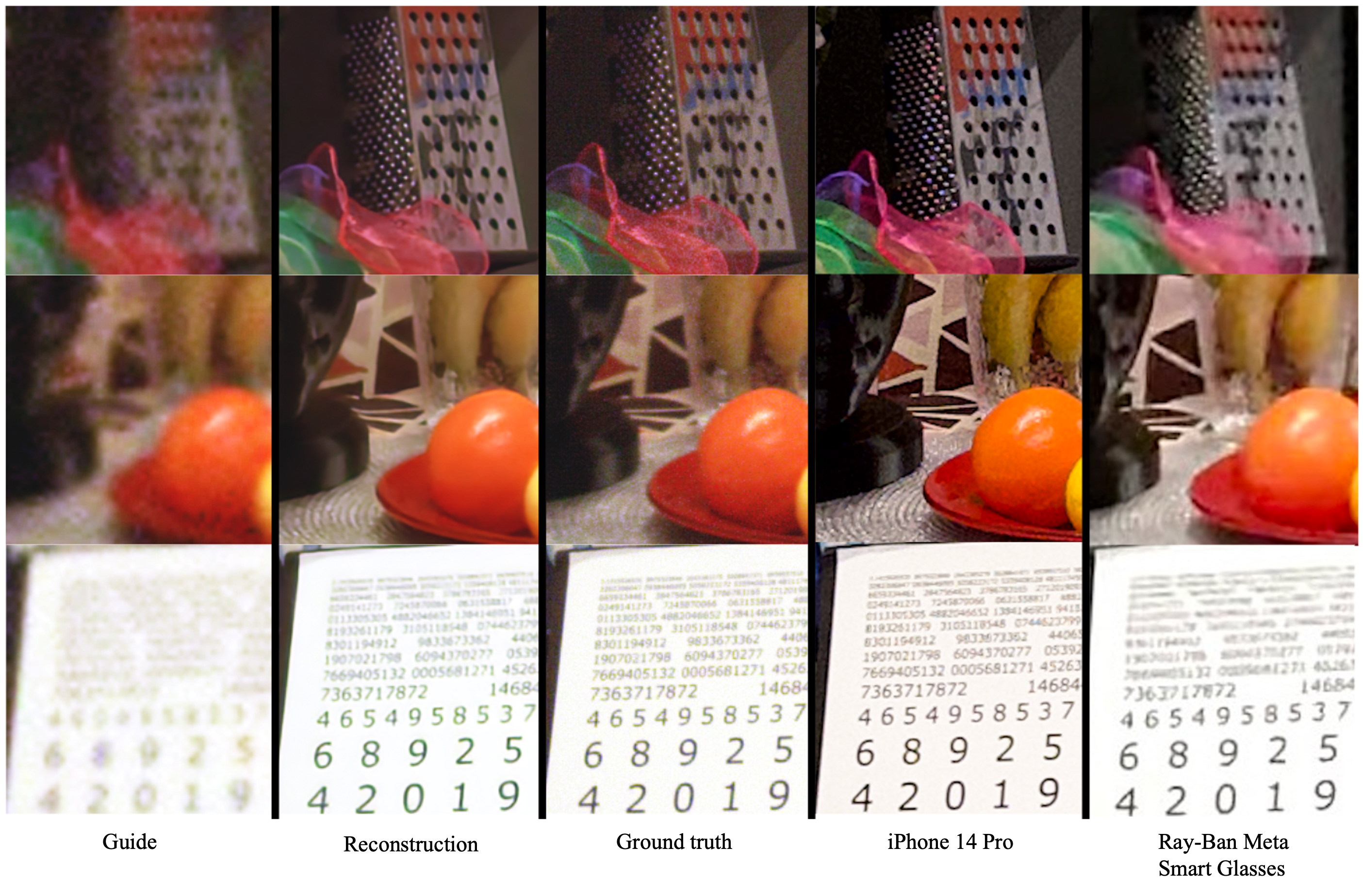}
\caption{\label{fig:fusion2}
    Comparing our pipeline and other state-of-the-art commercial devices. Top: full images. Bottom: selected detail crops. The image from the guide camera on the left resembles what can be achieved with a single camera for the large FOV with small form factor. Ground truth, iPhone 14 Pro and Ray-Ban Meta Smart Glasses all use significantly larger cameras. The reconstruction column shows the quality that can be achieved with our distributed imaging approach with a set of simulated ten tiny cameras (9 detail +  1 guide). As visible, our approach outperforms Ray-Ban Meta Smart glasses and comes close to the performance of the iPhone 14 Pro's main camera.  
    }
\end{figure*}

We compare our approach with two state-of-the-art commercial devices: the primary camera of the iPhone 14 Pro capturing at 12\,MP and the Ray-Ban Meta Smart Glasses. Examples of the achieved image quality are shown in Fig.~\ref{fig:fusion2}. 
Note that our prototype uses large cameras, and we obtain the guide and detail input images for our pipeline through simulation. The  ``ground truth'' is the image from the guide camera before degrading it with the settings of our simulated guide camera. It resembles what commercially available cameras with the required FOV can currently achieve (without image-enhancing ISPs as in the Ray-Ban Meta Smart Glasses and the iPhone). Thus, matching the ground truth would mean that we are able to undo the degradation that happened due to the small form factor. 

Ray-Ban Meta Smart Glasses is a single camera solution that is fully constrained by the glass form-factor, which is subject to all the limitations imposed described in Sec.~\ref{sec:limits}. As visible---albeit having significantly smaller cameras--our approach outperforms it and comes close to the iPhone, which is not glasses form factor but uses a substantially larger camera and advanced ISP.     

The quality can be most intuitively evaluated using the digits of varying sizes in the last row of the figure. Note that most of these digits are completely lost in the guide image. 
The iPhone 14 Pro image shows better readability than the Ray-Ban Meta smart glasses image. 
The quality of the reconstructed image from our pipeline is similar to that of the iPhone. The smallest digits are slightly blurrier in our results than from iPhone, but the mid-sized digits look smoother and are more easily recognizable. Overall, the results demonstrate that we can obtain high-quality images close to the quality of the iPhone, despite using several tiny cameras instead of a single large module. 

Note that the observations from this experiment does not violate the optics limits we derived in Sec.~\ref{sec:limits}. The detail cameras in the desk-mounted prototype used in this experiment are able to achieve approximately 1\,arcmin resolution,  similar to the iPhone 14 Pro at 12\,MP, because we accepted a shallow depth of field. Meanwhile the 2\,arcmin angular resolution was derived based on the assumption that the fixed-focus camera has a large depth of field covering from 40\,cm to infinity.

\subsubsection{Machine Perception Applications}

One important use case for AR glasses is to be able to read QR codes without needing a secondary device and without needing wearers to reposition their head to align on-board cameras with the QR code. Note that in a production AR machine perception system, QR code recognition could be performed simply on raw images captured from individual detail cameras rather than requiring the full super-resolution pipeline, but this form of evaluation lets us determine to what degree the originally captured scene from detail cameras is available in the final output.

\begin{figure}
    \centering
    \includegraphics[width=\linewidth]{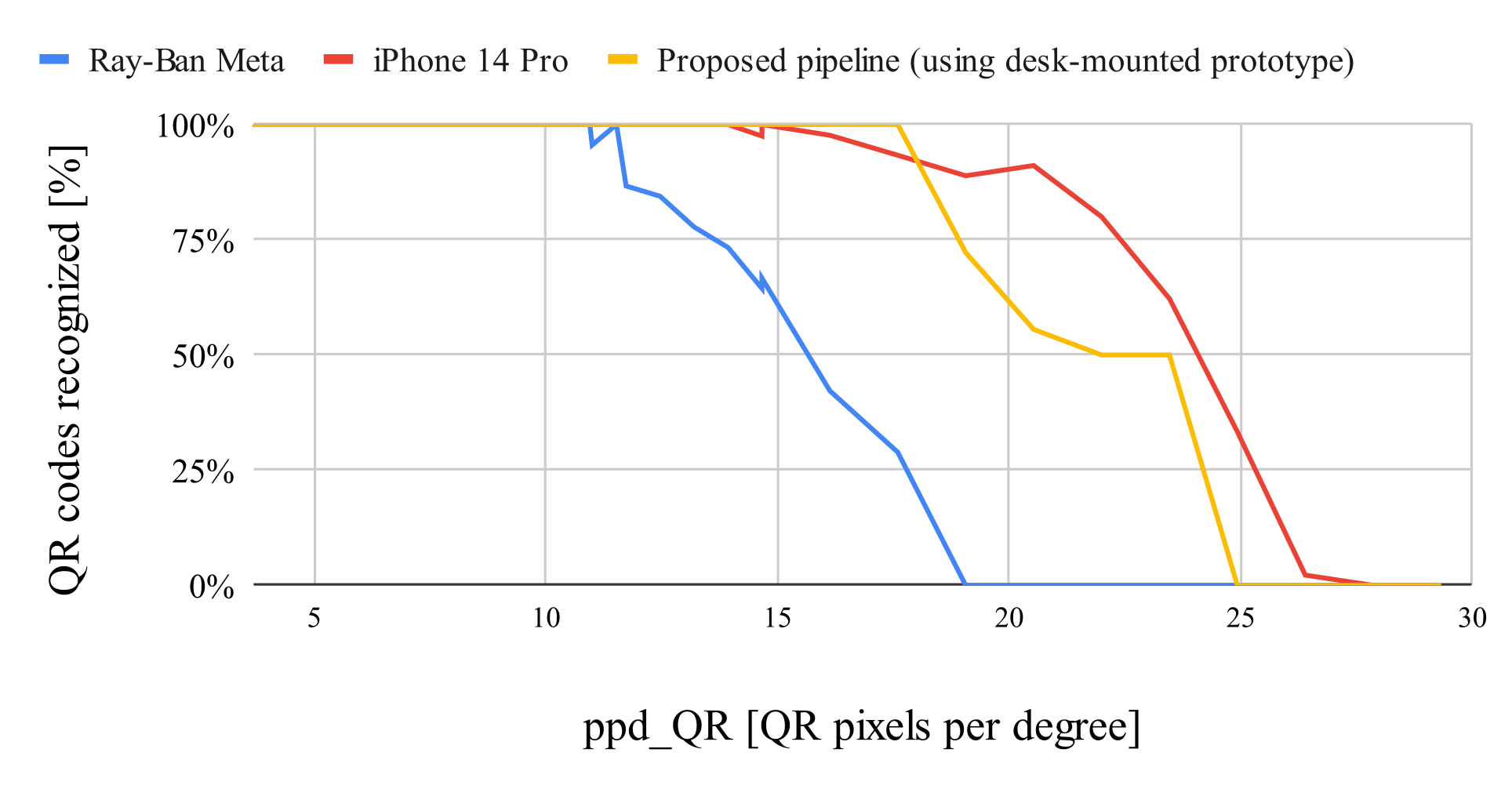}
    \caption{Percentage of QR codes successfully recognized over $ppd_{QR}$ on captured/reconstructed images from various devices (Ray-Ban Meta glasses, iPhone 14 Pro, and the results of our reconstruction pipeline on data captured from our desk-mounted prototype).
    As visible, our prototype significantly outperforms Ray-Ban Meta and comes close to iPhones. 
    }
    \label{fig:qr_evaluation}
\end{figure}

QR codes vary in printed size and in complexity, so we unify this variance along with viewing distance by defining a measure of ``QR pixels per viewing degree at a certain distance''. Given a QR code of physical width $w$ and pixel count $p\times p$, observed at a distance $d$, the QR ``pixels'' per viewing degree $ppd_{QR}$ is:
\[
ppd_{QR} = \frac{p}{2 \arctan(\frac{w}{2d})} \,\mathrm{.}
\]
As $ppd_{QR}$ increases beyond some threshold (with distance from the camera, with code complexity, or with decreased physical size), the ability of cameras to reliably resolve QR codes decreases. We again captured images of several posters with QR codes at 1-2\,m distance.
QR code recognition was performed on each image using an Android smartphone app and pointing the onboard camera at the magnified image displayed on a monitor\footnote{While QR recognition libraries such as zbar and zxing are available, we found such tools to introduce an unacceptable number of false-negatives in QR recognition that did not reflect the experience of typical smartphone usage.}. We recorded the percentage of visible QR codes recognized by this method. Fig.~\ref{fig:qr_evaluation} plots the relation between QR code recognition percentage and $ppd_{QR}$ for each imaging method.

The results show that QR recognition degrades rapidly on the Ray-Ban Meta Smart Glasses ($ppd_{QR} = [10,20]$), while photos from state-of-the-art smartphones and from our super-resolution pipeline have similar performance, degrading around $ppd_{QR} = [15,25]$. This demonstrates our ability to faithfully represent real-world scene information, similar to modern smartphones.

\subsection{Evaluation for Static Image Reconstruction}
\label{sec:results:static_pipeline}

We evaluate our image reconstruction pipeline on both synthetic and real data. To generate the test synthetic data, we used the Aria synthetic environments dataset \cite{ASE}. We simulated images captured from a synthetic rig from 100 random scenes.  
We also collected real-world images from 107 scenes using our desk-mounted prototype to test our reconstruction pipeline for static images. 
As in Section~\ref{sec:comparison_photography}, we use the high-resolution image from the guide-view camera as the ground-truth image for reference, before degrading it to a low-resolution image with our simulation. 

\begin{figure}
\centering
\includegraphics[width=\linewidth]{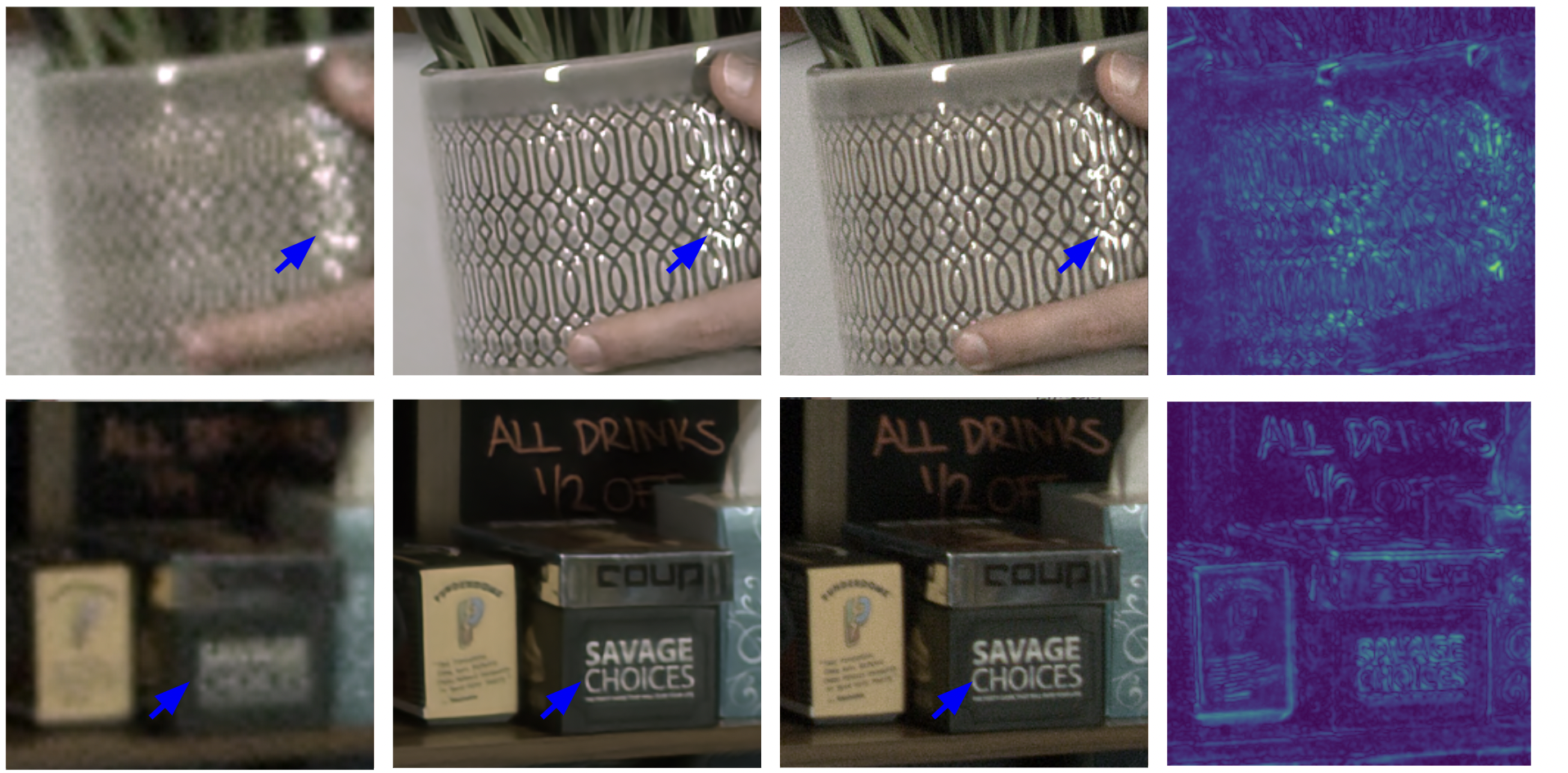}
\includegraphics[width=\linewidth]{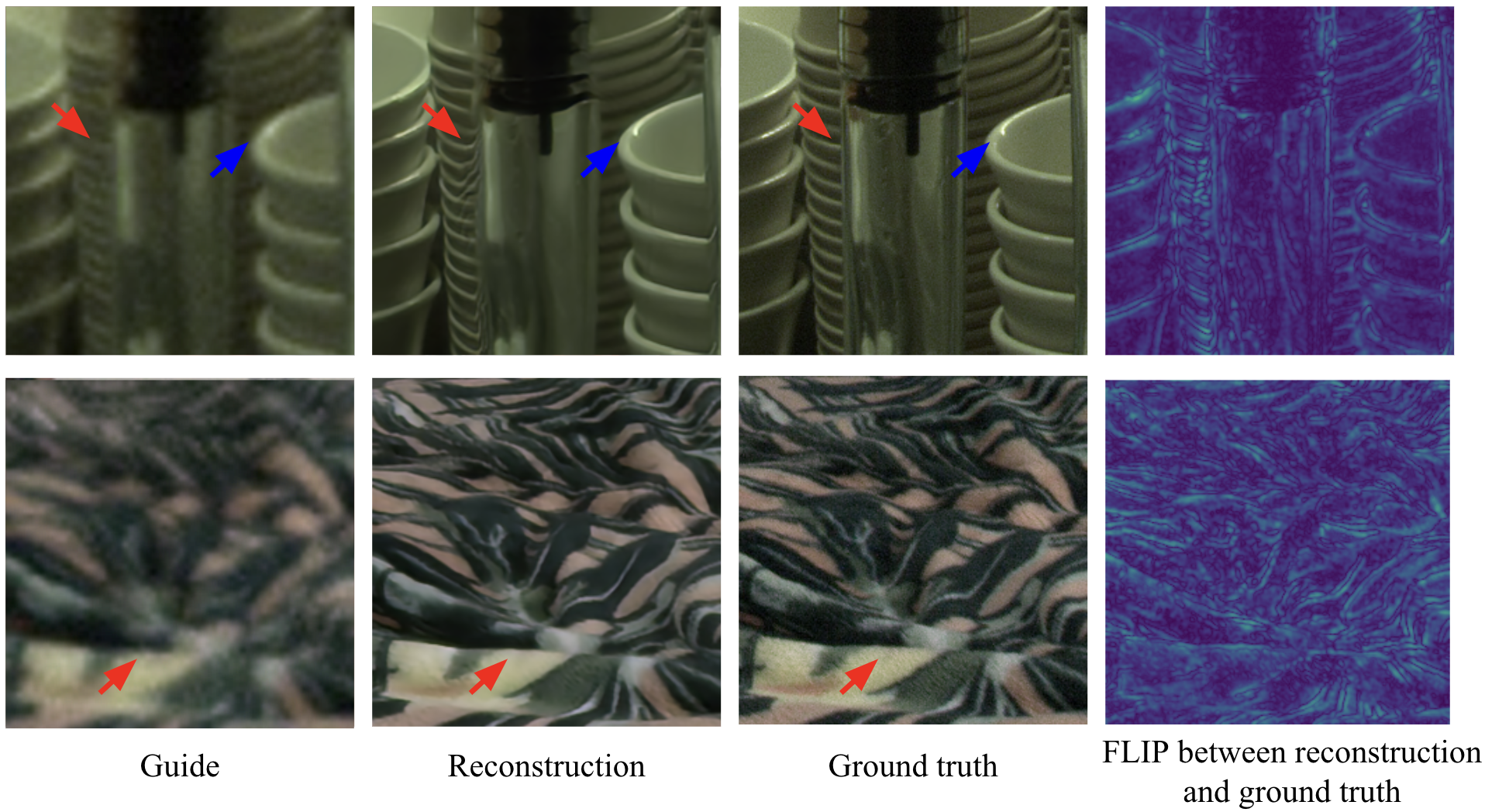}
\caption{\label{fig:fusion1}
    Examples demonstrating some success and failure cases of our full pipeline. From left to right: input guide image, reconstructed image, ground truth and the FLIP metric between reconstruction and ground truth, with a scale from 0 to 0.8. Blue arrows point to successful recovery of the details, while red arrows point to challenging cases in which our method is suboptimal due to occlusions (3rd row) or texture that is considered as noise (4th row).
}
\end{figure}

We inspected the results after the OFW, RSR and fusion stages of our pipeline and evaluated the reconstruction with standard image quality metrics: Table \ref{tab:metrics} summarizes the results based on FLIP~\cite{Andersson:2020:FDE}, peak signal-to-noise ratio (PSNR), and structural similarity index (SSIM)~\cite{Wang:2003:IQA}. The metrics are calculated between the reconstructed image from each pipeline stage and the ground-truth image.
For each dataset, the OFW pipeline in general has the worst metrics because of its inability to handle occlusions and the direct pixel-level interpolation involved in the warping step. However, visually it performs significantly better in regions with detail, as is for example visible in the fourth row of Fig.~\ref{fig:gamma_vs_beta}. The fusion yields a general improvement over both, OFW and RSR. 
Please note that the metrics for the real-world dataset are affected by the fact that the reference images come from a real camera and have noise themselves, and thus only provide an orientation. We recommend assessing the qualitative results from Figures~\ref{fig:fusion2} and~\ref{fig:fusion1} instead.

\begin{table}[]
    \centering
    \begin{tabular}{|c|c|c|c|c|}
    \hline
         Dataset & Sub-pipline & FLIP $\downarrow$ & PSNR $\uparrow$ & SSIM $\uparrow$ \\ \hline
         \multirow{3}{*}{Synthetic}& OFW & 0.101 & 23.74 & 0.884 \\ \cline{2-5}
         ~ & RSR & \textbf{0.058} & 37.37 & 0.917 \\ \cline{2-5}
         ~ & Fusion & 0.060 & \textbf{37.42} & \textbf{0.925} \\ \hline
         \multirow{3}{*}{Real-world}& OFW & 0.167 & 22.59 & 0.761\\\cline{2-5}
         ~ & RSR & \textbf{0.074} & \textbf{33.31} & 0.802\\\cline{2-5}
         ~ & Fusion & 0.082 & 33.17 & \textbf{0.804} \\ \hline
    \end{tabular}
    \caption{
    Quantitative evaluation of the reconstruction results for each stage of the pipeline. The RSR approach generally provides better numbers than the OFW one, although visual details from OFW are often better (see figure~\ref{fig:gamma_vs_beta}). The fusion qualitatively preserves the details and leads to the overall best metrics.     
    } 
    \label{tab:metrics}
\end{table}

We provide some additional qualitative examples with success and failure cases of our full pipeline in Fig.~\ref{fig:fusion1}. Note that it is able to recover details such as text and fine periodic structures that are greatly contaminated by degradation in the guide image. The pipeline is also capable of keeping the specular highlights at the correct locations, despite the fact that these highlights are view-dependent. As expected, it struggles in occluded regions, especially when the occluded regions have high-frequency details, and may also smooth structure out which is close  to noise, such as the furry blanket texture in the last row.

\subsection{Evaluation for Video Reconstruction}

\subsubsection{Frame-by-Frame Reconstruction Without Burst Mode}

\begin{figure}
\centering
\includegraphics[width=0.48\linewidth]{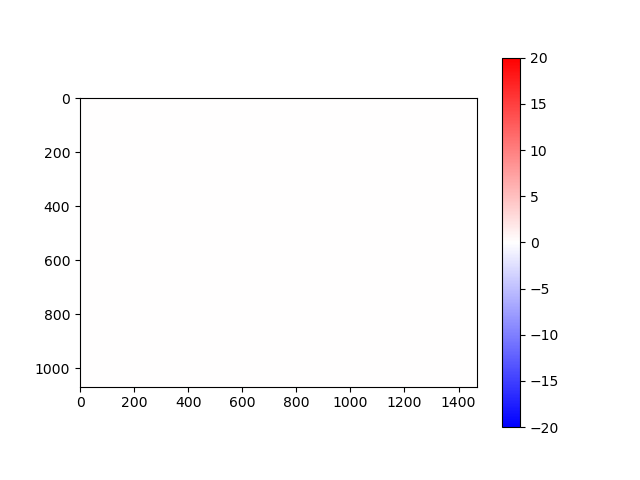}
\includegraphics[width=0.48\linewidth]{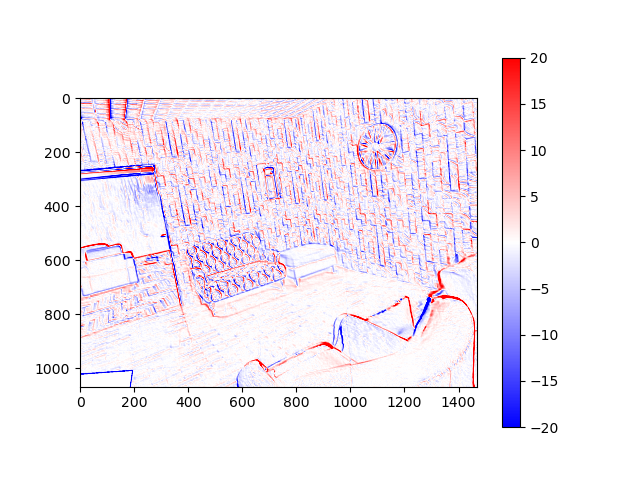}
\caption{\label{fig:temporal_consistency}
    Frame-by-frame processing without burst mode: temporally varying noise in the guide image 
    causes temporal inconsistencies in the reconstruction. We show the results of the OFW approach. Left: difference from two neighboring reconstructed frames for a static scene with noise-free input data. Right: difference from two neighboring reconstructed frames for a static scene with noise-contaminated input data.}
\end{figure}

With a frame-by-frame processing technique without burst mode, the image quality of individual frames would be the same as discussed in Section~\ref{sec:results:static_pipeline}. 
However, for videos the temporal consistency is of additional importance. Due to the temporally varying noise in the guide image, we observe some jitter even when observing a static scene from a fixed camera as illustrated in Figure~\ref{fig:temporal_consistency}. 

\subsubsection{Frame-by-Frame Reconstruction With Burst Mode}

\begin{figure}
\centering
\includegraphics[width=0.48\linewidth]{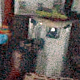}
\includegraphics[width=0.48\linewidth]{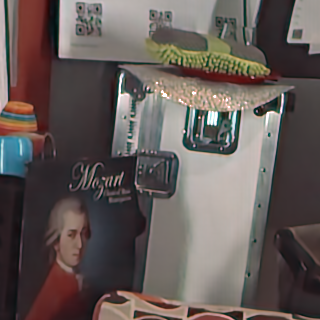}
\caption{\label{fig:burst_full_pipeline}
    Zoomed-in samples from the reconstruction of video with burst mode.
    Left: a single guide image from the burst. Right: full pipeline results.}
\end{figure}

Next, we evaluate our pipeline with burst mode on dynamic datasets captured using the head-mounted prototype. The exposure time for each frame was restricted to 1\,ms, and the gain value was set to the largest value to ensure a maximum use of dynamic range, given the short exposure time. 
As shown in Fig.~\ref{fig:burst_full_pipeline}, the burst mode algorithm performs well in avoiding motion blur and removing the noise accompanying the short exposure time. The textures are recovered properly and the semantically-meaningful text regions become readable again.

\subsubsection{Video Reconstruction With Sparse Inputs}

\begin{figure}
\centering
\includegraphics[width=0.98\linewidth]{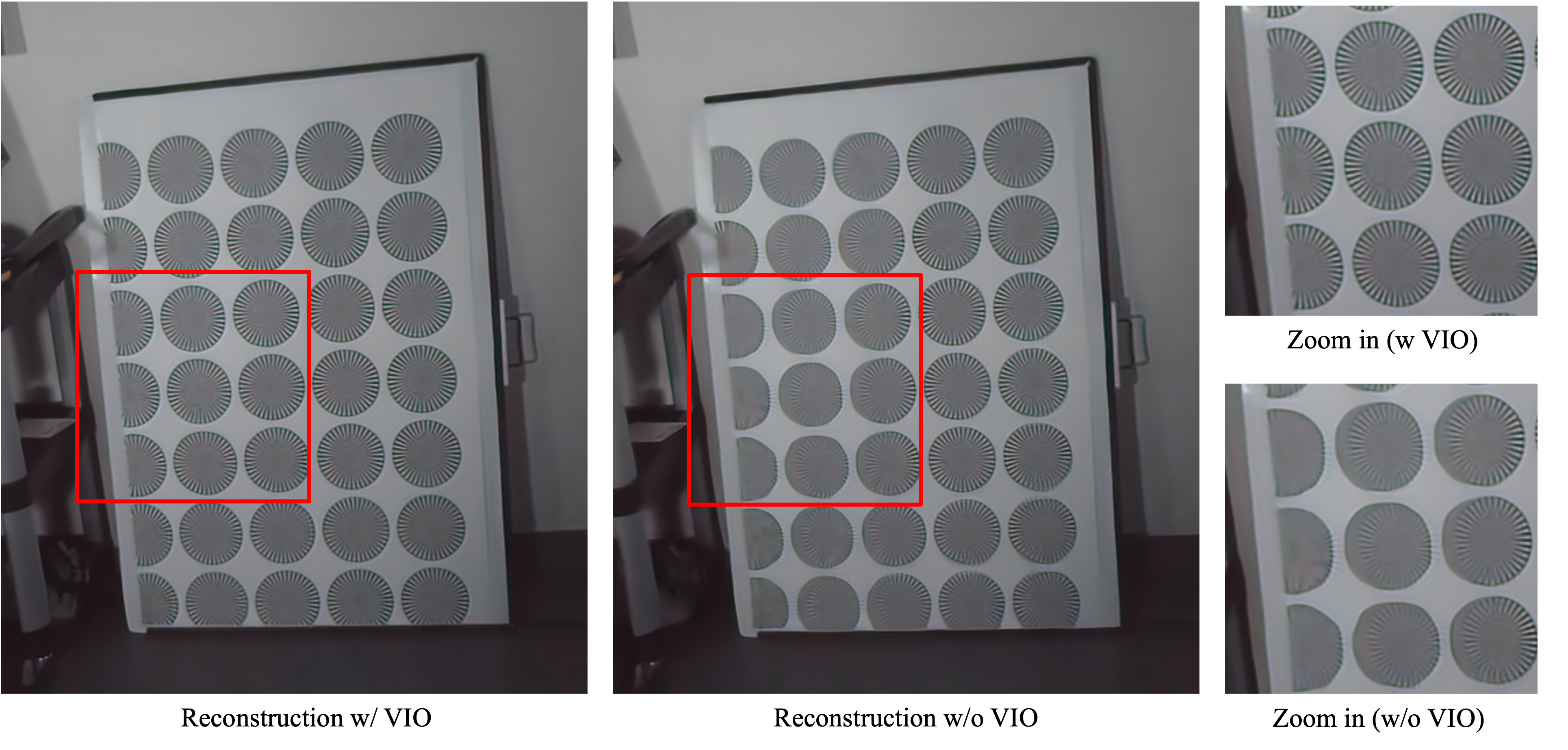}
\caption{\label{fig:SLAM_video}
    Sample frames from video reconstruction results with (\textbf{Column 1}) and without (\textbf{Column 2}) VIO trajectories. 
    \textbf{Column 3} shows the zoom-in view of \textcolor{red}{red} highlighted, illustrating the differences in reconstruction quality of the star patterns.}
\end{figure}

To evaluate the pipeline's ability to cope with sparse input data, we collected video sequences of a static scene using the moving head-mounted prototype with the non-burst mode option and reduced the frame rate of the  detail cameras to one seventh the frame rate of the guide camera. We used a bright light source that allows adequate illumination even for a short exposure time to avoid motion blur. We also recorded inertial data in order to  support VIO trajectory generation. 
Fig.~\ref{fig:SLAM_video} shows a comparison between reconstructed video frames with and without VIO. Without VIO, the epipolar constraint is invalid, so the pipeline is not able to find the correct matching region; thus, the star pattern becomes a blurred blob. The problem is much alleviated with VIO trajectory correcting the epipolar constraints, leading to successful reconstruction of the star pattern. Note that this assumes a static scene. Once the scene content is changing, the epipolar constraint will no longer be valid regardless of how accurately we track the camera poses.

\section{Discussion}
\label{sec:discussion}

\paragraph{Compute and power} In this paper, we explicitly excluded constraints on (on-device) computation and power use. Minimizing both is essential in any practical implementation, since they determine whether an device is all-day usable and impact its weight and form factor (in particular via the size of the required battery). Besides general optimization strategies applied in any hardware development, it is possible to reduce the power consumption by replacing the processing pipeline with approaches that are better suited for on-device implementation. An alternative approach is to increase the sparsity, e.g., by only capturing detail images when needed while most of the time relying on the lower resolution guide images, potentially offloading expensive computation off-device. All of these strategies need to keep the full overall system in mind and trade off the compute and power costs incurred.

\paragraph{Image quality and reliability} A fundamental challenge for the distributed imaging method is the lack of original data in occluded regions and the potential for artifacts introduced in our pipeline, e.g., if correspondence is not reliably estimated and we need to rely on the GAN-based reconstruction provided by the RSR pipeline. This opens up an opportunity to extend our proposed approach to include more recent generative AI-based reconstruction techniques that provide photorealistic results  \cite{rombach2022high}. Still, all of these methods are able to introduce hallucinations, which have the potential to fundamentally alter the image content. Our proposed distributed imaging approach provides one key advantage: All input images we capture are true images from miniaturized, standard photographic cameras. If needed, we can  rely on these for an accurate but lower quality representation of the scene.

\paragraph{Privacy}  As pointed out by Engel et al. \cite{Engel:2023:PAA}, privacy is an important consideration for the development of any smart glasses device. In their research, they follow strict responsible innovation principles and set clear guidelines for the use of smart glasses devices. Any implementation of our proposed approach will need to follow similar standards to protect users and bystanders alike.

\section{Conclusion}
\label{sec:conclusion}

Imaging for all-day wearable smart glasses poses a unique set of challenges, both in terms of the camera hardware design and the algorithms required to generate high-quality imagery for both photography and AI applications. Considering such a system from a holistic view is essential, since it exposes the overall complexity of the problem and opportunities that may otherwise be overlooked. 
In this work, we proposed a solution from the perspective of a full end-to-end system ranging from physics to software and have shown for the first time that an alternative solution with a distributed imaging system can provide quality close to modern mobile phones, while maintaining the necessary form-factor to seamlessly integrate into smart glasses. 
Although many challenges remain, we see it as a significant step forward 
and hope our work will spur a new line of research in the future.

\section*{Acknowledgements}
The authors would like to thank Michael Proulx for his input on the properties of the human visual system.

\bibliographystyle{ACM-Reference-Format}
\bibliography{paper}


\begin{thebibliography}{65}


\ifx \showCODEN    \undefined \def \showCODEN     #1{\unskip}     \fi
\ifx \showDOI      \undefined \def \showDOI       #1{#1}\fi
\ifx \showISBNx    \undefined \def \showISBNx     #1{\unskip}     \fi
\ifx \showISBNxiii \undefined \def \showISBNxiii  #1{\unskip}     \fi
\ifx \showISSN     \undefined \def \showISSN      #1{\unskip}     \fi
\ifx \showLCCN     \undefined \def \showLCCN      #1{\unskip}     \fi
\ifx \shownote     \undefined \def \shownote      #1{#1}          \fi
\ifx \showarticletitle \undefined \def \showarticletitle #1{#1}   \fi
\ifx \showURL      \undefined \def \showURL       {\relax}        \fi
\providecommand\bibfield[2]{#2}
\providecommand\bibinfo[2]{#2}
\providecommand\natexlab[1]{#1}
\providecommand\showeprint[2][]{arXiv:#2}

\bibitem[Alakarhu(2007)]%
        {Alakarhu:2007:ISI}
\bibfield{author}{\bibinfo{person}{Juha Alakarhu}.}
  \bibinfo{year}{2007}\natexlab{}.
\newblock \showarticletitle{Image Sensors and Image Quality in Mobile Phones}.
  In \bibinfo{booktitle}{\emph{2007 International Image Sensor Workshop}}.
\newblock


\bibitem[Andersson et~al\mbox{.}(2020)]%
        {Andersson:2020:FDE}
\bibfield{author}{\bibinfo{person}{Pontus Andersson}, \bibinfo{person}{Jim
  Nilsson}, \bibinfo{person}{Tomas Akenine-M\"{o}ller}, \bibinfo{person}{Magnus
  Oskarsson}, \bibinfo{person}{Kalle \r{A}str\"{o}m}, {and}
  \bibinfo{person}{Mark~D. Fairchild}.} \bibinfo{year}{2020}\natexlab{}.
\newblock \showarticletitle{FLIP: A Difference Evaluator for Alternating
  Images}.
\newblock \bibinfo{journal}{\emph{Proc. ACM Comput. Graph. Interact. Tech.}}
  \bibinfo{volume}{3}, \bibinfo{number}{2}, Article \bibinfo{articleno}{15}
  (\bibinfo{date}{aug} \bibinfo{year}{2020}), \bibinfo{numpages}{23}~pages.
\newblock
\urldef\tempurl%
\url{https://doi.org/10.1145/3406183}
\showDOI{\tempurl}


\bibitem[Bekerman et~al\mbox{.}(2014)]%
        {Bekerman2014VariationsIE}
\bibfield{author}{\bibinfo{person}{Inessa Bekerman}, \bibinfo{person}{Paul
  Gottlieb}, {and} \bibinfo{person}{Michael Vaiman}.}
  \bibinfo{year}{2014}\natexlab{}.
\newblock \showarticletitle{Variations in eyeball diameters of the healthy
  adults.}
\newblock \bibinfo{journal}{\emph{Journal of Ophthalmology}}
  (\bibinfo{year}{2014}).
\newblock
\urldef\tempurl%
\url{https://doi.org/10.1155/2014/503645}
\showDOI{\tempurl}


\bibitem[Bhat et~al\mbox{.}(2021)]%
        {bhat2021deep}
\bibfield{author}{\bibinfo{person}{Goutam Bhat}, \bibinfo{person}{Martin
  Danelljan}, \bibinfo{person}{Luc Van~Gool}, {and} \bibinfo{person}{Radu
  Timofte}.} \bibinfo{year}{2021}\natexlab{}.
\newblock \showarticletitle{Deep burst super-resolution}. In
  \bibinfo{booktitle}{\emph{Proceedings of the IEEE/CVF Conference on Computer
  Vision and Pattern Recognition}}. \bibinfo{pages}{9209--9218}.
\newblock


\bibitem[Bipat et~al\mbox{.}(2019)]%
        {Bipat:2019:AUC}
\bibfield{author}{\bibinfo{person}{Taryn Bipat},
  \bibinfo{person}{Maarten~Willem Bos}, \bibinfo{person}{Rajan Vaish}, {and}
  \bibinfo{person}{Andr\'{e}s Monroy-Hern\'{a}ndez}.}
  \bibinfo{year}{2019}\natexlab{}.
\newblock \showarticletitle{Analyzing the Use of Camera Glasses in the Wild}.
  In \bibinfo{booktitle}{\emph{Proceedings of the 2019 CHI Conference on Human
  Factors in Computing Systems}} (Glasgow, Scotland Uk)
  \emph{(\bibinfo{series}{CHI '19})}. \bibinfo{publisher}{Association for
  Computing Machinery}, \bibinfo{address}{New York, NY, USA},
  \bibinfo{pages}{1–8}.
\newblock
\showISBNx{9781450359702}
\urldef\tempurl%
\url{https://doi.org/10.1145/3290605.3300651}
\showDOI{\tempurl}


\bibitem[Brooks et~al\mbox{.}(2018)]%
        {Brooks:2018:UIL}
\bibfield{author}{\bibinfo{person}{Tim Brooks}, \bibinfo{person}{Ben
  Mildenhall}, \bibinfo{person}{Tianfan Xue}, \bibinfo{person}{Jiawen Chen},
  \bibinfo{person}{Dillon Sharlet}, {and} \bibinfo{person}{Jonathan~T.
  Barron}.} \bibinfo{year}{2018}\natexlab{}.
\newblock \showarticletitle{Unprocessing Images for Learned Raw Denoising}.
\newblock \bibinfo{journal}{\emph{CoRR}}  \bibinfo{volume}{abs/1811.11127}
  (\bibinfo{year}{2018}).
\newblock
\showeprint[arXiv]{1811.11127}
\urldef\tempurl%
\url{http://arxiv.org/abs/1811.11127}
\showURL{%
\tempurl}


\bibitem[Brown and Lowe(2007)]%
        {brown2007automatic}
\bibfield{author}{\bibinfo{person}{Matthew Brown} {and}
  \bibinfo{person}{David~G Lowe}.} \bibinfo{year}{2007}\natexlab{}.
\newblock \showarticletitle{Automatic panoramic image stitching using invariant
  features}.
\newblock \bibinfo{journal}{\emph{International journal of computer vision}}
  \bibinfo{volume}{74} (\bibinfo{year}{2007}), \bibinfo{pages}{59--73}.
\newblock


\bibitem[Brown et~al\mbox{.}(2003)]%
        {brown2003recognising}
\bibfield{author}{\bibinfo{person}{Matthew Brown}, \bibinfo{person}{David~G
  Lowe}, {et~al\mbox{.}}} \bibinfo{year}{2003}\natexlab{}.
\newblock \showarticletitle{Recognising panoramas}. In
  \bibinfo{booktitle}{\emph{ICCV}}, Vol.~\bibinfo{volume}{3}.
  \bibinfo{pages}{1218}.
\newblock


\bibitem[Cao et~al\mbox{.}(2022)]%
        {cao2022reference}
\bibfield{author}{\bibinfo{person}{Jiezhang Cao}, \bibinfo{person}{Jingyun
  Liang}, \bibinfo{person}{Kai Zhang}, \bibinfo{person}{Yawei Li},
  \bibinfo{person}{Yulun Zhang}, \bibinfo{person}{Wenguan Wang}, {and}
  \bibinfo{person}{Luc~Van Gool}.} \bibinfo{year}{2022}\natexlab{}.
\newblock \showarticletitle{Reference-based image super-resolution with
  deformable attention transformer}. In \bibinfo{booktitle}{\emph{ECCV}}.
  Springer, \bibinfo{pages}{325--342}.
\newblock


\bibitem[Chakravarthula et~al\mbox{.}(2023)]%
        {Chakravarthula:2023:TON}
\bibfield{author}{\bibinfo{person}{Praneeth Chakravarthula},
  \bibinfo{person}{Jipeng Sun}, \bibinfo{person}{Xiao Li},
  \bibinfo{person}{Chenyang Lei}, \bibinfo{person}{Gene Chou},
  \bibinfo{person}{Mario Bijelic}, \bibinfo{person}{Johannes Froesch},
  \bibinfo{person}{Arka Majumdar}, {and} \bibinfo{person}{Felix Heide}.}
  \bibinfo{year}{2023}\natexlab{}.
\newblock \showarticletitle{Thin On-Sensor Nanophotonic Array Cameras}.
\newblock \bibinfo{journal}{\emph{ACM Trans. Graph.}} \bibinfo{volume}{42},
  \bibinfo{number}{6}, Article \bibinfo{articleno}{249} (\bibinfo{date}{Dec.}
  \bibinfo{year}{2023}), \bibinfo{numpages}{18}~pages.
\newblock
\showISSN{0730-0301}
\urldef\tempurl%
\url{https://doi.org/10.1145/3618398}
\showDOI{\tempurl}


\bibitem[Debevec and Malik(1997)]%
        {debevec1997recovering}
\bibfield{author}{\bibinfo{person}{Paul~E Debevec} {and}
  \bibinfo{person}{Jitendra Malik}.} \bibinfo{year}{1997}\natexlab{}.
\newblock \showarticletitle{Recovering high dynamic range radiance maps from
  photographs}. In \bibinfo{booktitle}{\emph{Proceedings of the 24th annual
  conference on Computer graphics and interactive techniques}}.
  \bibinfo{pages}{369--378}.
\newblock


\bibitem[Engel et~al\mbox{.}(2014)]%
        {engel2014lsd}
\bibfield{author}{\bibinfo{person}{Jakob Engel}, \bibinfo{person}{Thomas
  Sch{\"o}ps}, {and} \bibinfo{person}{Daniel Cremers}.}
  \bibinfo{year}{2014}\natexlab{}.
\newblock \showarticletitle{{LSD-SLAM}: Large-scale direct monocular {SLAM}}.
  In \bibinfo{booktitle}{\emph{Computer Vision--ECCV 2014: 13th European
  Conference, Zurich, Switzerland, September 6-12, 2014, Proceedings, Part II
  13}}. Springer, \bibinfo{pages}{834--849}.
\newblock


\bibitem[Engel et~al\mbox{.}(2023)]%
        {Engel:2023:PAA}
\bibfield{author}{\bibinfo{person}{Jakob Engel}, \bibinfo{person}{Kiran
  Somasundaram}, \bibinfo{person}{Michael Goesele}, \bibinfo{person}{Albert
  Sun}, \bibinfo{person}{Alexander Gamino}, \bibinfo{person}{Andrew Turner},
  \bibinfo{person}{Arjang Talattof}, \bibinfo{person}{Arnie Yuan},
  \bibinfo{person}{Bilal Souti}, \bibinfo{person}{Brighid Meredith},
  \bibinfo{person}{Cheng Peng}, \bibinfo{person}{Chris Sweeney},
  \bibinfo{person}{Cole Wilson}, \bibinfo{person}{Dan Barnes},
  \bibinfo{person}{Daniel DeTone}, \bibinfo{person}{David Caruso},
  \bibinfo{person}{Derek Valleroy}, \bibinfo{person}{Dinesh Ginjupalli},
  \bibinfo{person}{Duncan Frost}, \bibinfo{person}{Edward Miller},
  \bibinfo{person}{Elias Mueggler}, \bibinfo{person}{Evgeniy Oleinik},
  \bibinfo{person}{Fan Zhang}, \bibinfo{person}{Guruprasad Somasundaram},
  \bibinfo{person}{Gustavo Solaira}, \bibinfo{person}{Harry Lanaras},
  \bibinfo{person}{Henry Howard-Jenkins}, \bibinfo{person}{Huixuan Tang},
  \bibinfo{person}{Hyo~Jin Kim}, \bibinfo{person}{Jaime Rivera},
  \bibinfo{person}{Ji Luo}, \bibinfo{person}{Jing Dong},
  \bibinfo{person}{Julian Straub}, \bibinfo{person}{Kevin Bailey},
  \bibinfo{person}{Kevin Eckenhoff}, \bibinfo{person}{Lingni Ma},
  \bibinfo{person}{Luis Pesqueira}, \bibinfo{person}{Mark Schwesinger},
  \bibinfo{person}{Maurizio Monge}, \bibinfo{person}{Nan Yang},
  \bibinfo{person}{Nick Charron}, \bibinfo{person}{Nikhil Raina},
  \bibinfo{person}{Omkar Parkhi}, \bibinfo{person}{Peter Borschowa},
  \bibinfo{person}{Pierre Moulon}, \bibinfo{person}{Prince Gupta},
  \bibinfo{person}{Raul Mur-Artal}, \bibinfo{person}{Robbie Pennington},
  \bibinfo{person}{Sachin Kulkarni}, \bibinfo{person}{Sagar Miglani},
  \bibinfo{person}{Santosh Gondi}, \bibinfo{person}{Saransh Solanki},
  \bibinfo{person}{Sean Diener}, \bibinfo{person}{Shangyi Cheng},
  \bibinfo{person}{Simon Green}, \bibinfo{person}{Steve Saarinen},
  \bibinfo{person}{Suvam Patra}, \bibinfo{person}{Tassos Mourikis},
  \bibinfo{person}{Thomas Whelan}, \bibinfo{person}{Tripti Singh},
  \bibinfo{person}{Vasileios Balntas}, \bibinfo{person}{Vijay Baiyya},
  \bibinfo{person}{Wilson Dreewes}, \bibinfo{person}{Xiaqing Pan},
  \bibinfo{person}{Yang Lou}, \bibinfo{person}{Yipu Zhao},
  \bibinfo{person}{Yusuf Mansour}, \bibinfo{person}{Yuyang Zou},
  \bibinfo{person}{Zhaoyang Lv}, \bibinfo{person}{Zijian Wang},
  \bibinfo{person}{Mingfei Yan}, \bibinfo{person}{Carl Ren},
  \bibinfo{person}{Renzo~De Nardi}, {and} \bibinfo{person}{Richard Newcombe}.}
  \bibinfo{year}{2023}\natexlab{}.
\newblock \bibinfo{title}{Project Aria: A New Tool for Egocentric Multi-Modal
  AI Research}.
\newblock
\newblock
\showeprint[arxiv]{2308.13561}~[cs.HC]


\bibitem[Fang et~al\mbox{.}(2015)]%
        {Fang:2015:EHC}
\bibfield{author}{\bibinfo{person}{Yu Fang}, \bibinfo{person}{Ryoichi
  Nakashima}, \bibinfo{person}{Kazumichi Matsumiya}, \bibinfo{person}{Ichiro
  Kuriki}, {and} \bibinfo{person}{Satoshi Shioiri}.}
  \bibinfo{year}{2015}\natexlab{}.
\newblock \showarticletitle{Eye-Head Coordination for Visual Cognitive
  Processing}.
\newblock \bibinfo{journal}{\emph{PLOS ONE}} \bibinfo{volume}{10},
  \bibinfo{number}{3} (\bibinfo{date}{03} \bibinfo{year}{2015}),
  \bibinfo{pages}{1--17}.
\newblock
\urldef\tempurl%
\url{https://doi.org/10.1371/journal.pone.0121035}
\showDOI{\tempurl}


\bibitem[Gallo et~al\mbox{.}(2015)]%
        {Gallo:2015:LNR}
\bibfield{author}{\bibinfo{person}{Orazio Gallo}, \bibinfo{person}{Alejandro
  Troccoli}, \bibinfo{person}{Jun Hu}, \bibinfo{person}{Kari Pulli}, {and}
  \bibinfo{person}{Jan Kautz}.} \bibinfo{year}{2015}\natexlab{}.
\newblock \showarticletitle{Locally non-rigid registration for mobile HDR
  photography}. In \bibinfo{booktitle}{\emph{2015 IEEE Conference on Computer
  Vision and Pattern Recognition Workshops (CVPRW)}}. \bibinfo{pages}{48--55}.
\newblock
\urldef\tempurl%
\url{https://doi.org/10.1109/CVPRW.2015.7301366}
\showDOI{\tempurl}


\bibitem[Gharbi et~al\mbox{.}(2016)]%
        {gharbi2016}
\bibfield{author}{\bibinfo{person}{Micha{\"e}l Gharbi}, \bibinfo{person}{Gaurav
  Chaurasia}, \bibinfo{person}{Sylvain Paris}, {and} \bibinfo{person}{Fr{\'e}do
  Durand}.} \bibinfo{year}{2016}\natexlab{}.
\newblock \showarticletitle{Deep joint demosaicking and denoising}.
\newblock \bibinfo{journal}{\emph{ACM Transactions on Graphics (ToG)}}
  \bibinfo{volume}{35}, \bibinfo{number}{6} (\bibinfo{year}{2016}),
  \bibinfo{pages}{1--12}.
\newblock


\bibitem[Gortler et~al\mbox{.}(1996)]%
        {gortler1996lumigraph}
\bibfield{author}{\bibinfo{person}{Steven~J. Gortler}, \bibinfo{person}{Radek
  Grzeszczuk}, \bibinfo{person}{Richard Szeliski}, {and}
  \bibinfo{person}{Michael~F. Cohen}.} \bibinfo{year}{1996}\natexlab{}.
\newblock \showarticletitle{The lumigraph}. In
  \bibinfo{booktitle}{\emph{Proceedings of the 23rd Annual Conference on
  Computer Graphics and Interactive Techniques}}
  \emph{(\bibinfo{series}{SIGGRAPH '96})}. \bibinfo{publisher}{Association for
  Computing Machinery}, \bibinfo{address}{New York, NY, USA},
  \bibinfo{pages}{43–54}.
\newblock
\showISBNx{0897917464}
\urldef\tempurl%
\url{https://doi.org/10.1145/237170.237200}
\showDOI{\tempurl}


\bibitem[Grauman et~al\mbox{.}(2022)]%
        {Ego4D2022CVPR}
\bibfield{author}{\bibinfo{person}{Kristen Grauman}, \bibinfo{person}{Andrew
  Westbury}, \bibinfo{person}{Eugene Byrne}, \bibinfo{person}{Zachary Chavis},
  \bibinfo{person}{Antonino Furnari}, \bibinfo{person}{Rohit Girdhar},
  \bibinfo{person}{Jackson Hamburger}, \bibinfo{person}{Hao Jiang},
  \bibinfo{person}{Miao Liu}, \bibinfo{person}{Xingyu Liu},
  \bibinfo{person}{Miguel Martin}, \bibinfo{person}{Tushar Nagarajan},
  \bibinfo{person}{Ilija Radosavovic}, \bibinfo{person}{Santhosh~Kumar
  Ramakrishnan}, \bibinfo{person}{Fiona Ryan}, \bibinfo{person}{Jayant Sharma},
  \bibinfo{person}{Michael Wray}, \bibinfo{person}{Mengmeng Xu},
  \bibinfo{person}{Eric~Zhongcong Xu}, \bibinfo{person}{Chen Zhao},
  \bibinfo{person}{Siddhant Bansal}, \bibinfo{person}{Dhruv Batra},
  \bibinfo{person}{Vincent Cartillier}, \bibinfo{person}{Sean Crane},
  \bibinfo{person}{Tien Do}, \bibinfo{person}{Morrie Doulaty},
  \bibinfo{person}{Akshay Erapalli}, \bibinfo{person}{Christoph Feichtenhofer},
  \bibinfo{person}{Adriano Fragomeni}, \bibinfo{person}{Qichen Fu},
  \bibinfo{person}{Christian Fuegen}, \bibinfo{person}{Abrham Gebreselasie},
  \bibinfo{person}{Cristina Gonzalez}, \bibinfo{person}{James Hillis},
  \bibinfo{person}{Xuhua Huang}, \bibinfo{person}{Yifei Huang},
  \bibinfo{person}{Wenqi Jia}, \bibinfo{person}{Weslie Khoo},
  \bibinfo{person}{Jachym Kolar}, \bibinfo{person}{Satwik Kottur},
  \bibinfo{person}{Anurag Kumar}, \bibinfo{person}{Federico Landini},
  \bibinfo{person}{Chao Li}, \bibinfo{person}{Yanghao Li},
  \bibinfo{person}{Zhenqiang Li}, \bibinfo{person}{Karttikeya Mangalam},
  \bibinfo{person}{Raghava Modhugu}, \bibinfo{person}{Jonathan Munro},
  \bibinfo{person}{Tullie Murrell}, \bibinfo{person}{Takumi Nishiyasu},
  \bibinfo{person}{Will Price}, \bibinfo{person}{Paola~Ruiz Puentes},
  \bibinfo{person}{Merey Ramazanova}, \bibinfo{person}{Leda Sari},
  \bibinfo{person}{Kiran Somasundaram}, \bibinfo{person}{Audrey Southerland},
  \bibinfo{person}{Yusuke Sugano}, \bibinfo{person}{Ruijie Tao},
  \bibinfo{person}{Minh Vo}, \bibinfo{person}{Yuchen Wang},
  \bibinfo{person}{Xindi Wu}, \bibinfo{person}{Takuma Yagi},
  \bibinfo{person}{Yunyi Zhu}, \bibinfo{person}{Pablo Arbelaez},
  \bibinfo{person}{David Crandall}, \bibinfo{person}{Dima Damen},
  \bibinfo{person}{Giovanni~Maria Farinella}, \bibinfo{person}{Bernard Ghanem},
  \bibinfo{person}{Vamsi~Krishna Ithapu}, \bibinfo{person}{C.~V. Jawahar},
  \bibinfo{person}{Hanbyul Joo}, \bibinfo{person}{Kris Kitani},
  \bibinfo{person}{Haizhou Li}, \bibinfo{person}{Richard Newcombe},
  \bibinfo{person}{Aude Oliva}, \bibinfo{person}{Hyun~Soo Park},
  \bibinfo{person}{James~M. Rehg}, \bibinfo{person}{Yoichi Sato},
  \bibinfo{person}{Jianbo Shi}, \bibinfo{person}{Mike~Zheng Shou},
  \bibinfo{person}{Antonio Torralba}, \bibinfo{person}{Lorenzo Torresani},
  \bibinfo{person}{Mingfei Yan}, {and} \bibinfo{person}{Jitendra Malik}.}
  \bibinfo{year}{2022}\natexlab{}.
\newblock \showarticletitle{Ego4D: Around the {W}orld in 3,000 {H}ours of
  {E}gocentric {V}ideo}. In \bibinfo{booktitle}{\emph{IEEE/CVF Computer Vision
  and Pattern Recognition (CVPR)}}.
\newblock


\bibitem[Hecht et~al\mbox{.}(1942)]%
        {Hecht:1942:EQV}
\bibfield{author}{\bibinfo{person}{Selig Hecht}, \bibinfo{person}{Simon
  Shlaer}, {and} \bibinfo{person}{Maurice~Henri Pirenne}.}
  \bibinfo{year}{1942}\natexlab{}.
\newblock \showarticletitle{Energy, quanta, and vision}.
\newblock \bibinfo{journal}{\emph{The Journal of general physiology}}
  \bibinfo{volume}{25}, \bibinfo{number}{6} (\bibinfo{year}{1942}),
  \bibinfo{pages}{819--840}.
\newblock


\bibitem[Hood and Finkelstein(1986)]%
        {Hood:1986:STL}
\bibfield{author}{\bibinfo{person}{DC Hood} {and} \bibinfo{person}{MA
  Finkelstein}.} \bibinfo{year}{1986}\natexlab{}.
\newblock \bibinfo{booktitle}{\emph{Handbook of Perception and Human
  Performance}}. Vol.~\bibinfo{volume}{1}.
\newblock \bibinfo{publisher}{Wiley Interscience}.
\newblock


\bibitem[Howard and Rogers(1995)]%
        {Howard:1995:BVS}
\bibfield{author}{\bibinfo{person}{Ian~P. Howard} {and}
  \bibinfo{person}{Brian~J. Rogers}.} \bibinfo{year}{1995}\natexlab{}.
\newblock \bibinfo{booktitle}{\emph{Binocular Vision and Stereopsis}}.
\newblock \bibinfo{publisher}{Oxford University Press}.
\newblock
\showISBNx{9780195084764}
\showLCCN{94045728}


\bibitem[Huang et~al\mbox{.}(2022)]%
        {huang2022task}
\bibfield{author}{\bibinfo{person}{Yixuan Huang}, \bibinfo{person}{Xiaoyun
  Zhang}, \bibinfo{person}{Yu Fu}, \bibinfo{person}{Siheng Chen},
  \bibinfo{person}{Ya Zhang}, \bibinfo{person}{Yan-Feng Wang}, {and}
  \bibinfo{person}{Dazhi He}.} \bibinfo{year}{2022}\natexlab{}.
\newblock \showarticletitle{Task decoupled framework for reference-based
  super-resolution}. In \bibinfo{booktitle}{\emph{Proceedings of the IEEE/CVF
  Conference on Computer Vision and Pattern Recognition}}.
  \bibinfo{pages}{5931--5940}.
\newblock


\bibitem[ISO 12232:2019(E)(2019)]%
        {ISO12232}
ISO 12232:2019(E) \bibinfo{year}{2019}\natexlab{}.
\newblock \bibinfo{booktitle}{\emph{{Photography -- Digital still cameras --
  Determination of exposure index, ISO speed ratings, standard output
  sensitivity, and recommended exposure index}}}.
\newblock \bibinfo{type}{Standard}. \bibinfo{institution}{International
  Organization for Standardization}, \bibinfo{address}{Geneva, CH}.
\newblock


\bibitem[ISO 12870:2016(E)(2016)]%
        {ISO12870}
ISO 12870:2016(E) \bibinfo{year}{2016}\natexlab{}.
\newblock \bibinfo{booktitle}{\emph{{Ophthalmic optics — Spectacle frames —
  Requirements and test methods}}}.
\newblock \bibinfo{type}{Standard}. \bibinfo{institution}{International
  Organization for Standardization}, \bibinfo{address}{Geneva, CH}.
\newblock


\bibitem[Jiang et~al\mbox{.}(2021)]%
        {jiang2021robust}
\bibfield{author}{\bibinfo{person}{Yuming Jiang}, \bibinfo{person}{Kelvin~CK
  Chan}, \bibinfo{person}{Xintao Wang}, \bibinfo{person}{Chen~Change Loy},
  {and} \bibinfo{person}{Ziwei Liu}.} \bibinfo{year}{2021}\natexlab{}.
\newblock \showarticletitle{Robust reference-based super-resolution via
  c2-matching}. In \bibinfo{booktitle}{\emph{CVPR}}.
  \bibinfo{pages}{2103--2112}.
\newblock


\bibitem[Kim et~al\mbox{.}(2023)]%
        {kim2023efficient}
\bibfield{author}{\bibinfo{person}{Youngrae Kim}, \bibinfo{person}{Jinsu Lim},
  \bibinfo{person}{Hoonhee Cho}, \bibinfo{person}{Minji Lee},
  \bibinfo{person}{Dongman Lee}, \bibinfo{person}{Kuk-Jin Yoon}, {and}
  \bibinfo{person}{Ho-Jin Choi}.} \bibinfo{year}{2023}\natexlab{}.
\newblock \showarticletitle{Efficient Reference-based Video Super-Resolution
  (ERVSR): Single Reference Image Is All You Need}. In
  \bibinfo{booktitle}{\emph{Proceedings of the IEEE/CVF Winter Conference on
  Applications of Computer Vision}}. \bibinfo{pages}{1828--1837}.
\newblock


\bibitem[Kim et~al\mbox{.}(2021)]%
        {Kim:2021:WCP}
\bibfield{author}{\bibinfo{person}{Yong~Min Kim}, \bibinfo{person}{Sangwoo
  Bahn}, {and} \bibinfo{person}{Myung~Hwan Yun}.}
  \bibinfo{year}{2021}\natexlab{}.
\newblock \showarticletitle{Wearing comfort and perceived heaviness of smart
  glasses}.
\newblock \bibinfo{journal}{\emph{Human Factors and Ergonomics in Manufacturing
  \& Service Industries}}  \bibinfo{volume}{31} (\bibinfo{year}{2021}),
  \bibinfo{pages}{484--495}.
\newblock
Issue 5.


\bibitem[Kirillov et~al\mbox{.}(2023)]%
        {kirillov2023segment}
\bibfield{author}{\bibinfo{person}{Alexander Kirillov}, \bibinfo{person}{Eric
  Mintun}, \bibinfo{person}{Nikhila Ravi}, \bibinfo{person}{Hanzi Mao},
  \bibinfo{person}{Chloe Rolland}, \bibinfo{person}{Laura Gustafson},
  \bibinfo{person}{Tete Xiao}, \bibinfo{person}{Spencer Whitehead},
  \bibinfo{person}{Alexander~C Berg}, \bibinfo{person}{Wan-Yen Lo},
  {et~al\mbox{.}}} \bibinfo{year}{2023}\natexlab{}.
\newblock \showarticletitle{Segment anything}.
\newblock \bibinfo{journal}{\emph{arXiv preprint arXiv:2304.02643}}
  (\bibinfo{year}{2023}).
\newblock


\bibitem[Lawrence et~al\mbox{.}(2021)]%
        {lawrence2021project}
\bibfield{author}{\bibinfo{person}{Jason Lawrence}, \bibinfo{person}{Danb
  Goldman}, \bibinfo{person}{Supreeth Achar}, \bibinfo{person}{Gregory~Major
  Blascovich}, \bibinfo{person}{Joseph~G Desloge}, \bibinfo{person}{Tommy
  Fortes}, \bibinfo{person}{Eric~M Gomez}, \bibinfo{person}{Sascha
  H{\"a}berling}, \bibinfo{person}{Hugues Hoppe}, \bibinfo{person}{Andy
  Huibers}, {et~al\mbox{.}}} \bibinfo{year}{2021}\natexlab{}.
\newblock \showarticletitle{Project starline: a high-fidelity telepresence
  system}.
\newblock \bibinfo{journal}{\emph{ACM Transactions on Graphics (TOG)}}
  \bibinfo{volume}{40}, \bibinfo{number}{6} (\bibinfo{year}{2021}),
  \bibinfo{pages}{1--16}.
\newblock


\bibitem[Lee et~al\mbox{.}(2022)]%
        {lee2022reference}
\bibfield{author}{\bibinfo{person}{Junyong Lee}, \bibinfo{person}{Myeonghee
  Lee}, \bibinfo{person}{Sunghyun Cho}, {and} \bibinfo{person}{Seungyong Lee}.}
  \bibinfo{year}{2022}\natexlab{}.
\newblock \showarticletitle{Reference-based video super-resolution using
  multi-camera video triplets}. In \bibinfo{booktitle}{\emph{Proceedings of the
  IEEE/CVF Conference on Computer Vision and Pattern Recognition}}.
  \bibinfo{pages}{17824--17833}.
\newblock


\bibitem[Levoy and Hanrahan(1996)]%
        {levoy1996light}
\bibfield{author}{\bibinfo{person}{Marc Levoy} {and} \bibinfo{person}{Pat
  Hanrahan}.} \bibinfo{year}{1996}\natexlab{}.
\newblock \showarticletitle{Light field rendering}. In
  \bibinfo{booktitle}{\emph{SIGGRAPH 96}}. \bibinfo{pages}{31--42}.
\newblock


\bibitem[Liba et~al\mbox{.}(2019)]%
        {Liba:2019:HMP}
\bibfield{author}{\bibinfo{person}{Orly Liba}, \bibinfo{person}{Kiran Murthy},
  \bibinfo{person}{Yun-Ta Tsai}, \bibinfo{person}{Tim Brooks},
  \bibinfo{person}{Tianfan Xue}, \bibinfo{person}{Nikhil Karnad},
  \bibinfo{person}{Qiurui He}, \bibinfo{person}{Jonathan~T. Barron},
  \bibinfo{person}{Dillon Sharlet}, \bibinfo{person}{Ryan Geiss},
  \bibinfo{person}{Samuel~W. Hasinoff}, \bibinfo{person}{Yael Pritch}, {and}
  \bibinfo{person}{Marc Levoy}.} \bibinfo{year}{2019}\natexlab{}.
\newblock \showarticletitle{Handheld Mobile Photography in Very Low Light}.
\newblock \bibinfo{journal}{\emph{ACM Trans. Graph.}} \bibinfo{volume}{38},
  \bibinfo{number}{6}, Article \bibinfo{articleno}{164} (\bibinfo{date}{nov}
  \bibinfo{year}{2019}), \bibinfo{numpages}{16}~pages.
\newblock
\showISSN{0730-0301}
\urldef\tempurl%
\url{https://doi.org/10.1145/3355089.3356508}
\showDOI{\tempurl}


\bibitem[Lucas and Kanade(1981)]%
        {lucas1981iterative}
\bibfield{author}{\bibinfo{person}{Bruce~D Lucas} {and} \bibinfo{person}{Takeo
  Kanade}.} \bibinfo{year}{1981}\natexlab{}.
\newblock \showarticletitle{An iterative image registration technique with an
  application to stereo vision}. In \bibinfo{booktitle}{\emph{IJCAI'81: 7th
  international joint conference on Artificial intelligence}},
  Vol.~\bibinfo{volume}{2}. \bibinfo{pages}{674--679}.
\newblock


\bibitem[Mann(2013)]%
        {Mann:2013:MAL}
\bibfield{author}{\bibinfo{person}{Steve Mann}.}
  \bibinfo{year}{2013}\natexlab{}.
\newblock \showarticletitle{My \"Augmediated\" Life}.
\newblock \bibinfo{journal}{\emph{IEEE Spectrum}} (\bibinfo{year}{2013}).
\newblock


\bibitem[{Meta Reality Labs-R}(2022)]%
        {APD}
\bibfield{author}{\bibinfo{person}{{Meta Reality Labs-R}}.}
  \bibinfo{year}{2022}\natexlab{}.
\newblock \bibinfo{title}{Aria Pilot Dataset}.
\newblock
  \bibinfo{howpublished}{\url{https://www.projectaria.com/datasets/apd/}}.
\newblock


\bibitem[{Meta Reality Labs-R}(2023)]%
        {ASE}
\bibfield{author}{\bibinfo{person}{{Meta Reality Labs-R}}.}
  \bibinfo{year}{2023}\natexlab{}.
\newblock \bibinfo{title}{Aria Synthetic Environments Dataset}.
\newblock
  \bibinfo{howpublished}{\url{https://www.projectaria.com/datasets/ase/}}.
\newblock


\bibitem[Mildenhall et~al\mbox{.}(2022)]%
        {mildenhall2022rawnerf}
\bibfield{author}{\bibinfo{person}{Ben Mildenhall}, \bibinfo{person}{Peter
  Hedman}, \bibinfo{person}{Ricardo Martin-Brualla}, \bibinfo{person}{Pratul~P.
  Srinivasan}, {and} \bibinfo{person}{Jonathan~T. Barron}.}
  \bibinfo{year}{2022}\natexlab{}.
\newblock \showarticletitle{{NeRF} in the Dark: High Dynamic Range View
  Synthesis from Noisy Raw Images}.
\newblock \bibinfo{journal}{\emph{CVPR}} (\bibinfo{year}{2022}).
\newblock


\bibitem[Mildenhall et~al\mbox{.}(2021)]%
        {mildenhall2021nerf}
\bibfield{author}{\bibinfo{person}{Ben Mildenhall}, \bibinfo{person}{Pratul~P
  Srinivasan}, \bibinfo{person}{Matthew Tancik}, \bibinfo{person}{Jonathan~T
  Barron}, \bibinfo{person}{Ravi Ramamoorthi}, {and} \bibinfo{person}{Ren Ng}.}
  \bibinfo{year}{2021}\natexlab{}.
\newblock \showarticletitle{Nerf: Representing scenes as neural radiance fields
  for view synthesis}.
\newblock \bibinfo{journal}{\emph{Commun. ACM}} \bibinfo{volume}{65},
  \bibinfo{number}{1} (\bibinfo{year}{2021}), \bibinfo{pages}{99--106}.
\newblock


\bibitem[Mordi and Ciuffreda(1998)]%
        {Mordi:1998:PAM}
\bibfield{author}{\bibinfo{person}{John~A Mordi} {and}
  \bibinfo{person}{Kenneth~J Ciuffreda}.} \bibinfo{year}{1998}\natexlab{}.
\newblock \showarticletitle{Static aspects of accommodation: age and
  presbyopia}.
\newblock \bibinfo{journal}{\emph{Vision Research}} \bibinfo{volume}{38},
  \bibinfo{number}{11} (\bibinfo{year}{1998}), \bibinfo{pages}{1643--1653}.
\newblock
\showISSN{0042-6989}
\urldef\tempurl%
\url{https://doi.org/10.1016/S0042-6989(97)00336-2}
\showDOI{\tempurl}


\bibitem[Mourikis and Roumeliotis(2007)]%
        {mourikis2007multi}
\bibfield{author}{\bibinfo{person}{Anastasios~I Mourikis} {and}
  \bibinfo{person}{Stergios~I Roumeliotis}.} \bibinfo{year}{2007}\natexlab{}.
\newblock \showarticletitle{A multi-state constraint {K}alman filter for
  vision-aided inertial navigation}. In \bibinfo{booktitle}{\emph{Proceedings
  2007 IEEE International Conference on Robotics and Automation}}. IEEE,
  \bibinfo{pages}{3565--3572}.
\newblock


\bibitem[Mur-Artal and Tard{\'o}s(2017)]%
        {mur2017orb}
\bibfield{author}{\bibinfo{person}{Raul Mur-Artal} {and}
  \bibinfo{person}{Juan~D Tard{\'o}s}.} \bibinfo{year}{2017}\natexlab{}.
\newblock \showarticletitle{{ORB-SLAM2}: An open-source slam system for
  monocular, stereo, and {RBG-D} cameras}.
\newblock \bibinfo{journal}{\emph{IEEE Transactions on Robotics}}
  \bibinfo{volume}{33}, \bibinfo{number}{5} (\bibinfo{year}{2017}),
  \bibinfo{pages}{1255--1262}.
\newblock


\bibitem[Nomura et~al\mbox{.}(2007)]%
        {nomura2007scene}
\bibfield{author}{\bibinfo{person}{Yoshikuni Nomura}, \bibinfo{person}{Li
  Zhang}, {and} \bibinfo{person}{Shree~K Nayar}.}
  \bibinfo{year}{2007}\natexlab{}.
\newblock \showarticletitle{Scene collages and flexible camera arrays}. In
  \bibinfo{booktitle}{\emph{Proceedings of the 18th Eurographics conference on
  Rendering Techniques}}. \bibinfo{pages}{127--138}.
\newblock


\bibitem[Pears(1998)]%
        {pears1998strategic}
\bibfield{author}{\bibinfo{person}{Alan Pears}.}
  \bibinfo{year}{1998}\natexlab{}.
\newblock \bibinfo{booktitle}{\emph{Strategic study of household energy and
  greenhouse issues}}.
\newblock \bibinfo{publisher}{Sustainable Solutions Australia}.
\newblock


\bibitem[Perazzi et~al\mbox{.}(2015)]%
        {perazzi2015panoramic}
\bibfield{author}{\bibinfo{person}{Federico Perazzi},
  \bibinfo{person}{Alexander Sorkine-Hornung}, \bibinfo{person}{Henning
  Zimmer}, \bibinfo{person}{Peter Kaufmann}, \bibinfo{person}{Oliver Wang},
  \bibinfo{person}{Scott Watson}, {and} \bibinfo{person}{Markus Gross}.}
  \bibinfo{year}{2015}\natexlab{}.
\newblock \showarticletitle{Panoramic video from unstructured camera arrays}.
  In \bibinfo{booktitle}{\emph{Computer Graphics Forum}},
  Vol.~\bibinfo{volume}{34}. Wiley Online Library, \bibinfo{pages}{57--68}.
\newblock


\bibitem[Pesavento et~al\mbox{.}(2021)]%
        {pesavento2021attention}
\bibfield{author}{\bibinfo{person}{Marco Pesavento}, \bibinfo{person}{Marco
  Volino}, {and} \bibinfo{person}{Adrian Hilton}.}
  \bibinfo{year}{2021}\natexlab{}.
\newblock \showarticletitle{Attention-based multi-reference learning for image
  super-resolution}. In \bibinfo{booktitle}{\emph{Proceedings of the IEEE/CVF
  International Conference on Computer Vision}}. \bibinfo{pages}{14697--14706}.
\newblock


\bibitem[Rayleigh(1879)]%
        {Rayleigh:1879:IOS}
\bibfield{author}{\bibinfo{person}{Rayleigh}.} \bibinfo{year}{1879}\natexlab{}.
\newblock \showarticletitle{XXXI. Investigations in optics, with special
  reference to the spectroscope}.
\newblock \bibinfo{journal}{\emph{The London, Edinburgh, and Dublin
  Philosophical Magazine and Journal of Science}} \bibinfo{volume}{8},
  \bibinfo{number}{49} (\bibinfo{year}{1879}), \bibinfo{pages}{261--274}.
\newblock
\urldef\tempurl%
\url{https://doi.org/10.1080/14786447908639684}
\showDOI{\tempurl}


\bibitem[Rombach et~al\mbox{.}(2022)]%
        {rombach2022high}
\bibfield{author}{\bibinfo{person}{Robin Rombach}, \bibinfo{person}{Andreas
  Blattmann}, \bibinfo{person}{Dominik Lorenz}, \bibinfo{person}{Patrick
  Esser}, {and} \bibinfo{person}{Bj{\"o}rn Ommer}.}
  \bibinfo{year}{2022}\natexlab{}.
\newblock \showarticletitle{High-resolution image synthesis with latent
  diffusion models}. In \bibinfo{booktitle}{\emph{Proceedings of the IEEE/CVF
  conference on computer vision and pattern recognition}}.
  \bibinfo{pages}{10684--10695}.
\newblock


\bibitem[Rosenholtz(2016)]%
        {Rosenholtz:2016:CLP}
\bibfield{author}{\bibinfo{person}{Ruth Rosenholtz}.}
  \bibinfo{year}{2016}\natexlab{}.
\newblock \showarticletitle{Capabilities and Limitations of Peripheral Vision}.
\newblock \bibinfo{journal}{\emph{Annual Review of Vision Science}}
  \bibinfo{volume}{2}, \bibinfo{number}{1} (\bibinfo{year}{2016}),
  \bibinfo{pages}{437--457}.
\newblock
\urldef\tempurl%
\url{https://doi.org/10.1146/annurev-vision-082114-035733}
\showDOI{\tempurl}
\showeprint{https://doi.org/10.1146/annurev-vision-082114-035733}
\newblock
\shownote{PMID: 28532349}.


\bibitem[Sahin and Laroia(2017)]%
        {Sahin:2017:LLC}
\bibfield{author}{\bibinfo{person}{Furkan~E. Sahin} {and}
  \bibinfo{person}{Rajiv Laroia}.} \bibinfo{year}{2017}\natexlab{}.
\newblock \showarticletitle{Light L16 Computational Camera}, In
  \bibinfo{booktitle}{Imaging and Applied Optics 2017 (3D, AIO, COSI, IS, MATH,
  pcAOP)}.
\newblock \bibinfo{journal}{\emph{Imaging and Applied Optics 2017 (3D, AIO,
  COSI, IS, MATH, pcAOP)}}, \bibinfo{pages}{JTu5A.20}.
\newblock
\urldef\tempurl%
\url{https://doi.org/10.1364/3D.2017.JTu5A.20}
\showDOI{\tempurl}


\bibitem[Teed and Deng(2020)]%
        {teed2020raft}
\bibfield{author}{\bibinfo{person}{Zachary Teed} {and} \bibinfo{person}{Jia
  Deng}.} \bibinfo{year}{2020}\natexlab{}.
\newblock \showarticletitle{Raft: Recurrent all-pairs field transforms for
  optical flow}. In \bibinfo{booktitle}{\emph{ECCV}}. Springer,
  \bibinfo{pages}{402--419}.
\newblock


\bibitem[Tinsley et~al\mbox{.}(2016)]%
        {Tinsley:2016:DDS}
\bibfield{author}{\bibinfo{person}{Jonathan~N. Tinsley},
  \bibinfo{person}{Maxim~I. Molodtsov}, \bibinfo{person}{Robert Prevedel},
  \bibinfo{person}{David Wartmann}, \bibinfo{person}{Jofre Espigul{\'e}-Pons},
  \bibinfo{person}{Mattias Lauwers}, {and} \bibinfo{person}{Alipasha Vaziri}.}
  \bibinfo{year}{2016}\natexlab{}.
\newblock \showarticletitle{Direct detection of a single photon by humans}.
\newblock \bibinfo{journal}{\emph{Nature Communications}} \bibinfo{volume}{7},
  \bibinfo{number}{1} (\bibinfo{year}{2016}), \bibinfo{pages}{12172}.
\newblock
\showISBNx{2041-1723}
\urldef\tempurl%
\url{https://doi.org/10.1038/ncomms12172}
\showDOI{\tempurl}


\bibitem[Trinidad et~al\mbox{.}(2019)]%
        {trinidad2019multi}
\bibfield{author}{\bibinfo{person}{Marc~Comino Trinidad},
  \bibinfo{person}{Ricardo~Martin Brualla}, \bibinfo{person}{Florian Kainz},
  {and} \bibinfo{person}{Janne Kontkanen}.} \bibinfo{year}{2019}\natexlab{}.
\newblock \showarticletitle{Multi-view image fusion}. In
  \bibinfo{booktitle}{\emph{Proceedings of the IEEE/CVF International
  Conference on Computer Vision}}. \bibinfo{pages}{4101--4110}.
\newblock


\bibitem[Venkataraman et~al\mbox{.}(2013)]%
        {Venkataraman:2013:PUT}
\bibfield{author}{\bibinfo{person}{Kartik Venkataraman}, \bibinfo{person}{Dan
  Lelescu}, \bibinfo{person}{Jacques Duparr\'{e}}, \bibinfo{person}{Andrew
  McMahon}, \bibinfo{person}{Gabriel Molina}, \bibinfo{person}{Priyam
  Chatterjee}, \bibinfo{person}{Robert Mullis}, {and} \bibinfo{person}{Shree
  Nayar}.} \bibinfo{year}{2013}\natexlab{}.
\newblock \showarticletitle{PiCam: an ultra-thin high performance monolithic
  camera array}.
\newblock \bibinfo{journal}{\emph{ACM Trans. Graph.}} \bibinfo{volume}{32},
  \bibinfo{number}{6}, Article \bibinfo{articleno}{166} (\bibinfo{date}{Nov.}
  \bibinfo{year}{2013}), \bibinfo{numpages}{13}~pages.
\newblock
\showISSN{0730-0301}
\urldef\tempurl%
\url{https://doi.org/10.1145/2508363.2508390}
\showDOI{\tempurl}


\bibitem[Wang et~al\mbox{.}(2023)]%
        {Wang_2023_CVPR}
\bibfield{author}{\bibinfo{person}{Jialiang Wang}, \bibinfo{person}{Daniel
  Scharstein}, \bibinfo{person}{Akash Bapat}, \bibinfo{person}{Kevin
  Blackburn-Matzen}, \bibinfo{person}{Matthew Yu}, \bibinfo{person}{Jonathan
  Lehman}, \bibinfo{person}{Suhib Alsisan}, \bibinfo{person}{Yanghan Wang},
  \bibinfo{person}{Sam Tsai}, \bibinfo{person}{Jan-Michael Frahm},
  \bibinfo{person}{Zijian He}, \bibinfo{person}{Peter Vajda},
  \bibinfo{person}{Michael~F. Cohen}, {and} \bibinfo{person}{Matt
  Uyttendaele}.} \bibinfo{year}{2023}\natexlab{}.
\newblock \showarticletitle{A Practical Stereo Depth System for Smart Glasses}.
  In \bibinfo{booktitle}{\emph{Proceedings of the IEEE/CVF Conference on
  Computer Vision and Pattern Recognition (CVPR)}}.
  \bibinfo{pages}{21498--21507}.
\newblock


\bibitem[Wang et~al\mbox{.}(2021)]%
        {wang2021dual}
\bibfield{author}{\bibinfo{person}{Tengfei Wang}, \bibinfo{person}{Jiaxin Xie},
  \bibinfo{person}{Wenxiu Sun}, \bibinfo{person}{Qiong Yan}, {and}
  \bibinfo{person}{Qifeng Chen}.} \bibinfo{year}{2021}\natexlab{}.
\newblock \showarticletitle{Dual-camera super-resolution with aligned attention
  modules}. In \bibinfo{booktitle}{\emph{ICCV}}. \bibinfo{pages}{2001--2010}.
\newblock


\bibitem[Wang et~al\mbox{.}(2004)]%
        {Wang:2003:IQA}
\bibfield{author}{\bibinfo{person}{Zhou Wang}, \bibinfo{person}{A.C. Bovik},
  \bibinfo{person}{H.R. Sheikh}, {and} \bibinfo{person}{E.P. Simoncelli}.}
  \bibinfo{year}{2004}\natexlab{}.
\newblock \showarticletitle{Image quality assessment: from error visibility to
  structural similarity}.
\newblock \bibinfo{journal}{\emph{IEEE Transactions on Image Processing}}
  \bibinfo{volume}{13}, \bibinfo{number}{4} (\bibinfo{year}{2004}),
  \bibinfo{pages}{600--612}.
\newblock
\urldef\tempurl%
\url{https://doi.org/10.1109/TIP.2003.819861}
\showDOI{\tempurl}


\bibitem[Wilburn et~al\mbox{.}(2005)]%
        {wilburn2005high}
\bibfield{author}{\bibinfo{person}{Bennett Wilburn}, \bibinfo{person}{Neel
  Joshi}, \bibinfo{person}{Vaibhav Vaish}, \bibinfo{person}{Eino-Ville
  Talvala}, \bibinfo{person}{Emilio Antunez}, \bibinfo{person}{Adam Barth},
  \bibinfo{person}{Andrew Adams}, \bibinfo{person}{Mark Horowitz}, {and}
  \bibinfo{person}{Marc Levoy}.} \bibinfo{year}{2005}\natexlab{}.
\newblock \showarticletitle{High performance imaging using large camera
  arrays}.
\newblock In \bibinfo{booktitle}{\emph{ACM SIGGRAPH 2005 Papers}}.
  \bibinfo{pages}{765--776}.
\newblock


\bibitem[Wronski et~al\mbox{.}(2019)]%
        {Wronski:2019:HMS}
\bibfield{author}{\bibinfo{person}{Bartlomiej Wronski},
  \bibinfo{person}{Ignacio Garcia-Dorado}, \bibinfo{person}{Manfred Ernst},
  \bibinfo{person}{Damien Kelly}, \bibinfo{person}{Michael Krainin},
  \bibinfo{person}{Chia-Kai Liang}, \bibinfo{person}{Marc Levoy}, {and}
  \bibinfo{person}{Peyman Milanfar}.} \bibinfo{year}{2019}\natexlab{}.
\newblock \showarticletitle{Handheld Multi-Frame Super-Resolution}.
\newblock \bibinfo{journal}{\emph{ACM Trans. Graph.}} \bibinfo{volume}{38},
  \bibinfo{number}{4}, Article \bibinfo{articleno}{28} (\bibinfo{date}{jul}
  \bibinfo{year}{2019}), \bibinfo{numpages}{18}~pages.
\newblock
\showISSN{0730-0301}
\urldef\tempurl%
\url{https://doi.org/10.1145/3306346.3323024}
\showDOI{\tempurl}


\bibitem[Wu et~al\mbox{.}(2023)]%
        {wu2023efficient}
\bibfield{author}{\bibinfo{person}{Xiaotong Wu}, \bibinfo{person}{Wei-Sheng
  Lai}, \bibinfo{person}{Yichang Shih}, \bibinfo{person}{Charles Herrmann},
  \bibinfo{person}{Michael Krainin}, \bibinfo{person}{Deqing Sun}, {and}
  \bibinfo{person}{Chia-Kai Liang}.} \bibinfo{year}{2023}\natexlab{}.
\newblock \showarticletitle{Efficient Hybrid Zoom Using Camera Fusion on Mobile
  Phones}.
\newblock \bibinfo{journal}{\emph{ACM Transactions on Graphics (TOG)}}
  \bibinfo{volume}{42}, \bibinfo{number}{6} (\bibinfo{year}{2023}),
  \bibinfo{pages}{1--12}.
\newblock


\bibitem[Yuan et~al\mbox{.}(2017)]%
        {yuan2017multiscale}
\bibfield{author}{\bibinfo{person}{Xiaoyun Yuan}, \bibinfo{person}{Lu Fang},
  \bibinfo{person}{Qionghai Dai}, \bibinfo{person}{David~J Brady}, {and}
  \bibinfo{person}{Yebin Liu}.} \bibinfo{year}{2017}\natexlab{}.
\newblock \showarticletitle{Multiscale gigapixel video: A cross resolution
  image matching and warping approach}. In \bibinfo{booktitle}{\emph{2017 IEEE
  International Conference on Computational Photography (ICCP)}}. IEEE,
  \bibinfo{pages}{1--9}.
\newblock


\bibitem[Zhang et~al\mbox{.}(2023)]%
        {zhang2023lmr}
\bibfield{author}{\bibinfo{person}{Lin Zhang}, \bibinfo{person}{Xin Li},
  \bibinfo{person}{Dongliang He}, \bibinfo{person}{Fu Li},
  \bibinfo{person}{Errui Ding}, {and} \bibinfo{person}{Zhaoxiang Zhang}.}
  \bibinfo{year}{2023}\natexlab{}.
\newblock \showarticletitle{LMR: A Large-Scale Multi-Reference Dataset for
  Reference-based Super-Resolution}. In \bibinfo{booktitle}{\emph{Proceedings
  of the IEEE/CVF International Conference on Computer Vision}}.
  \bibinfo{pages}{13118--13127}.
\newblock


\bibitem[Zhang et~al\mbox{.}(2018)]%
        {Zhang:2018:ASE}
\bibfield{author}{\bibinfo{person}{Yupeng Zhang}, \bibinfo{person}{Liyan Liu},
  \bibinfo{person}{Weitao Gong}, \bibinfo{person}{Haihua Yu},
  \bibinfo{person}{Wei Wang}, \bibinfo{person}{Chongying Zhao},
  \bibinfo{person}{Peng Wang}, {and} \bibinfo{person}{Toshitsugu Ueda}.}
  \bibinfo{year}{2018}\natexlab{}.
\newblock \showarticletitle{Autofocus System and Evaluation Methodologies: A
  Literature Review}.
\newblock \bibinfo{journal}{\emph{Sensors and Materials}} \bibinfo{volume}{30},
  \bibinfo{number}{5} (\bibinfo{year}{2018}), \bibinfo{pages}{1165--1174}.
\newblock


\bibitem[Zhang et~al\mbox{.}(2019)]%
        {zhang2019image}
\bibfield{author}{\bibinfo{person}{Zhifei Zhang}, \bibinfo{person}{Zhaowen
  Wang}, \bibinfo{person}{Zhe Lin}, {and} \bibinfo{person}{Hairong Qi}.}
  \bibinfo{year}{2019}\natexlab{}.
\newblock \showarticletitle{Image super-resolution by neural texture transfer}.
  In \bibinfo{booktitle}{\emph{CVPR}}. \bibinfo{pages}{7982--7991}.
\newblock


\bibitem[Zou et~al\mbox{.}(2023a)]%
        {zou2023refvsr++}
\bibfield{author}{\bibinfo{person}{Han Zou}, \bibinfo{person}{Masanori
  Suganuma}, {and} \bibinfo{person}{Takayuki Okatani}.}
  \bibinfo{year}{2023}\natexlab{a}.
\newblock \showarticletitle{RefVSR++: Exploiting Reference Inputs for
  Reference-based Video Super-resolution}.
\newblock \bibinfo{journal}{\emph{arXiv preprint arXiv:2307.02897}}
  (\bibinfo{year}{2023}).
\newblock


\bibitem[Zou et~al\mbox{.}(2023b)]%
        {zou2023geometry}
\bibfield{author}{\bibinfo{person}{Han Zou}, \bibinfo{person}{Liang Xu}, {and}
  \bibinfo{person}{Takayuki Okatani}.} \bibinfo{year}{2023}\natexlab{b}.
\newblock \showarticletitle{Geometry Enhanced Reference-Based Image
  Super-Resolution}. In \bibinfo{booktitle}{\emph{Proceedings of the IEEE/CVF
  Conference on Computer Vision and Pattern Recognition}}.
  \bibinfo{pages}{6123--6132}.
\newblock


\end{thebibliography}

\clearpage

\appendix

\section{Prototype Details}
\label{app:prototypes}

\begin{figure}
\centering
\includegraphics[width=0.48\linewidth]{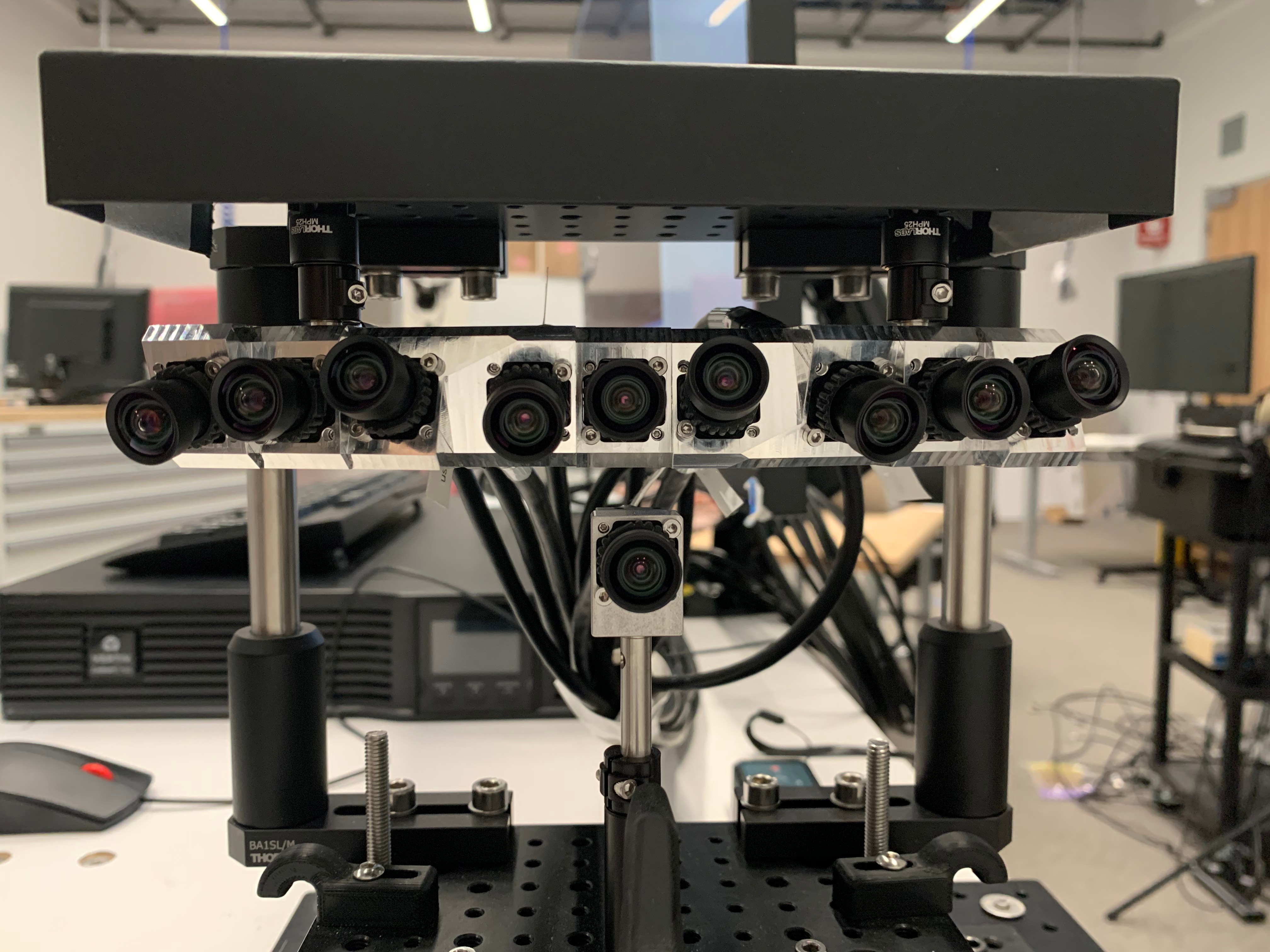}
\includegraphics[width=0.48\linewidth]{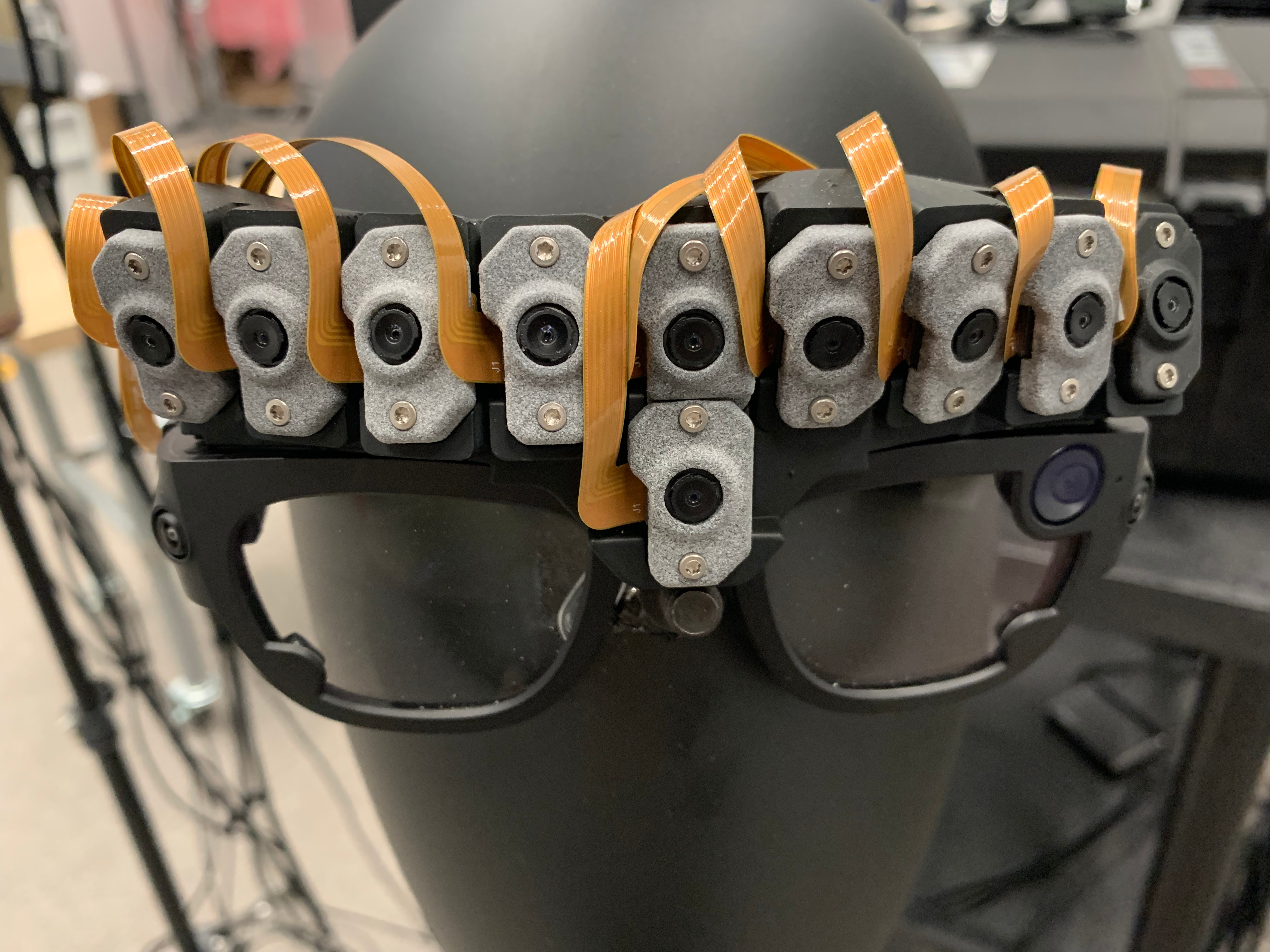}
\caption{\label{fig:protoA_protoB}
    Pictures of the prototypes. Left: desk-mounted prototype. Right: head-mounted prototype}
\end{figure}

\subsection{Desk-Mounted Prototype} 
As shown in the left of Fig~\ref{fig:protoA_protoB}, this system consists of 10 XIMEA xiMU RGB 18\,MPixel cameras (MU181CR-ON, 1.25\,$\mu$m pixel size) with identical Commonlands CIL039 lenses (3.8\,mm focal length, f-number 2.8). 
Nine detail cameras are placed along a line to approximate how tiny cameras can roughly be placed at the eyebrow region of the glasses. As discussed in the main paper,  they are oriented such that they cover the entire FOV of the guide camera in a $3\times3$ grid starting from a minimum distance of 0.8\,m.
The amount of overlap between adjacent cameras increases with further distance. 
The guide camera is placed in the center below the detail cameras, which maps to the nose bridge region in real glasses.
We note that this prototype has an average angular resolution that is close to 1\,arcmin and thus its depth of field is not sufficient to cover the full range from 0.4\,m to infinity.
The prototype is mounted on a rigid desk and is capable of capturing images of static scenes.

We cropped the images as needed to make sure that the system meets our design that the detail cameras cover everything placed 0.8\,m and beyond away from the camera inside the guide camera's FOV in a $3\times3$ grid. 
We took the central $3072\times3072$ pixels from the guide camera to be our target image. This image was downscaled to $768\times768$ to mimic the guide image using a prefiltering blur kernel and adding appropriate noise. 
We obtained the detail images by cropping out the central 1200x1200 pixels from the detail cameras, and there too added the necessary simulated degradation. We provide more details about the degradation process for both guide and detail cameras are provided in Appendix \ref{app:degrade}.

\subsection{Head-Mounted Prototype}

The XIMEA cameras are not capable of capturing high-speed videos that are suitable for burst mode. 
Thus, we built the second prototype for head-mounted dynamic capture which is shown in the right panel of Fig.~\ref{fig:protoA_protoB}. It uses 10 camera modules with 8 megapixel resolution that are identical to the POV cameras on Project Aria glasses \cite{Engel:2023:PAA}. 
The cameras are mounted on top of a modified pair of Aria glasses, which can provide inertial data from the built-in IMUs, and can be worn by a user. 
The detail cameras are placed close to the eyebrow line. The guide camera is mounted right below the detail cameras in the center of the rig. 
The cameras are connected to local storage through flexible printed circuits (FPC) in order to minimize the weight of the head-mounted device and to allow close to natural head motion.

Similar to what we did for the desk-mounted prototype, we cropped out the central $1600\times1200$ pixels in the guide camera as the target image, and cropped the central 640x480 pixels from the detail cameras as detail images. 
When we need high frame rate capture, the guide image is captured in $2\times2$ binning mode to save bandwidth. The highest frame rate we can achieve with stable capture is 120\,fps. The $2\times2$ binning operation is applied by the camera's firmware and keeps the Bayer pattern after the binning. Thus, even though we get a raw image resolution at $800\times600$, the effective resolution is lower.

\section{Degradation Strategies}
\label{app:degrade}

We model the main properties of the prototype cameras -- blur and noise -- experimentally and in simulation, to create models for the properties of cameras and lens systems at the approximate target scale. 
We will first talk about how we obtain the degradation models and how these models would be applied to degradation-free synthetic images. Then, we will elaborate how to apply the same degradation to images captured from real cameras, which are inevitably already degraded during the physical imaging process.

\subsection{Camera Blur}
\label{app:blur}

We designed lenses for the guide camera in our target system that are capable of achieving the target specs while meeting the form-factor requirements. 
We then model the blur for the lenses through Zemax simulation. 
Thus, for each camera design, we simulate the point spread function (PSF) over the FOV in a grid (7x7 for desk-mounted prototype simulation and 6x8 for head-mounted prototype simulation). 
To simulate a fully spatially varying blurred image from clean synthetic images, we break up the FOV into $7\times7$ or $6\times8$ overlapping squares. In each square, we convolve with the corresponding PSF, and then we smoothly blend between the different regions to create the image.

\subsection{Camera Noise}
\label{sec:noise}

Multiple factors could affect the noise level in camera sensors. Without the ability to build the target camera system, it is hard to derive precisely what would be the noise level in the system. Luckily, we can theoretically reason that the noise in both detail and guide cameras in the target system will be similar to the noise level in the cameras from our existing prototypes. 
Consequently, we can measure the noise levels in our prototype cameras and treat them as the targeted noise model. When we add noise to a clean synthetic image, we just add simulated noise according to this targeted noise model. 

Now, all we need is the noise model from the prototype cameras (XIMEA/Aria cameras). 
For short exposure times, noise in camera images is dominated by two noise sources: \textit{Shot noise} is caused by the photon nature of light, i.e., by the random arrival times of individual photons at the sensor. It can be modeled via a Poisson distribution and changes with the amount of light collected. \textit{Read noise} is the noise arising from imprecision in the signal readout process. It is independent of the amount of light collected and the exposure time. Overall, the relationship between incident illumination $x$ and digital values $y$ can be modeled as 
\[ y \sim \mathcal{N}(\mu = x, \sigma^2 = \lambda_{\text{read}} + \lambda_{\text{shot}}x)
\label{eq:image-noise}
\]
\cite{Brooks:2018:UIL}. 
\begin{figure}
    \centering
    \includegraphics[width=0.48\columnwidth]{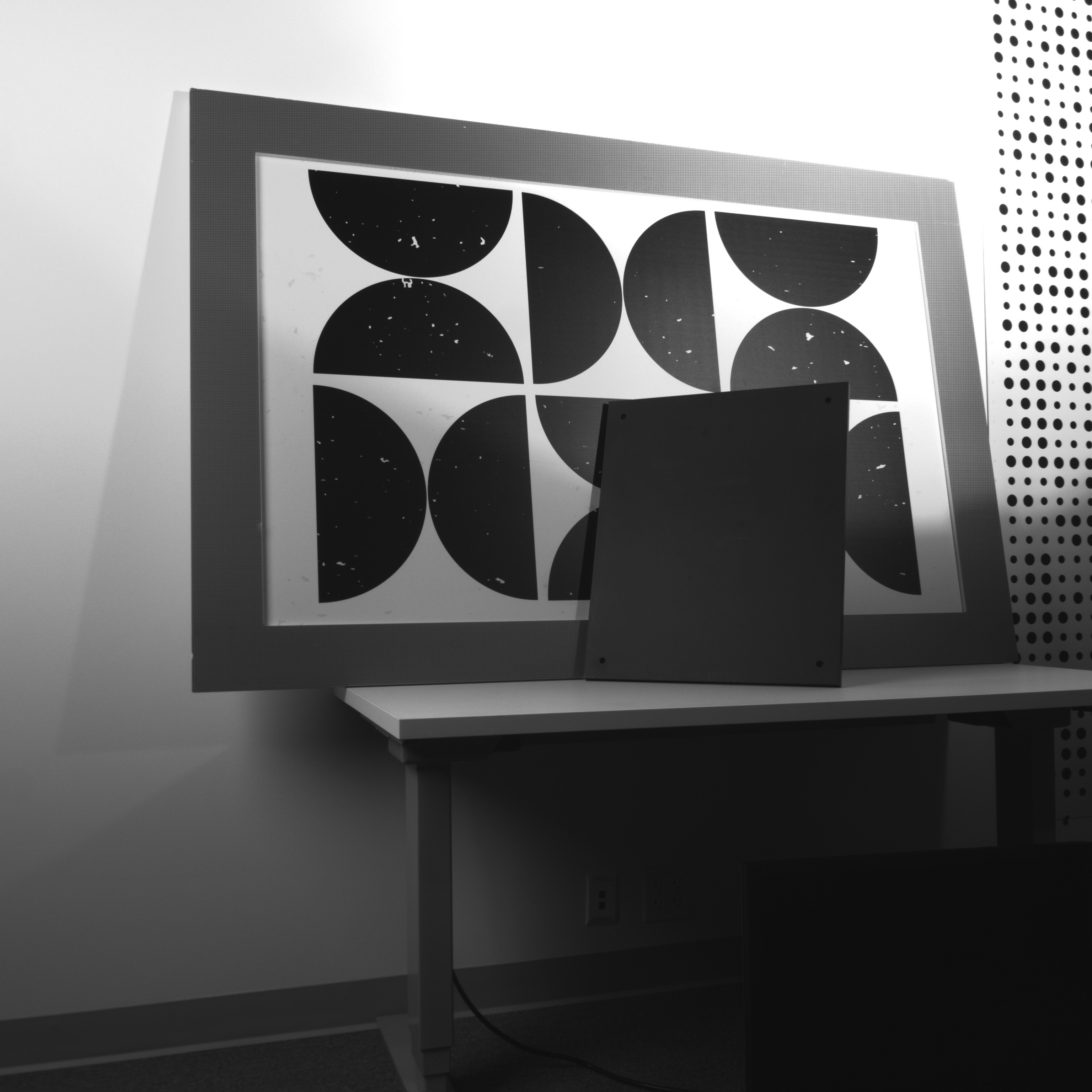}
    \hfill
    \includegraphics[width=0.48\columnwidth]{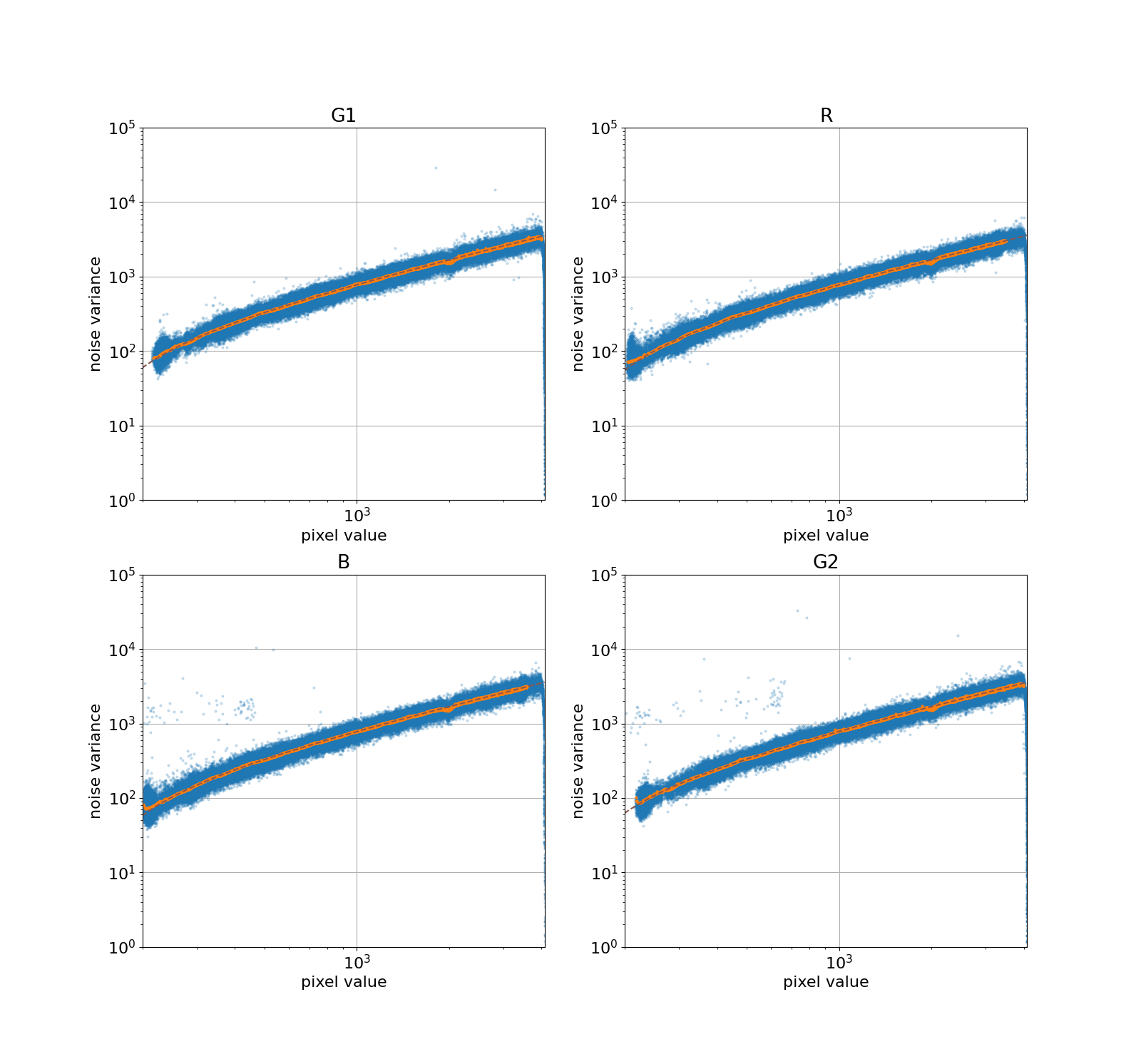}
    \caption{Left: Example of a noise measurement image from XIMEA camera using gain=1. In total 100 images shot at exactly the same condition were used to calculate the mean variance plot and create the fitted noise model. Right: Mean variance plot that includes fitted noise model in dashed red line as well as the raw variance measurements in blue dots. The orange dots show the average of the raw variance measurements for each input intensity value.}
    \label{fig:ximea-mean-variance}
\end{figure}
In order to determine the (signal-dependent) variance for a specific camera and gain setting, we collect a large number of images of a static scene with fixed camera pose and settings. We then determine for each pixel the mean and variance of its sampled values and plot them in a mean variance plot as in Figure \ref{fig:ximea-mean-variance}. Fitting a straight line to the data determines the parameters of the signal-dependent noise model in Eq. \ref{eq:image-noise} that we can use to add simulated noise to a clean, noise-free image. 

\begin{table}[]
    \centering
    \begin{tabular}{|c|c|c|c|c|}
    \hline
         Prototype & Gain & Color & $\lambda_\text{shot}$ & $\lambda_\text{read}$ \\ \hline
         Desk-mounted & 1 & R/G/B & $2.4\times10^{-4}$ & $1.5\times10^{-6}$ \\ \hline
         Head-mounted & 1 & R & $1.1\times10^{-4}$  & $2.9\times10^{-6}$ \\ \hline
         Head-mounted & 1 & G/B & $1.2\times10^{-4}$ & $2.9\times10^{-6}$ \\ \hline
         Head-mounted & 22 (max) & R/B & $2.1\times10^{-3}$ & $1.7\times10^{-5}$ \\ \hline
         Head-mounted & 22 (max) & G & $2.2\times10^{-3}$ & $1.7\times10^{-5}$ \\ \hline
    \end{tabular}
    \caption{
    The fitted noise parameters for the two prototypes assuming signal is normalized, i.e. $x\in[0, 1]$. } 
    \label{tab:noise}
\end{table}

Furthermore, we can add additional noise to a noisy signal $y_1$ using the sum of normally distributed variable in order to simulate a higher noise signal $y_2$ from a lower noise capture as
\[ y_2 \sim \mathcal{N} \left(\mu = y_1, \sigma^2 = \lambda_{\text{read,2}}-\lambda_{\text{read,1}} 
+ \left(\lambda_{\text{shot,2}} - \lambda_{\text{shot,1}}\right) y_1\right)
\label{eq:image-noise-transfer}
\]
This is an approximation since $y_1$ is a noisy measurement of $x_1$, which cannot be precisely measured.

\subsection{Degradation Applied to Real-World Captured Images}

When we are using real XIMEA/Aria cameras to capture data, the raw images we obtain are already blurred and noise-contaminated, so we cannot just add all these degradation on top of the raw data as if they are clean images. We examine them one by one:

\begin{itemize}
\item \textbf{Guide camera's blur:} For guide cameras, because we directly capture high-resolution target images, we need to downsample them by an aggressive factor of 4 to obtain the real guide image. 
We estimated the PSF using the blackbox lens model for the exact lens used in the system, and the PSF kernel is much smaller than a 4x4 box, indicating that we can safely assume the existing blur from the lens is negligible after downsampling.
As a result, we need to apply the blurring degradation as if the input image is blur-free. 
There is an exception when we use the 2x2 binning for the head-mounted prototype. Because the effective resolution is lower than the digital pixels we get due to the binning process, we do not apply extra blur degradation to the raw data.
\item \textbf{Guide camera's noise:} Because the 4x4 downsampling is quite aggressive, we can also assume that after downsampling, the existing noise is negligible. Thus, we need to add the noise as if the input image is noise-free.
\item \textbf{Detail camera's blur:} The detail images from our prototypes have more blur than detail images from our tiny guide camera. The lenses has a larger f-number, so the diffraction limited spot size is larger than in our detail camera lens. As a result, we do not add additional blur to the captured raw detail images.
\item \textbf{Detail camera's noise:} As we earlier concluded the expected tiny camera noise is similar to XIMEA/Aria camera noise level, we do not add any additional noise to the captured raw detail images.
\end{itemize}

\section{Head motion while keeping intentionally still}
\label{sec:head_motion_user_stabilized_supplementary}

\begin{figure*}[tp]
    \centering
    \includegraphics[width=1\linewidth]{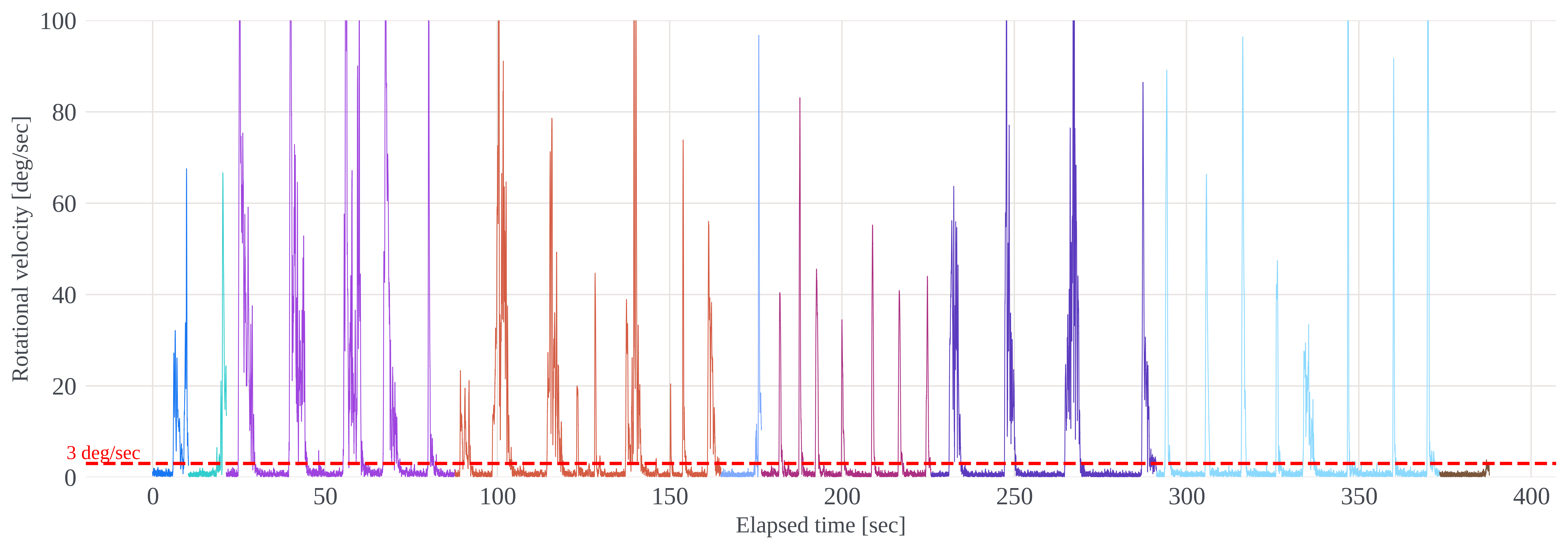}
    \caption{Visualization of Aria rotational velocity within a set of recordings in which users wore Aria glasses and intentionally fixated on static scene objects. Each colored plot represents a separate recording. The spikes in rotational velocity are due to the wearer shifting their fixation from one target to another. The red line indicates a threshold of 3 deg/sec between head-still and head-moving modes of user movement.}
    \label{fig:aria_head_still_threshold}
\end{figure*}

As described in the main paper, we investigated  different user behavior as it applies to head motion during two categories of user motion: (1) general egocentric motion where a human moves naturally through their environment and performs everyday tasks, and (2) intentional head-still motion where a human is fixated on an object of interest. Fig.~\ref{fig:aria_head_still_threshold} illustrates how we determined a threshold of rotational velocity for head-still activities. Our recordings consisted mostly of low rotational velocity, punctuated by intermittent high rotation as the user moved their fixation between different static objects in their surroundings. A threshold of 3 deg/sec categorizes this behavior well.

\begin{figure}[tp]
    \includegraphics[width=\linewidth]{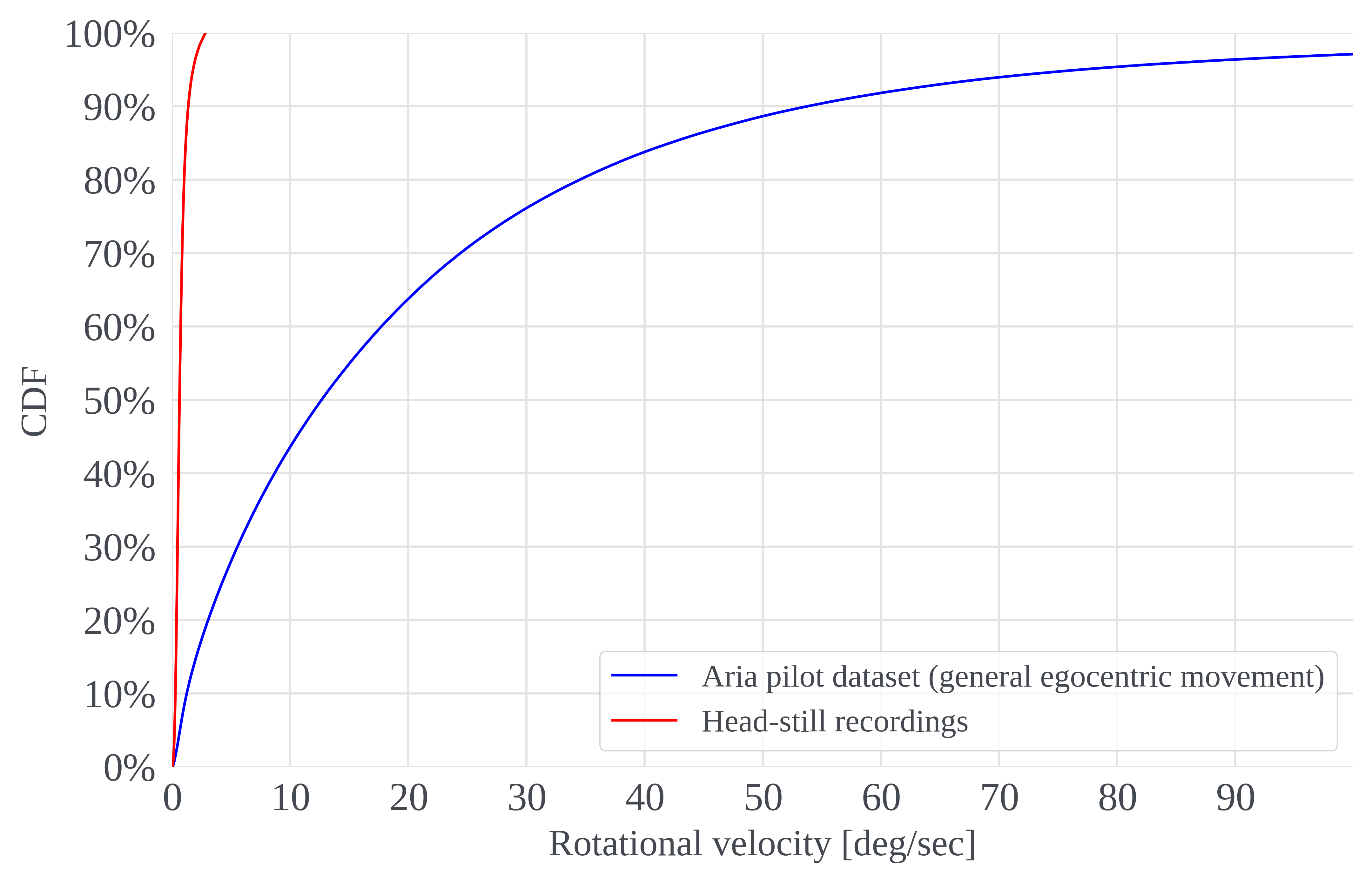}
    \caption{Cumulative distribution function (CDF) of instantaneous rotational velocities $\omega$ from egocentric Aria recordings. Blue line: Aria Pilot Dataset containing general egocentric motion. Note the high rotational velocity. Red line: selected sub-sequences from a different set of recordings in which the wearers were asked to kept their head intentionally still (< 3 deg/sec) while observing objects in their field of view. Parts of sequences where the wearer's head moved faster (i.e. moving from one object of interest to another) are discarded. See Fig.~\ref{fig:aria_head_still_threshold} for an illustration of how we found this threshold.} 
    \label{fig:head_motion}
\end{figure}

Fig.~\ref{fig:head_motion} shows the distribution of rotational velocities for the two sets of recordings we used in our analysis: the Aria Pilot Dataset (blue line) and our selected portions of recordings of head-still activities (red line). Unsurprisingly, as we constrain our head-still data to 3 deg/sec, the distribution of rotational velocities in that data is similarly constrained.

\begin{figure*}
    \centering
    \includegraphics[width=1\linewidth]{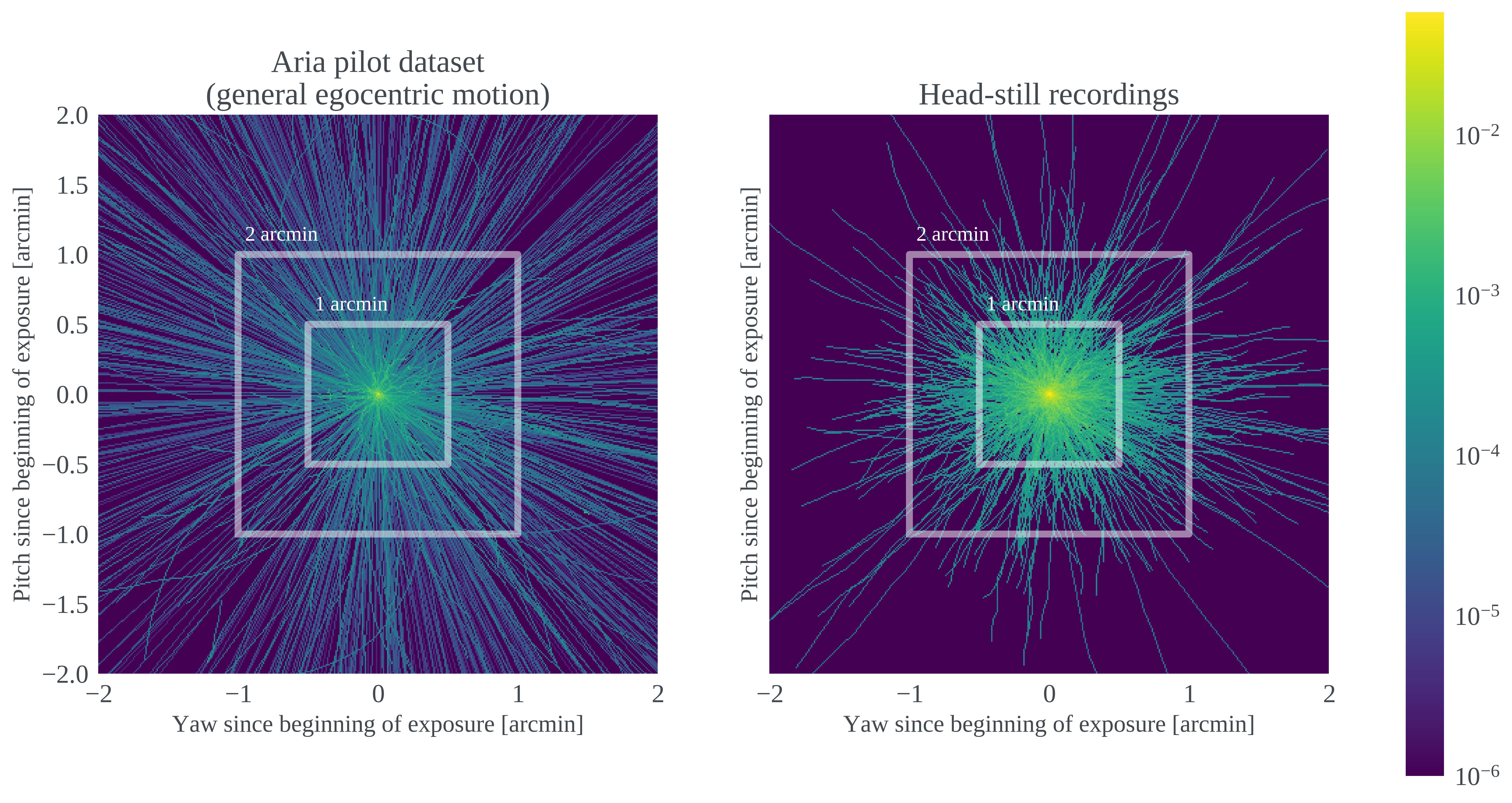}
    \caption{Illustration of head-motion traces for Aria pilot dataset (left) and head-still recordings (right). All traces were drawn by selecting a random timestamp in the dataset and integrating rotational IMU data forward in time by 25ms. X/Y positions in the plot illustrate yaw/pitch movement since the beginning of the simulated exposure. The pixels of each trace were normalized before summing all traces together, so that the total contribution of each trace would represent the same amount of light during an exposure. Values are colored here in log-scale for visibility.}
    \label{fig:head_motion_traces}
\end{figure*}

Further insight into these two categories of head motion can be seen when visualizing traces of rotational motion during simulated exposures (Fig.~\ref{fig:head_motion_traces}). For each dataset, we simulated 1000 exposures, each of exposure time 25ms, by randomly selecting 1000 timestamps and integrating the rotational IMU data for 25ms. We then plotted these traces where the X-axis represents the yaw since the beginning of the exposure, and where the Y-axis represents the pitch since the beginning of the exposure (each trace begins at the origin). The brightness of each trace was normalized to accurately represent the spread of a constant point light source over a fixed exposure of 25ms (i.e., each point on a longer trace would be drawn more dimly than the points on a shorter trace).

The resulting traces show that for the most part, general egocentric rotational motion is more linear (Fig.~\ref{fig:head_motion_traces}, left) when compared with head-still activities (Fig.~\ref{fig:head_motion_traces}, right). Furthermore, much of the head-still traces remain close (within 1-2 arcmin) to the origin, suggesting that a user can adequately stabilize their head during exposure.

\end{document}